\documentclass{article}

\usepackage{arxiv}

\usepackage[T1]{fontenc}
\usepackage{microtype}
\usepackage{xspace}
\usepackage[dvipsnames]{xcolor}
\usepackage{graphicx}
\usepackage{amsmath}
\usepackage{amssymb}
\usepackage{booktabs}
\usepackage[hyphens]{url}
\usepackage[numbers,sort&compress]{natbib}
\usepackage{etoolbox}
\usepackage[format=plain,labelformat=simple,labelsep=period,font=small,compatibility=false]{caption}
\usepackage[font=footnotesize,skip=3pt,subrefformat=parens]{subcaption}
\usepackage[shortlabels,inline]{enumitem}
\usepackage{wrapfig}

\usepackage{multirow}
\usepackage{algorithm}
\usepackage{algpseudocode}

\usepackage{tcolorbox}
\newtcolorbox{promptbox}[1]{colback=gray!6,colframe=gray!55,boxrule=0.5pt,arc=1pt,left=4pt,right=4pt,top=3pt,bottom=3pt,fonttitle=\small\bfseries,title={#1}}

\definecolor{linkblue}{rgb}{0.21,0.49,0.74}
\usepackage[pagebackref,breaklinks,colorlinks,allcolors=linkblue]{hyperref}
\usepackage[capitalize]{cleveref}
\crefname{section}{Sec.}{Secs.}
\Crefname{section}{Section}{Sections}
\crefname{table}{Tab.}{Tabs.}
\Crefname{table}{Table}{Tables}

\makeatletter
\DeclareRobustCommand\onedot{\futurelet\@let@token\@onedot}
\def\@onedot{\ifx\@let@token.\else.\null\fi\xspace}
\def\eg{\emph{e.g}\onedot} 
 
\def\etal{\emph{et al}\onedot}

\let\arxiv@figure\figure
\let\arxiv@endfigure\endfigure
\renewenvironment{figure}[1][tbp]
  {\arxiv@figure[#1]\centering\begin{minipage}{0.75\textwidth}}
  {\end{minipage}\arxiv@endfigure}
\let\arxiv@table\table
\let\arxiv@endtable\endtable
\renewenvironment{table}[1][tbp]
  {\arxiv@table[#1]\centering\begin{minipage}{0.6\textwidth}}
  {\end{minipage}\arxiv@endtable}
\newsavebox{\arxiv@box}
\let\arxiv@resizebox\resizebox
\renewcommand{\resizebox}[3]{%
  \sbox{\arxiv@box}{\arxiv@resizebox{#1}{#2}{#3}}%
  \ifdim\wd\arxiv@box>\linewidth\arxiv@resizebox{\linewidth}{!}{\usebox{\arxiv@box}}\else\usebox{\arxiv@box}\fi}
\makeatother

\title{ProbFaS: Probabilistic Video Frame Selection for MLLMs via Multi-Fidelity Calibration}

\author{%
  Tianru Zhang\textsuperscript{1}
  \quad Mayank Nautiyal\textsuperscript{1,2} 
  \quad Hugo Eidmann\textsuperscript{1}
  \quad Prashant Singh\textsuperscript{1,2}
  \quad Ekta Vats\textsuperscript{1}\\[3pt]
  {\normalfont\textsuperscript{1}Department of Information Technology, Uppsala University, Uppsala, Sweden}\\
  {\normalfont\textsuperscript{2}SciLifeLab, Uppsala University, Uppsala, Sweden}%
}

\begin{document}
\maketitle
\addtocontents{toc}{\protect\setcounter{tocdepth}{-1}}
\begin{abstract}
Processing long videos with multimodal large language models (MLLMs) is prohibitively expensive, making informative frame selection essential for accurate video understanding.
The default uniform sampling often misses task-relevant keyframes, while recent training-free selectors instead pick frames by how well each one matches the question under an image--text alignment (ITA) score.
However, we show that ITA signals correspond poorly with the MLLM's true needs: their rankings correlate weakly with downstream accuracy and are even negatively correlated on over 40\% of questions, whereas the MLLM's internal yes/no judgment yields a much stronger signal.
We present ProbFaS, a training- and label-free probabilistic pipeline that fuses both signals for key frame selection.
ProbFaS treats dense ITA scores as cheap, biased measurements and sparse MLLM judgments as accurate, costly ones, unifying them via co-kriging, a multi-fidelity Gaussian process, to calibrate relevance across all frames.
A temporal determinantal point process then selects frames capturing relevance peaks while preserving temporal coverage.
Across Video-MME, MLVU and LongVideoBench, ProbFaS is the most accurate among fourteen state-of-the-art training-free selectors and uniform sampling in 13 of 15 settings (8--128 frames), outperforming the strongest baseline by up to 3.1 percentage points.
On MLVU and LongVideoBench, 32 ProbFaS frames match or surpass the top baseline at 128 frames and uniform sampling with up to 768 frames.
\end{abstract}

\section{Introduction}
\label{sec:intro}

\begin{wrapfigure}{r}{0.5\textwidth}
    \centering
    \vspace{-\baselineskip}
    \includegraphics[width=0.55\linewidth]{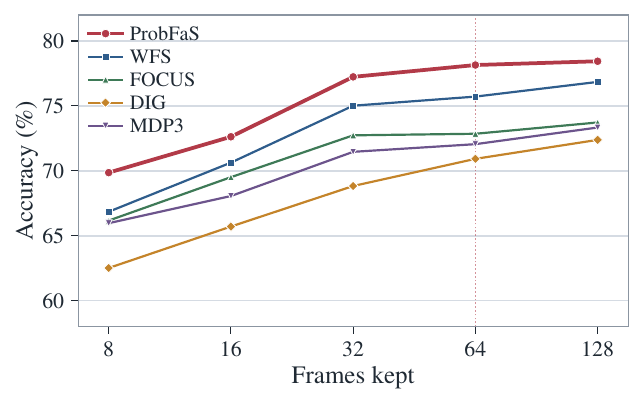}\hfill
    \includegraphics[width=0.44\linewidth]{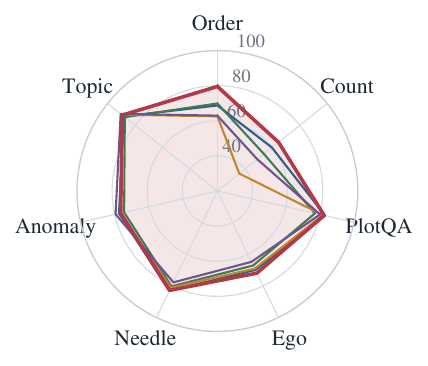}
    \caption{ProbFaS against state-of-the-art training-free selectors on MLVU dev~\cite{zhou2025mlvu}: official score against the number of selected frames (left) and accuracy on the seven tasks at 64 frames (right; dotted line on the left).}
    \label{fig:quantitative}
\end{wrapfigure}

\begin{figure}[t]
    \centering
    \includegraphics[width=\linewidth]{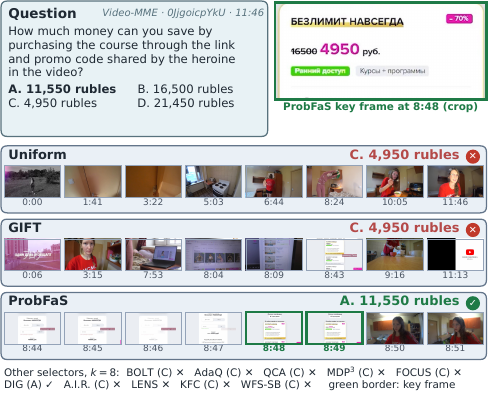}
    \caption{Frames chosen by uniform sampling, GIFT and ProbFaS ($k=8$) for a Video-MME question, and the answers of Qwen3-VL-8B.
    ProbFaS follows the checkout from the promo code (8:44) to the discounted price (green; crop at top right), from which the saving follows.
    Uniform sampling shows no price, and GIFT only the full price before the code is applied (8:43); without the discount in view, the model answers 4,950 rubles, as it does with no frame at all.}
    \label{fig:teaser_qual}
\end{figure}

Current multimodal large language models (MLLMs) typically represent video inputs via sampled frames, each encoded into a sequence of visual tokens~\cite{bai2025qwen3vl,wang2025internvl35,zhang2024llavavideo}.
A long video holds thousands of frames, but the context window, memory and latency limit a model to only a few dozen or a few hundred frames, making frame selection critical to performance. Standard uniform sampling can easily miss the brief window a question focuses on (\cref{fig:teaser_qual}; more examples in Supp.~\ref{sec:suppl_qualitative}).
Training-free frame selectors offer an attractive solution, as they are lightweight, work with any frozen MLLM, and require no fine-tuning.

Most existing training-free selectors share a common paradigm~\cite{tang2025aks,liu2025bolt,sun2025mdp3,chen2026wfssb,zhang2026adaq,peng2026qca}: an off-the-shelf vision-language model, such as CLIP, SigLIP, or BLIP~\cite{radford2021clip,zhai2023siglip,li2023blip2}, scores the image--text alignment (ITA) between each frame and the question, after which a selection rule turns the scores into a frame set that balances relevance against diversity or coverage.
Though selection rules vary, they all assume that a frame ranked highly by the scoring model is necessarily more useful to the downstream MLLM.
We test this assumption directly (\cref{sec:motivation}) by measuring how much each frame helps the MLLM answer.
The agreement is weak: on Video-MME~\cite{fu2025videomme} the median within-question rank correlation of these alignment scores with actual frame usefulness is at most 0.08, with the ranking negatively correlated on over 40\% of questions, and neither scaling to stronger scoring models nor fine-grained peak searching resolves this misalignment.
The answering MLLM itself is a consistently stronger judge: evaluating frames via a simple label-free yes/no query yields a top-ranked frame that correctly answers 60.1\% of questions, compared to at most 54.4\% for alignment scores.
However, querying the downstream MLLM across every frame of a long video introduces prohibitive computational costs, recreating the exact inference bottleneck frame selection aims to solve.

We introduce ProbFaS (probabilistic frame selection), which treats the relevance of a frame as a latent function of time observed through two measurements of different fidelity (\cref{fig:ProbFaS_pipeline}): dense alignment scores, which serve as a cheap but biased prior, and yes/no judgments of the answering MLLM on a small subset of frames, which act as accurate but costly references.
ProbFaS fuses them by co-kriging~\cite{kennedy2000cokriging}, a multi-fidelity Gaussian process (GP) that learns per question how the alignment score maps to the MLLM's judgment and propagates the remaining bias along the timeline while retaining its sharp peaks.
A temporal determinantal point process (DPP) then selects the $k$ optimal frames from this calibrated relevance, capturing relevance peaks while maintaining temporal coverage over the video.
ProbFaS requires no training, no labels and no auxiliary language model, incurring only minimal overhead from single-token logit evaluations on a small subset of frames.

Our contributions are:
\begin{itemize}
    \item An analysis showing that the alignment scores used by training-free frame selectors poorly reflect what the answering MLLM needs, whereas an MLLM's own label-free judgment is a significantly more accurate signal.
    \item ProbFaS, to our knowledge the first frame selector that casts relevance estimation as multi-fidelity probabilistic inference: co-kriging fuses dense low-fidelity alignment scores with sparse judgments of the answering MLLM into a calibrated per-frame relevance, while a probabilistic sampling model, temporal DPP, selects relevant yet temporally diverse frames without a hand-set trade-off. The pipeline is training- and label-free and works with any pre-trained MLLM.
    \item We demonstrate the effectiveness of ProbFaS on Video-MME, MLVU~\cite{zhou2025mlvu} and LongVideoBench~\cite{wu2024longvideobench} against fourteen training-free selectors at budgets of 8 to 128 frames (\cref{fig:quantitative}): it is the most accurate in 13 of 15 settings, its 32 frames match or surpass the best baseline at 128 frames on MLVU and LongVideoBench (\cref{tab:efficiency_frames}), and its frame selections transfer across different answering models.
\end{itemize}

\section{Related Work}
\label{sec:related}

\paragraph{Long video understanding with MLLMs.}
Recent MLLMs extend the context length of the language model (LongVA~\cite{zhang2024longva}) or train directly on long videos (LLaVA-Video~\cite{zhang2024llavavideo}, Qwen2.5-VL~\cite{bai2025qwen25vl}, Qwen3-VL~\cite{bai2025qwen3vl}, InternVL3.5~\cite{wang2025internvl35}), while alternative pipelines compress visual tokens via pruning (FastV~\cite{chen2024fastv}, DyCoke~\cite{tao2025dycoke}) or merging (LongVU~\cite{shen2025longvu}, VideoChat-Flash~\cite{li2026videochatflash}).
However, memory consumption and latency still scale with frame count; thus, MLLMs in practice receive a constrained frame budget, making selective frame sampling essential for effective long video understanding.

\paragraph{Training-free frame selection.}
Most training-free selectors score each frame against the question using an image--text alignment score (e.g., CLIP~\cite{radford2021clip}, SigLIP~\cite{zhai2023siglip}, or BLIP~\cite{li2023blip2}) and balance relevance against coverage by recursive splitting (AKS~\cite{tang2025aks}), sampling from the score distribution (BOLT~\cite{liu2025bolt}, AdaQ~\cite{zhang2026adaq}, Q-Frame~\cite{zhang2025qframe}, which also lowers the resolution of less relevant frames), a DPP (MDP$^3$~\cite{sun2025mdp3}, LDDR~\cite{chen2026lddr}), an integer quadratic program (Nar-KFC~\cite{fang2026narkfc}), greedy subset objectives (MarKey~\cite{shi2026markey}, FORGE~\cite{kaviani2026forge}), relevance-conditioned uniqueness (GIFT~\cite{ma2026gift}), wavelet denoising with segment boundaries (WFS-SB~\cite{chen2026wfssb}), the shape of the score curve (RIDGE~\cite{xu2026ridge}) or query-guided partitions (QCA~\cite{peng2026qca}).
Other approaches extract attention weights from a smaller MLLM (DAFS~\cite{wang2026dafs}), score candidate frames (FRAG~\cite{huang2025frag}, UG-Sample~\cite{kim2025ugsample}) or frame groups (GridProbe~\cite{eltahir2026gridprobe}) with the answering MLLM, spread sparse MLLM scores linearly (OTFS~\cite{zhao2026otfs}) or over a segment graph (VideoDetective~\cite{yang2026videodetective}), ground evidence via coverage heuristics (GCR~\cite{wei2026gcr}), adapt to the query by routing (DIG~\cite{li2026dig}) or spatial--temporal zooms (LENS~\cite{zhang2026lens}), validate CLIP proposals with MLLMs (A.I.R.~\cite{zou2026air}), refine the sampling distribution from object detections (T*~\cite{ye2025tstar}, Logic-in-Frames~\cite{guo2025vsls}), reduce scoring cost by multi-armed bandits (FOCUS~\cite{zhu2026focus}), adaptive budgets (CSES~\cite{zhang2026cses}) or a GP surrogate (FORTE~\cite{huang2026forte}) of the alignment score, or select frames maximizing feature volume without considering the query (MaxInfo~\cite{li2026maxinfo}); MetaSampling~\cite{dahal2026metasampling} re-runs any selector per interval to pass fewer frames.
Most of these selectors equate raw alignment scores with the MLLM's internal utility, mirroring information retrieval systems that conflate passage relevance with task utility~\cite{zhang2024utility,zhang2026llmutility}.
However, as shown in \cref{sec:motivation}, this alignment score is inherently weak and cannot be repaired by post-hoc score reshaping; ProbFaS instead measures the answering MLLM's own judgment on a sparse frame set and models it probabilistically across the entire frame sequence.
Unlike OTFS, which interpolates sparse MLLM rewards linearly, ProbFaS uses the calibrated alignment score to fill the gaps between probes.

\paragraph{Agent- and learning-based selection.}
Agentic methods let LLMs or VLMs iteratively caption, retrieve or re-query frames.
VideoAgent~\cite{wang2024videoagent} lets an LLM judge whether the captions so far suffice and otherwise retrieve frames with CLIP for queries it generates, VideoTree~\cite{wang2025videotree} clusters frames and expands the clusters an LLM rates relevant into a finer hierarchy, Temporal Chain of Thought~\cite{arnab2025tcot} lets the VLM itself pick the relevant frames of each segment before answering, and Vgent~\cite{shen2025vgent} retrieves clips from an entity graph and filters them by yes/no verification with an LVLM.
Each question thus triggers several rounds of generation by large models, often with captions of many frames, so selection can cost far more than the answer.

Learning-based selectors are trained on pseudo-labels or answerer feedback: SeViLA~\cite{yu2023sevila} refines a BLIP-2 keyframe localizer with pseudo-labels from its answerer, Frame-Voyager~\cite{yu2025framevoyager} learns to pick frame combinations ranked by a Video-LLM's prediction loss, the selector of Hu~\etal~\cite{hu2025mllmselector} is a small MLLM trained on pseudo-labels of frame importance, GenS~\cite{yao2025gens} generates relevant frame spans with scores learned from GPT-4o labels, VideoITG~\cite{wang2025videoitg} learns instructed frame selection from automatic annotations, ReQuest~\cite{kim2026request} distills an MLLM into a question-aware selector that rethinks when uncertain, and FFS~\cite{buch2025ffs} learns a policy that keeps a flexible number of frames.
Others train relevance scorers contrastively, FrameScout~\cite{hu2026framescout} with a frame--query objective for streaming selection and evidential keyframe sampling~\cite{wang2026evidential} on annotated answer-supporting segments, or learn to select by reinforcement learning: TSPO~\cite{tang2026tspo} trains an event-aware temporal agent, ViaRL~\cite{xu2025viarl} rewards a frame selector by answer accuracy, VideoMiner~\cite{cao2025videominer} searches a tree of segments, captions and clusters, K-frames~\cite{yao2025kframes} predicts query-relevant scene clips and converts them into any number of keyframes, and MSJoE~\cite{tan2026msjoe} jointly trains the MLLM and a sampler over CLIP similarities to the MLLM's own queries.
Video-MTR~\cite{xie2025videomtr} and FrameThinker~\cite{he2026framethinker} go further and train the MLLM itself to request segments or zoom into frames over several reasoning turns.
These approaches need curated data and backbone-specific training.
ProbFaS requires neither interactive generation loops nor fine-tuning: like SeViLA~\cite{yu2023sevila}, it uses the answering MLLM as a judge at test time, but reads a single-token logit per probed frame, with no generation and no auxiliary language model.

\paragraph{Multi-fidelity GPs and DPPs.}
Gaussian processes (GPs)~\cite{rasmussen2006gp} are standard surrogates for expensive functions, and GP-UCB~\cite{srinivas2010gpucb} uses their uncertainty to decide where to evaluate next.
Co-kriging~\cite{kennedy2000cokriging} models a high-fidelity, costly signal as a scaled low-fidelity observation plus a GP discrepancy, enabling sparse accurate measurements to calibrate dense biased priors; to our knowledge, it has not been applied to video frame selection.
FORTE~\cite{huang2026forte} also places a GP over time, but to decide which frames to score with the alignment model, whereas ProbFaS uses it to correct the alignment score toward the answering MLLM.
DPPs are widely used to sample diverse subsets~\cite{kulesza2012dpp,hennig2016exactdpp}, with greedy MAP inference~\cite{chen2018fastdpp} for the most probable subset, \eg in video summarization~\cite{gong2014seqdpp} and frame selection (MDP$^3$~\cite{sun2025mdp3}, LDDR~\cite{chen2026lddr}), where similarity is measured by visual features.
Simpler rules suppress the neighbors of each top-scoring pick (non-maximum suppression~\cite{hu2025mllmselector}) or greedily maximize a coverage objective~\cite{shi2026markey,kaviani2026forge}.
ProbFaS uses a temporal DPP, which measures similarity by the time between frames.

\section{Alignment Scores Miss MLLM Needs}
\label{sec:motivation}

Query-aware selectors assume that a frame with a higher alignment score is more useful to the answering MLLM.
Prior work doubts this from downstream accuracy or examples~\cite{kim2026request,wang2026evidential,wei2026gcr,yu2023sevila}; we quantify the gap frame by frame and independently of any budget, on Video-MME~\cite{fu2025videomme} with Qwen3-VL-8B~\cite{bai2025qwen3vl} as the answering model.

\paragraph{Usefulness and the MLLM probe.}
Given a question $q$ with its options and frame $f_t$ at time $t$, the MLLM is asked for an option letter. We define the \emph{usefulness} of the frame as the softmax of its next-token logits $\ell_c(t)$ over the letters $c\in\mathcal{C}$, evaluated at the correct letter $a^\star$:
\begin{equation}
u(t)=P(a^\star\mid f_t,q)=\exp\ell_{a^\star}(t)\big/\textstyle\sum_{c\in\mathcal{C}}\exp\ell_c(t).
\label{eq:usefulness}
\end{equation}
$u(t)$ needs the label and serves only as the reference.
We also conduct a label-free \emph{probe} that asks the same model whether $f_t$ helps answer $q$, Yes or No, and reads
\begin{equation}
y(t)=\ell_{\mathrm{Yes}}(t)-\ell_{\mathrm{No}}(t)=\log\frac{P(\mathrm{Yes}\mid f_t,q)}{P(\mathrm{No}\mid f_t,q)},
\label{eq:probe}
\end{equation}
so that its logistic sigmoid $\sigma(y(t))$ is the probability that the frame helps.
Both are computed on a small subset of frames per question (probe set of \cref{sec:probe}; prompts in Supp.~\ref{sec:suppl_prompts}) and compared with three alignment scores provided by CLIP-B/32~\cite{radford2021clip}, BLIP-2-ITM~\cite{li2023blip2}, and SigLIP-SO400M~\cite{zhai2023siglip} given the same frames and text.

\begin{figure}[h]
  \centering
  \includegraphics[width=\linewidth]{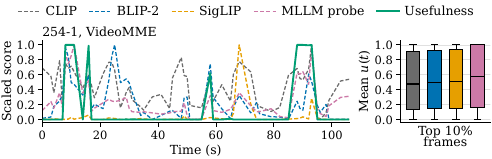}
  \caption{Usefulness, alignment scores and the MLLM probe on Video-MME. Left: results on question 254-1, each curve scaled to $[0,1]$. Right: per-question mean usefulness $u(t)$ of each score's top-10\% frames, over all questions.}
  \label{fig:score_usefulness}
\end{figure}

\paragraph{Alignment scores barely track usefulness, the probe does better.}
The median correlation of these scores with $u(t)$ is 0.06--0.08, it is negative on 41--44\% of questions, and answering from their top frame alone lifts accuracy only from 42.4\% with no image to 50.9--54.4\% (\cref{tab:score_usefulness}); a stronger alignment model barely helps.
On question 254-1 (\cref{fig:score_usefulness}, left), usefulness peaks at 8--11\,s and 88--93\,s, whereas SigLIP and BLIP-2 peak at 68 and 25\,s, on frames of zero usefulness.
The probe agrees clearly better: its median correlation is 0.17 (0.62 on question 254-1), answering from its top frame is correct on 60.1\% of questions, and its top 10\% of frames have a mean usefulness of 0.58 for the median question, against 0.48--0.51 for these scores (\cref{fig:score_usefulness}, right).
It correlates with BLIP-2 and SigLIP only at 0.43 and 0.50, so it carries information these scores lack.
Although Hu~\etal~\cite{hu2025mllmselector} found an MLLM's probability of Yes unreliable, this single-token judgment beats every alignment score, also on MLVU and LongVideoBench (Supp.~\ref{sec:suppl_bins}).

\begin{table}[h]
  \centering
  \small
  \setlength{\tabcolsep}{3.5pt}
  \begin{tabular}{lccc}
    \toprule
    Score & Spearman & Negative & Top-frame \\
    \midrule
    No image        & --   & --    & 42.4\% \\
    CLIP-B/32       & 0.06 & 43.6\% & 50.9\% \\
    BLIP-2-ITM      & 0.06 & 42.2\% & 52.1\% \\
    SigLIP-SO400M   & 0.08 & 41.4\% & 54.4\% \\
    Yes/no probe    & \textbf{0.17} & \textbf{36.7\%} & \textbf{60.1\%} \\
    \bottomrule
  \end{tabular}
  \vspace{0.1in}
  \caption{Agreement with single-frame usefulness $u(t)$ on Video-MME (2,700 questions). Spearman: median within-question rank correlation; Negative: questions where it is below zero; Top-frame: accuracy of the MLLM answering from the top-ranked frame alone.}
  \label{tab:score_usefulness}
\end{table}

\paragraph{Reshaping alignment scores does not help.}
If the alignment scores were the right target, finding more of their peaks would pay off.
Yet a search at half the scoring cost, which scores every fourth frame and then the gaps most likely to hide a higher score, finds 86\% of the 32 top-scoring frames (51\% for every second frame) without changing accuracy (67.1\% against 66.8\%; Supp.~\ref{sec:suppl_peak}).
Denoising~\cite{chen2026wfssb}, fitting~\cite{zhang2026adaq} or resampling~\cite{liu2025bolt} the score are functions of the same signal and cannot add what it lacks: the bias has to be measured, not filtered.

\paragraph{Implications.}
We keep the alignment score as a cheap, dense but biased measurement and correct it with the probe on a few frames.
We use only the label-free yes/no probe, since per-option probabilities can steer the answer: frames selected by the probability of a wrong option make the MLLM choose that option on 24.2\% of questions, against 10.6\% for BLIP-2-selected frames (Supp.~\ref{sec:suppl_steer}).
Diversity also matters: the 32 top-scoring frames give 65.2\%, against 67.0\% for a DPP on the same scores (Supp.~\ref{sec:suppl_topk}).
\section{Method}
\label{sec:method}

\begin{figure*}[t]
    \centering
    \includegraphics[width=\linewidth]{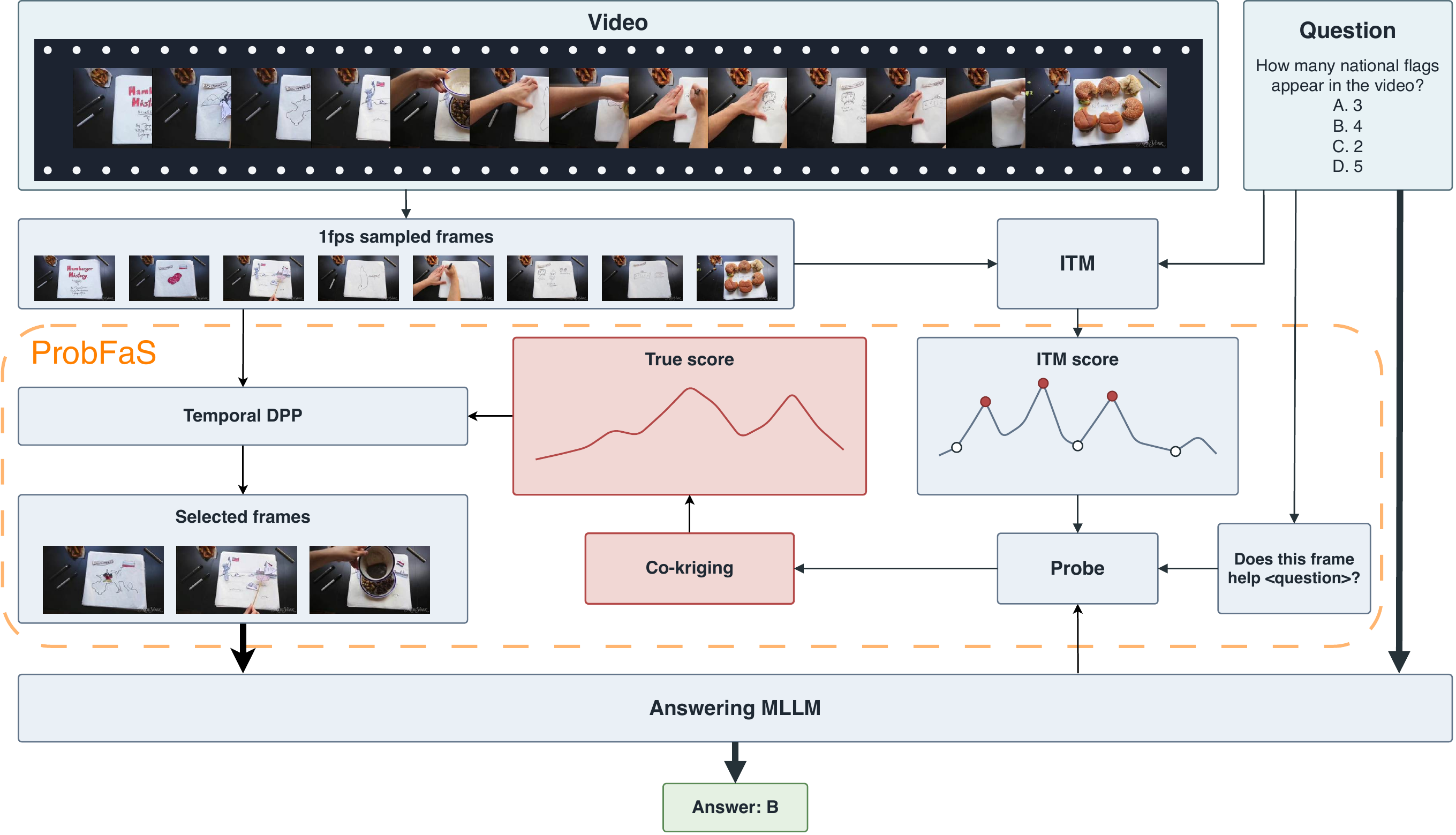}
    \caption{ProbFaS pipeline.
    An image--text alignment model first scores every frame of a one-frame-per-second candidate grid against the question, and then the answering MLLM judges $m$ of these frames with a label-free yes/no question.
    Co-kriging calibrates the low-fidelity alignment scores to these judgments on every frame, and a temporal DPP selects the $k$ frames that the MLLM answers from.}
    \vspace{-0.1in}
    \label{fig:ProbFaS_pipeline}
\end{figure*}

ProbFaS aims at selecting $k$ frames for a frozen MLLM that most effectively support the model in answering a question (\cref{fig:ProbFaS_pipeline}).
We cast selection as a probabilistic surrogate (\cref{sec:formulation}), infer the relevance it needs by fusing cheap alignment scores with a few MLLM probes (\cref{sec:probe,sec:cokrig}), and select frames with a temporal DPP (\cref{sec:dpp}).

\subsection{Problem formulation}
\label{sec:formulation}

Let a video consist of frames $(f_1,\ldots,f_N)$, let $q$ be the question with its options, and let $k\ll N$ be the budget.
Ideally we would choose
\begin{equation}
I^\star=\operatorname*{arg\,max}_{I\subset\{1,\ldots,N\},\,|I|=k}\;P\big(a^\star\mid \{f_i\}_{i\in I},\,q\big),
\label{eq:ideal}
\end{equation}
with $a^\star$ the correct answer and $P$ the probability from the answering MLLM, but $a^\star$ is unknown at test time and the objective is a set function over $\binom{N}{k}$ subsets.
ProbFaS replaces it in two steps.
First, the contribution of a frame is approximated by its label-free relevance $r(t)=\sigma(y(t))$, the probe's probability that the frame helps (\cref{eq:probe}), which tracks usefulness far better than alignment scores (\cref{sec:motivation}).
Measuring $r$ everywhere is too costly, so we infer it as $\hat r(t)=\sigma(g(t))$ from a posterior estimate $g$ of $y$ (\cref{sec:cokrig}).
Second, frame contributions are not additive, since a near-duplicate of a shown frame adds little, so we select by a quality--diversity objective over the $n$ candidate frames (\cref{sec:grid}),
\begin{equation}
I^\star\approx\operatorname*{arg\,max}_{I\subset\{0,\ldots,n-1\},\,|I|=k}\;\sum_{i\in I}\phi\big(\hat r(t_i)\big)+\lambda\,\Omega(I),
\label{eq:surrogate}
\end{equation}
with $\phi$ an increasing function of relevance (\eg identity or log), $\Omega$ a diversity or coverage term, and $\lambda\ge0$ their trade-off.
Many rules are instances that differ only in $\phi$, $\Omega$ and $\lambda$: top-$k$ ($\lambda=0$), non-maximum suppression~\cite{hu2025mllmselector}, relevance plus coverage~\cite{tang2025aks}, maximal marginal relevance~\cite{chen2026wfssb}, log-gap coverage~\cite{huang2026forte} and feature DPPs~\cite{sun2025mdp3}.
ProbFaS solves this by a temporal DPP (\cref{sec:dpp}), in which one probability model fixes $\phi$, $\Omega$ and $\lambda$ and diversity is measured in time.

\subsection{Candidate grid and probes}
\label{sec:grid}
\label{sec:probe}

Like most selectors, we keep one candidate frame per second: the grid $\mathcal{G}$ takes every $\eta$-th native frame, $\eta=\max(\lfloor\mathrm{fps}\rfloor,1)$, and its $n$ frames are indexed by their time $t\in\{0,\ldots,n-1\}$ in seconds.
An image--text alignment model scores each grid frame against $q$, $s(t)=P(\mathrm{match}\mid f_t,q)\in(0,1)$, a probability from an image--text matching (ITM) head or a normalized cosine similarity between image and text embeddings.

These scores are cheap but biased, so ProbFaS observes the bias with the answering MLLM itself: on a probe set $\mathcal{S}$ of at most $m$ grid frames it reads the yes/no log-odds $y(t)$ of \cref{eq:probe}, in batched single-frame forward passes with no generation and no label.
The probes must reveal how the score errs both where it is high and where it is low.
Half of $\mathcal{S}$ are therefore $m/2$ diverse peaks of $s(t)$, and the other half are $m/2$ evenly spaced frames.
We use $m=64$.

\subsection{Co-kriging calibration}
\label{sec:cokrig}

We treat $\mathrm{logit}\,s(t)$ as a low-fidelity and $y(t)$ as a high-fidelity measurement of the same latent relevance, both in log-odds, and relate them by autoregressive co-kriging~\cite{kennedy2000cokriging}:
\begin{equation}
y(t)=\rho\,\mathrm{logit}\,s(t)+b+\delta(t),
\qquad
\delta\sim\mathcal{GP}(0,\kappa_\theta).
\label{eq:cokrig}
\end{equation}
The scale $\rho$ and offset $b$ calibrate the alignment score to the answering model. As in standard probability calibration, the logit undoes the saturation of $s$ near 0 and 1, and the discrepancy $\delta$ is a smooth function of time that absorbs the bias missed by constant calibration.
We fit $\rho$ and $b$ by least squares on $\mathcal{S}$, with $\rho$ free in sign so a high alignment score can also count as evidence against a frame.
The residuals $\tilde{\boldsymbol\delta}_\mathcal{S}$ at the probes are noisy observations of $\delta$, modeled with a Mat\'ern kernel $\kappa_\theta$ in $t$ and white noise $\sigma_\varepsilon^2$ fitted by marginal likelihood (Supp.~\ref{sec:suppl_impl} and \ref{sec:suppl_kernel}).
Adding the GP posterior mean $\hat\delta$ gives the calibrated log-odds on every grid frame,
\begin{equation}
g(t)=\rho\,\mathrm{logit}\,s(t)+b+\underbrace{\kappa_\theta(t,\mathcal{S})\big(\mathbf{K}_\mathcal{S}+\sigma_\varepsilon^2\mathbf{I}\big)^{-1}\tilde{\boldsymbol\delta}_\mathcal{S}}_{\hat\delta(t)},
\end{equation}
with $\mathbf{K}_\mathcal{S}=\kappa_\theta(\mathcal{S},\mathcal{S})$, and the relevance $\hat r(t)=\sigma(g(t))$.
Only the discrepancy is smoothed, so a sharp peak of $s$ stays a sharp peak of $g$, since smoothing $s$ itself would blur the short events that questions often ask about.
At the probes, $g$ stays close to $y$.
Between them, it keeps the shape of $\rho\,\mathrm{logit}\,s+b$, shifted by $\hat\delta$, which interpolates nearby residuals and fades to zero far from the probes.
Under the model, $g$ is the best linear unbiased predictor of the noise-free $y$, with a variance that grows away from the probes (Supp.~\ref{sec:suppl_cokrig}, which also checks $g$ against probing every frame).

\begin{algorithm}[t]
\caption{ProbFaS for one video--question pair.}
\label{alg:probfas}
\scriptsize
\begin{algorithmic}[1]
\Require frames $f_{1:N}$ at frame rate fps; question $q$ with options; frozen MLLM $\mathcal{M}$; alignment model; budgets $k$, $m$
\Ensure $k$ frame indices in temporal order
\Statex \textit{Alignment scores on a 1\,fps grid} \Comment{\cref{sec:grid}}
\State $\eta\gets\max(\lfloor\mathrm{fps}\rfloor,1)$;\; $\mathcal{G}\gets(0,\eta,2\eta,\ldots)$;\; $n\gets|\mathcal{G}|$
\State \textbf{if} $n\le k$ \textbf{then return} $\mathcal{G}$
\State $s(t)\gets P(\mathrm{match}\mid f_{\mathcal{G}_t},q)$ for $t=0,\ldots,n-1$
\Statex \textit{Label-free probes of the answering MLLM} \Comment{\cref{sec:probe}}
\State $\mathcal{S}\gets\textsc{TDPP}(s,\,m/2)\cup\textsc{Even}(n,\,m/2)$ \Comment{score peaks and coverage}
\Statex $\triangleright$ \textsc{Even}$(n,c)$: $c$ evenly spaced indices of $\{0,\ldots,n-1\}$
\State $y(t)\gets\ell_{\mathrm{Yes}}(t)-\ell_{\mathrm{No}}(t)$ of $\mathcal{M}(f_{\mathcal{G}_t},q)$ for $t\in\mathcal{S}$ \Comment{batched, \cref{eq:probe}}
\Statex \textit{Co-kriging calibration} \Comment{\cref{sec:cokrig}}
\State $x(t)\gets\mathrm{logit}\,s(t)$ for $t=0,\ldots,n-1$
\State $(\rho,b)\gets\arg\min_{\rho,b}\sum_{t\in\mathcal{S}}\big(y(t)-\rho\,x(t)-b\big)^2$
\State $\tilde\delta(t)\gets y(t)-\rho\,x(t)-b$ for $t\in\mathcal{S}$ \Comment{probe residuals}
\State $(\theta,\sigma_\varepsilon^2)\gets$ maximize the GP marginal likelihood of $\tilde{\boldsymbol\delta}_\mathcal{S}$ \Comment{Mat\'ern kernel}
\State $\hat\delta(t)\gets\kappa_\theta(t,\mathcal{S})\big(\mathbf{K}_\mathcal{S}+\sigma_\varepsilon^2\mathbf{I}\big)^{-1}\tilde{\boldsymbol\delta}_\mathcal{S}$ for $t=0,\ldots,n-1$
\State $\hat r(t)\gets\sigma\big(\rho\,x(t)+b+\hat\delta(t)\big)$ \Comment{calibrated relevance}
\Statex \textit{Temporal DPP selection} \Comment{\cref{sec:dpp}}
\State $F\gets\{\mathcal{G}_t: t\in\textsc{TDPP}(\hat r,\,k)\}$ \Comment{native frame indices}
\State \textbf{if} $|F|<k$ \textbf{then} add $k-|F|$ frames spread over the gaps of $F$ in proportion to their length
\State \Return $F$ in temporal order
\\
\Function{TDPP}{$w,\,k'$} \Comment{greedy MAP of \cref{eq:dpp}}
  \State $\ell\gets\max(n/2k',1)$;\; $C_{ij}\gets e^{-(i-j)^2/2\ell^2}$;\; $I\gets\emptyset$
  \While{$|I|<k'$}
    \State $\Delta_i\gets w_i^2\big(1-C_{iI}C_I^{-1}C_{Ii}\big)$ for $i\notin I$ \Comment{relevance $\times$ novelty}
    \State \textbf{if} $\max_i\Delta_i\le0$ \textbf{then break}
    \State $I\gets I\cup\{\arg\max_i\Delta_i\}$
  \EndWhile
  \State \Return $I$
\EndFunction
\end{algorithmic}
\end{algorithm}

\subsection{Temporal DPP}
\label{sec:dpp}

A DPP~\cite{kulesza2012dpp} assigns a probability for a subset $I$ proportional to $\det L_I$ for a positive semidefinite kernel $L$ of item qualities and similarities.
Its most probable $k$-subset favors high-quality items that are dissimilar to each other, since $\det L_I$ grows with their qualities and shrinks with their similarity.
ProbFaS uses the calibrated relevance as quality and defines the similarity by an RBF kernel in time,
\begin{equation}
L_{ij}=\hat r(t_i)\,\hat r(t_j)\,C_{ij},\quad
C_{ij}=e^{-(t_i-t_j)^2/2\ell^2},
\label{eq:dpp}
\end{equation}
with $\ell=\max(n/(2k),1)$, half the spacing of $k$ evenly spaced frames.
Since $\log\det L_I=\sum_{i\in I}\log\hat r(t_i)^2+\log\det C_I$, its most probable $k$-subset is an instance of \cref{eq:surrogate} with $\phi=2\log$, $\Omega=\log\det C_I$ and $\lambda=1$, with no hand-set trade-off weight.
We find the $k$-subset by greedy MAP inference~\cite{chen2018fastdpp}, which repeatedly adds the frame with the largest gain
\begin{equation}
\Delta(i\mid I)=\hat r(t_i)^2\big(1-C_{iI}C_I^{-1}C_{Ii}\big),
\label{eq:gain}
\end{equation}
the product of its relevance and its temporal novelty given the chosen frames $I$, which is near zero within about $\ell$ of a chosen frame and near one far from all of them.
The greedy thus takes the relevance peaks first and spreads the remaining budget over uncovered parts of the video.
If it stops early because the gains vanish, or the grid has fewer than $k$ frames, the missing frames are spread over the gaps between the chosen ones.

\Cref{alg:probfas} gives the full pipeline. Only the DPP step depends on $k$: the grid, probes and calibration are computed once per question.

\section{Experiments}
\label{sec:experiments}

\subsection{Setup}
\label{sec:setup}

\paragraph{Benchmarks and protocol.}
Video-MME~\cite{fu2025videomme} has 2,700 questions on 900 short (under 2\,min), medium (4--15\,min) and long (30--60\,min) videos; we use it without subtitles.
MLVU~\cite{zhou2025mlvu} dev has 2,174 questions on 1,122 videos of 3\,min to over 2\,h in seven tasks; we report the official score, the mean of the task accuracies.
LongVideoBench~\cite{wu2024longvideobench} val has 1,337 questions on videos of 8\,s to 1\,h; we use its video-only variant.
All methods answer with the same frozen Qwen3-VL-8B-Instruct~\cite{bai2025qwen3vl} in LMMs-Eval~\cite{zhang2025lmmseval} with greedy decoding, at budgets $k\in\{8,16,32,64,128\}$; every frame is resized to at most 786,432 pixels, and the selected frames are passed as an ordered image sequence.
Two methods are compared on the same questions by the paired McNemar statistic~\cite{mcnemar1947} $z=(n_{10}-n_{01})/\sqrt{n_{10}+n_{01}}$, where $n_{10}$ and $n_{01}$ count the questions that only the first or only the second answers correctly; $|z|>1.96$ means $p<0.05$.

\paragraph{Baselines.}
We compare with uniform sampling and fourteen training-free selectors, BOLT~\cite{liu2025bolt}, AKS~\cite{tang2025aks}, AdaQ~\cite{zhang2026adaq}, QCA~\cite{peng2026qca}, MDP$^3$~\cite{sun2025mdp3}, FOCUS~\cite{zhu2026focus}, DIG~\cite{li2026dig}, A.I.R.~\cite{zou2026air}, LENS~\cite{zhang2026lens}, GIFT~\cite{ma2026gift}, KFC (Nar-KFC~\cite{fang2026narkfc} without its narratives), WFS-SB~\cite{chen2026wfssb}, FRAG~\cite{huang2025frag}, which scores 256 frames with the answering MLLM, and OTFS~\cite{zhao2026otfs}, which combines alignment scores with DIG's MLLM rewards by optimal transport, each run with its released code and defaults (Supp.~\ref{sec:suppl_baselines}).
LENS masks regions and tiles four neighboring frames into hyperframes, so we run its pipeline end to end with the same answering model (Supp.~\ref{sec:suppl_lens}).
Q-Frame~\cite{zhang2025qframe} mixes three resolutions and is compared at a matched token budget.

\paragraph{ProbFaS.}
We use BLIP-2-ITM~\cite{li2023blip2} (ViT-g) to score the frame grid with the question and options as text, and Qwen3-VL-8B as the answering MLLM, which is also the probe model with budget $m=64$ (Supp.~\ref{sec:suppl_impl}).
Nothing is tuned per benchmark or budget; ablations use $k=32$.
Our code and the selected frame indices will be released.

\begin{table*}[t]
  \centering
  \caption{Accuracy (\%) with Qwen3-VL-8B at frame budget $k$. Best in \textbf{bold}, second \underline{underlined}. $^\dagger$Empty selections for $k<32$. $^\ddagger$Own pipeline; some images tile four frames.}
  \label{tab:main}
  \small
  \setlength{\tabcolsep}{3.2pt}
  \begin{tabular}{l ccccc ccccc ccccc}
    \toprule
    & \multicolumn{5}{c}{Video-MME} & \multicolumn{5}{c}{MLVU} & \multicolumn{5}{c}{LongVideoBench} \\
    \cmidrule(lr){2-6}\cmidrule(lr){7-11}\cmidrule(lr){12-16}
    Method $\backslash$ $k$ & 8 & 16 & 32 & 64 & 128 & 8 & 16 & 32 & 64 & 128 & 8 & 16 & 32 & 64 & 128 \\
    \midrule
Uniform & 56.3 & 60.2 & 64.4 & 67.0 & 69.1 & 55.2 & 59.3 & 63.7 & 68.1 & 73.1 & 55.1 & 57.9 & 60.4 & 63.2 & 64.9 \\
BOLT~\cite{liu2025bolt} & 58.3 & 61.6 & 64.7 & 67.1 & 69.1 & 63.4 & 67.6 & 70.5 & 74.1 & 76.0 & 58.2 & 60.4 & 62.8 & 64.8 & 65.7 \\
FRAG~\cite{huang2025frag} & 60.3 & 62.1 & 64.8 & 66.7 & 69.3 & 66.4 & 69.2 & 72.2 & 72.5 & 75.0 & 61.7 & 62.7 & 63.4 & 65.1 & 66.0 \\
AKS$^\dagger$~\cite{tang2025aks} & -- & -- & 65.6 & 67.7 & 69.6 & -- & -- & 68.1 & 72.2 & 74.0 & -- & -- & 61.5 & 64.8 & 64.9 \\
AdaQ~\cite{zhang2026adaq} & 56.9 & 61.4 & 64.1 & 67.3 & 68.7 & 65.7 & 68.7 & 72.5 & 75.9 & \underline{77.2} & 58.9 & 60.9 & 63.4 & 64.9 & 65.4 \\
QCA~\cite{peng2026qca} & 60.0 & 61.6 & 65.6 & 67.9 & 69.6 & 65.2 & 67.0 & 72.5 & 75.9 & 75.8 & 59.2 & 61.2 & 65.4 & 66.3 & 66.7 \\
MDP$^3$~\cite{sun2025mdp3} & 59.8 & 62.8 & 66.1 & 68.4 & 69.4 & 66.0 & 68.1 & 71.5 & 72.1 & 73.3 & 59.2 & 61.0 & 63.3 & 63.6 & 63.9 \\
FOCUS~\cite{zhu2026focus} & 59.2 & 60.6 & 62.9 & 64.9 & 64.9 & 66.2 & 69.5 & 72.7 & 72.9 & 73.7 & 59.9 & 61.9 & 64.5 & 64.9 & 66.1 \\
DIG~\cite{li2026dig} & 59.3 & 62.2 & 65.1 & 67.8 & 68.9 & 62.5 & 65.7 & 68.8 & 70.9 & 72.4 & 59.8 & 62.1 & 64.1 & 66.7 & 67.2 \\
A.I.R.~\cite{zou2026air} & 60.6 & 62.7 & 63.0 & 63.8 & 64.1 & 67.8 & 70.5 & 71.5 & 71.9 & 71.5 & 62.2 & 63.6 & 62.8 & 63.4 & 63.6 \\
LENS$^\ddagger$~\cite{zhang2026lens} & \underline{62.4} & \underline{65.4} & \underline{67.0} & \underline{69.0} & \textbf{70.8} & 66.5 & 66.9 & 69.2 & 70.9 & 73.2 & 60.0 & 64.2 & 63.1 & 65.4 & 65.4 \\
GIFT~\cite{ma2026gift} & 60.9 & 63.9 & 66.4 & 68.5 & 69.0 & 66.6 & 70.9 & 72.4 & 73.2 & 72.9 & 59.0 & 61.8 & 63.9 & 66.0 & 65.3 \\
KFC~\cite{fang2026narkfc} & 59.0 & 61.8 & 64.4 & 67.4 & 69.3 & 65.8 & 71.7 & 73.3 & 74.6 & 76.3 & 56.7 & 60.8 & 64.1 & 66.6 & 66.1 \\
WFS-SB~\cite{chen2026wfssb} & 61.4 & 63.8 & 66.2 & 68.1 & 69.7 & 66.9 & 70.6 & \underline{75.0} & 75.7 & 76.8 & 62.9 & 64.3 & \underline{66.9} & \underline{67.9} & \textbf{68.7} \\
OTFS~\cite{zhao2026otfs} & 61.1 & 63.6 & 66.7 & 68.7 & \underline{70.4} & \underline{69.3} & \underline{72.1} & 74.2 & \underline{76.3} & 76.7 & \underline{63.6} & \underline{64.4} & 65.4 & 65.1 & 67.5 \\
\midrule
\textbf{ProbFaS (ours)} & \textbf{63.9} & \textbf{65.7} & \textbf{67.9} & \textbf{69.1} & \underline{70.4} & \textbf{69.9} & \textbf{72.6} & \textbf{77.2} & \textbf{78.1} & \textbf{78.4} & \textbf{64.8} & \textbf{67.5} & \textbf{69.2} & \textbf{69.0} & \underline{68.6} \\
    \bottomrule
  \end{tabular}
\end{table*}

\subsection{Main results}
\label{sec:main_results}

ProbFaS is the most accurate method in 13 of the 15 settings of \cref{tab:main} (MLVU in \cref{fig:quantitative}), and a close second in the other two.
Over the second-best method, ProbFaS gains 1.5, 0.6 and 1.2 points at $k=8$ and 0.9, 2.2 and 2.3 points at $k=32$ on Video-MME, MLVU and LongVideoBench, and over uniform sampling 7.6, 14.7 and 9.7 points at $k=8$, where choosing the right frames matters most.
At $k=32$ the margins over the runner-up WFS-SB on MLVU and LongVideoBench are significant ($z=3.3$ and $2.3$).
On Video-MME the runner-up for $k\le32$ is LENS, whose hyperframes show up to four frames per image, yet it trails ProbFaS by 3 to 8 points on MLVU and LongVideoBench ($z\ge2.8$).
At $k=128$, LENS on Video-MME and WFS-SB on LongVideoBench are slightly but not significantly ahead, as with many frames the choice matters less.
Averaged over the 15 settings, ProbFaS leads the next best by 1.8 points, at less cost than any MLLM-based selector (\cref{tab:cost}).

\begin{wraptable}[14]{r}{0.46\textwidth}
  \centering
  \caption{ProbFaS at small budgets against larger ones. Best baseline: best of the fourteen selectors at $k=128$. Q-Frame shows 4, 8 and 32 frames at $\frac{1}{2}$, $\tfrac14$ and $\tfrac18$ resolution, about the visual tokens of 8 full frames.}
  \label{tab:efficiency_frames}
  \small
  \setlength{\tabcolsep}{4pt}
  \begin{tabular}{lrccc}
    \toprule
    Method & Frames & Video-MME & MLVU & LVB \\
    \midrule
    ProbFaS & 8 & 63.9 & 69.9 & 64.8 \\
    Q-Frame~\cite{zhang2025qframe} & 4+8+32 & 62.7 & 68.0 & 62.1 \\
    \midrule
    ProbFaS & 32 & 67.9 & \textbf{77.2} & \textbf{69.2} \\
    Best baseline & 128 & 70.8 & \textbf{77.2} & 68.7 \\
    Uniform & 128 & 69.1 & 73.1 & 64.9 \\
    Uniform & 256 & 70.6 & 75.8 & 65.7 \\
    Uniform, 2\,fps & $\le$768 & \textbf{72.1} & 76.9 & 66.6 \\
    \bottomrule
  \end{tabular}
\end{wraptable}

\paragraph{Fewer frames for better accuracy.}
On MLVU and LongVideoBench, 32 ProbFaS frames match or exceed the best baseline at 128 frames and uniform sampling at 2\,fps with up to 768 frames (\cref{tab:efficiency_frames}).
On Video-MME, where many questions concern the whole video (synopsis, counting, order), accuracy keeps rising with the number of frames, and 32 ProbFaS frames match about 90 uniform ones.
Compared with Q-Frame at a similar token budget, 8 ProbFaS frames are more accurate on all three benchmarks.

\subsection{Other answering models}
\label{sec:other_mllm}

\begin{table*}[t]
  \centering
  \begin{minipage}[t]{0.495\textwidth}
    \centering
    \caption{Accuracy (\%) at $k=32$ with other answering models, replaying the frame lists of \cref{tab:main}; own probe: ProbFaS probed by the answering model. $^\dagger$Reported by DAFS~\cite{wang2026dafs} in its own setup; not ranked.}
    \label{tab:other_mllm}
    \small
    \setlength{\tabcolsep}{2.6pt}
    \resizebox{\linewidth}{!}{%
    \begin{tabular}{l ccc ccc}
      \toprule
      & \multicolumn{3}{c}{LLaVA-Video-7B} & \multicolumn{3}{c}{InternVL3.5-8B} \\
      \cmidrule(lr){2-4}\cmidrule(lr){5-7}
      Method & V-MME & MLVU & LVB & V-MME & MLVU & LVB \\
      \midrule
      Uniform & 62.5 & 65.0 & 58.5 & 63.6 & 67.4 & 60.2 \\
      BOLT~\cite{liu2025bolt} & 64.1 & 69.5 & 60.1 & 65.0 & 69.9 & 61.1 \\
      AKS~\cite{tang2025aks} & 64.4 & 67.6 & 59.8 & 64.7 & 69.9 & 62.1 \\
      AdaQ~\cite{zhang2026adaq} & 63.0 & 70.5 & 60.4 & 64.4 & 70.4 & 61.9 \\
      QCA~\cite{peng2026qca} & 63.5 & 69.5 & 60.6 & 64.9 & 70.9 & 63.6 \\
      MDP$^3$~\cite{sun2025mdp3} & 64.9 & 71.0 & 60.2 & 65.1 & 70.7 & 62.7 \\
      FOCUS~\cite{zhu2026focus} & 61.0 & 69.9 & 60.8 & 62.6 & 70.2 & 62.5 \\
      DIG~\cite{li2026dig} & 64.1 & 68.3 & \underline{62.0} & 64.1 & 69.3 & 63.1 \\
      GIFT~\cite{ma2026gift} & \underline{65.0} & 71.1 & 59.3 & 65.3 & 71.8 & 62.0 \\
      WFS-SB~\cite{chen2026wfssb} & 64.8 & \underline{71.2} & \textbf{63.1} & 65.4 & 71.5 & \textbf{64.8} \\
      \midrule
      ProbFaS (Qwen probe) & \textbf{65.6} & \textbf{72.3} & \textbf{63.1} & \underline{65.8} & \textbf{73.3} & \underline{64.2} \\
      ProbFaS (own probe) & 64.7 & 69.4 & 61.0 & \textbf{66.1} & \underline{72.7} & 63.2 \\
      \midrule
      Uniform$^\dagger$ & 62.1 & 65.8 & 57.4 & 64.0 & 68.6 & 62.9 \\
      DAFS$^\dagger$~\cite{wang2026dafs} & 65.2 & 68.5 & 60.1 & 67.0 & 71.3 & 65.8 \\
      \bottomrule
    \end{tabular}
    }
  \end{minipage}\hfill
  \begin{minipage}[t]{0.475\textwidth}
    \centering
    \caption{Ablations at $k=32$; each row changes one component of ProbFaS.}
    \vspace{0.07in}
    \label{tab:ablation}
    \footnotesize
    \setlength{\tabcolsep}{3pt}
    \resizebox{\linewidth}{!}{%
    \begin{tabular}{lccc}
      \toprule
      Variant & V-MME & MLVU & LVB \\
      \midrule
      Uniform sampling & 64.4 & 63.7 & 60.4 \\
      BLIP-2 only: DPP, no probe & 67.0 & 72.6 & 65.2 \\
      Probes only: DPP on the 64 probes, no GP & 67.9 & 75.1 & 66.5 \\
      \textbf{ProbFaS} & \textbf{67.9} & \textbf{77.2} & \textbf{69.2} \\
      \midrule
      \multicolumn{4}{l}{\emph{Probe placement}} \\
      \quad 64 evenly spaced & 66.8 & 75.9 & 66.2 \\
      \quad 128 mixed (64 DPP + 64 uniform) & 67.4 & 76.8 & 67.2 \\
      \quad every 1\,fps frame & 68.2 & 78.0 & 68.2 \\
      \multicolumn{4}{l}{\emph{Alignment and probe model}} \\
      \quad SigLIP 2 instead of BLIP-2 & 67.0 & 76.5 & 68.4 \\
      \quad Qwen3-VL-2B probe instead of 8B & 67.0 & 72.7 & 65.0 \\
      \multicolumn{4}{l}{\emph{Selection rule on $\hat r$ instead of the DPP}} \\
      \quad Top-$k$ & 66.8 & 76.4 & 69.2 \\
      \quad Inverse transform sampling (BOLT) & 66.1 & 72.3 & 68.5 \\
      \quad Judge-and-split (AKS) & 67.0 & 70.2 & 65.3 \\
      \quad Softmax top-$p$ sampling (AdaQ) & 67.0 & 71.9 & 67.1 \\
      \quad Wavelet + segments (WFS-SB) & 66.9 & 73.5 & 67.8 \\
      \quad Exact relevance + log coverage (FORTE) & 68.0 & 73.6 & 67.2 \\
      \bottomrule
    \end{tabular}
    }
  \end{minipage}
\end{table*}

We replay the frame lists of \cref{tab:main} with LLaVA-Video-7B~\cite{zhang2024llavavideo} and InternVL3.5-8B~\cite{wang2025internvl35} at $k=32$, and also run ProbFaS with the answering model as its own probe (Supp.~\ref{sec:suppl_other_mllm}).
A.I.R.\ and LENS run Qwen3-VL inside their pipelines and are not replayed.
With the Qwen3-VL-8B probe, ProbFaS is the most accurate or tied in 8 of the 9 model--benchmark pairs over the three answering models (\cref{tab:main,tab:other_mllm}) and trails only WFS-SB with InternVL on LongVideoBench, by 0.6 points.
Its margins over the runner-up shrink with the new models ($|z|\le1.7$), while those over uniform sampling stay large ($|z|\ge2.8$): frames chosen from one model's judgments serve models that were never probed.
As its own probe, InternVL3.5-8B gives the best Video-MME result (66.1), whereas LLaVA-Video-7B, whose yes/no margins span 8--12 times less, loses 0.9--2.9 points but stays above uniform sampling.
Over its own uniform baseline, DAFS~\cite{wang2026dafs} gains less than ProbFaS on MLVU and LongVideoBench and about as much on Video-MME (Supp.~\ref{sec:suppl_other_mllm}).

\subsection{Ablations}
\label{sec:ablation}

\paragraph{Components.}
The alignment score alone, with the same temporal DPP, already beats uniform sampling, and the probes with co-kriging add 0.9, 4.6 and 4.0 points (\cref{tab:ablation}).
The GP extrapolation is essential: a DPP on the 64 probed frames alone ties ProbFaS on Video-MME but loses 2.1 and 2.7 points on MLVU and LongVideoBench ($z=-3.3$ and $-3.1$), where the frames that matter often lie between probes.
Probing 64 evenly spaced frames loses on all three benchmarks ($z=-1.8$, $-2.7$ and $-3.7$), as the bias of the alignment score at its own peaks goes unobserved, and 128 mixed probes do not help.
SigLIP~2~\cite{tschannen2025siglip2} instead of BLIP-2 costs 0.7--0.9 points but still beats every baseline on MLVU and LongVideoBench.
A Qwen3-VL-2B probe falls to the BLIP-2-only level: it ranks frames like the 8B model (median Spearman ${\approx}0.7$), but its margins span three times less, which flattens $\hat r$ (Supp.~\ref{sec:suppl_sensitivity}).

\paragraph{Selection rule.}
We feed the same $\hat r$ to top-$k$, to the rules of four baselines and to the exact relevance-plus-log-coverage optimum of FORTE~\cite{huang2026forte} (Supp.~\ref{sec:suppl_plugin}).
No rule is significantly better than the temporal DPP, which is best or within 0.1 points in all 18 comparisons and significantly better in 9.
Top-$k$ comes closest, while coverage rules spread frames into stretches that $\hat r$ rates irrelevant and lose up to 7.0 points on MLVU ($z=-8.1$).
Most of the gain thus comes from the calibration, which also lifts BOLT and AKS over their own scores on all three benchmarks, by 1.4--5.7 points.

\begin{wraptable}{r}{0.46\textwidth}
  \centering
  \caption{Probe budget $m$ against frame budget $k$ on Video-MME. Every frame: the whole 1\,fps grid is probed (about 16$\times$ the cost of $m=64$).}
  \label{tab:probe_budget}
  \small
  \setlength{\tabcolsep}{4.5pt}
  \begin{tabular}{lccccc}
    \toprule
    Probes $m$ $\backslash$ $k$ & 8 & 16 & 32 & 64 & 128 \\
    \midrule
    32 & 62.1 & 64.8 & 67.0 & 68.7 & \underline{70.5} \\
    64 (default) & \underline{63.9} & \underline{65.7} & \underline{67.9} & \underline{69.1} & 70.4 \\
    128 & \underline{63.9} & 65.1 & 67.4 & 69.0 & 70.1 \\
    Every frame & \textbf{64.3} & \textbf{65.9} & \textbf{68.2} & \textbf{69.8} & \textbf{71.2} \\
    \bottomrule
  \end{tabular}
\end{wraptable}

\paragraph{Probe budget.}
With $m=32$, accuracy drops by 0.9--1.8 points at $k\le32$ (paired $z=-3.3$ pooled), since with few frames the probes have to pin down the right moments, whereas $m=128$ never helps significantly (\cref{tab:probe_budget}).
At $k\le32$, $m=64$ is second only to probing every frame, which costs 16 times more and adds only 0.2--0.4 points, so co-kriging from 64 probes recovers nearly all that single-frame judgments offer.
At $k\ge64$ the sparse budgets are within half a point, as coverage matters more than a precise ranking.
\cref{tab:main} therefore uses $m=64$ as the default setup.
Other ways to spend the probes, and settings tuned on a slice of Video-MME, did not further help (Supp.~\ref{sec:suppl_sensitivity}, \ref{sec:suppl_negative}).

\subsection{Analysis}
\label{sec:analysis}

\begin{wraptable}[23]{r}{0.46\textwidth}
  \centering
  \caption{Selection time per question (s) on one GH200 GPU, for 90 Video-MME questions on 30 videos (mean length 1.2, 8.5 and 40.1\,min), model loading excluded. The answer pass adds 2.0\,s at $k=32$ and 6.5\,s at $k=128$.
  $^\ast$Whole run divided by the number of questions.
  $^\dagger$Without its 80B query router.
  $^\ddagger$Same BLIP-2 grid as ProbFaS.
  $^\S$From its full Video-MME run (Supp.~\ref{sec:suppl_cost}).}
  \label{tab:cost}
  \small
  \setlength{\tabcolsep}{4pt}
  \begin{tabular}{lcccc}
    \toprule
    Method & Short & Medium & Long & All \\
    \midrule
    BOLT~\cite{liu2025bolt} & 0.5 & 2.2 & 9.9 & 4.2 \\
    MDP$^3$~\cite{sun2025mdp3} & 5.6 & 6.2 & 6.4 & 6.1 \\
    WFS-SB$^\ddagger$~\cite{chen2026wfssb} & 0.5 & 3.8 & 19.5 & 8.0 \\
    AdaQ~\cite{zhang2026adaq} & 0.9 & 5.6 & 26.0 & 10.8 \\
    FRAG~\cite{huang2025frag} & 13.0 & 12.3 & 12.5 & 12.6 \\
    FOCUS$^\ast$~\cite{zhu2026focus} & -- & -- & -- & 14.7 \\
    LENS$^\S$~\cite{zhang2026lens} & 16.0 & 29.7 & 86.5 & 44.1 \\
    QCA~\cite{peng2026qca} & 4.7 & 23.7 & 114.4 & 47.6 \\
    AKS~\cite{tang2025aks} & 3.9 & 24.9 & 119.9 & 49.5 \\
    A.I.R.$^\ast$~\cite{zou2026air} & -- & -- & -- & $\approx$75 \\
    DIG$^\dagger$~\cite{li2026dig} & 25.2 & 51.5 & 178.7 & 85.1 \\
    OTFS$^\dagger$~\cite{zhao2026otfs} & 29.3 & 78.4 & 308.4 & 138.7 \\
    \midrule
    \textbf{ProbFaS} & 3.3 & 7.6 & 23.7 & 11.5 \\
    \quad BLIP-2 grid & 0.5 & 3.8 & 19.5 & 8.0 \\
    \quad 64 probes & 2.7 & 3.7 & 4.0 & 3.5 \\
    \quad Co-kriging + DPP & 0.04 & 0.08 & 0.14 & 0.09 \\
    \bottomrule
  \end{tabular}
\end{wraptable}

\paragraph{Video length and question type.}
At $k=32$ ProbFaS is the most accurate method on the longest videos of MLVU (70.3\% over 20\,min, next best 69.6\%) and LongVideoBench (62.4\% at 15--60\,min, next best 59.8\%), and second on long Video-MME videos (57.2\%) behind LENS (58.8\%).
Its gains concentrate on questions about one or a few moments, such as Video-MME OCR (82.0\% against at most 77.0\%) and MLVU plot QA (83.7\% against at most 81.1\%), and are smallest when many moments or the whole video matter, such as Video-MME counting and synopsis (Supp.~\ref{sec:suppl_detailed}).

\paragraph{Cost.}
ProbFaS takes 11.5\,s per question (\cref{tab:cost}), close to AdaQ and FRAG, about a quarter of LENS, QCA and AKS, a seventh of DIG and a twelfth of OTFS.
The BLIP-2 grid grows linearly with the video length and dominates on long videos (19.5 of 23.7\,s), although BLIP-2 encodes each frame once and reuses it for all questions on the video (16.9\,s per question without this reuse).
The 64 probes take about 3.5\,s whatever the length, and co-kriging plus the DPP take under 0.15\,s.
FRAG, which spends about as long on 256 passes of the answering MLLM, is 3.7 points less accurate on average (\cref{tab:main}).
WFS-SB scores the same grid: ProbFaS takes 3.5\,s more and is 1.7, 2.2 and 2.3 points more accurate at $k=32$ (\cref{tab:main}).

\section{Conclusion}
\label{sec:conclusion}

Our findings show that the image--text alignment scores behind training-free frame selectors correlate poorly with an MLLM's true downstream utility, whereas the model's own yes/no judgment provides a much stronger selection signal.
To leverage both, ProbFaS formulates frame relevance as a multi-fidelity probabilistic process: co-kriging calibrates dense, low-cost relevance scores against sparse label-free judgments from the answering MLLM, and a temporal DPP selects a subset that balances relevance peaks with temporal coverage.
Without training or labels, it achieves the highest accuracy in 13 of 15 settings while keeping computational overhead low. On MLVU and LongVideoBench its 32 selected frames match or surpass the best baseline at 128 frames and uniform sampling with up to 768 frames.

\paragraph{Limitations.}
Our single-token probe evaluates frames independently, so it does not explicitly capture inter-frame compositional dependencies; consequently, ProbFaS yields smaller gains on tasks requiring multi-event synthesis, such as Video-MME counting or global video synopses, and its relative advantage decreases on Video-MME under higher frame budgets.
Additionally, probe evaluation introduces $m$ single-frame forward passes and relies on an MLLM capable of reliable single-frame confidence calibration: performance degrades when paired with smaller or weaker judge models (\eg LLaVA-Video-7B or Qwen3-VL-2B).

{
    \small
    \bibliographystyle{ieeenat_fullname}
    \bibliography{main}
}

\clearpage
\newcounter{arxivpage}
\setcounter{arxivpage}{\value{page}}
\newcommand{\maketitlesupplementary}{%
  \setcounter{page}{\value{arxivpage}}%
  \begin{center}{\LARGE\scshape Appendix}\end{center}%
}
\clearpage
\setcounter{page}{1}
\maketitlesupplementary
\setcounter{section}{0}
\renewcommand{\thesection}{\Alph{section}}
\renewcommand{\theHsection}{suppl.\Alph{section}} 
\addtocontents{toc}{\protect\setcounter{tocdepth}{2}}
\tableofcontents

\section{Motivation analysis}
\label{sec:suppl_motivation}

This section gives the prompts of \cref{sec:motivation} and extends its analysis.
The experiments of \cref{sec:suppl_peak,sec:suppl_steer,sec:suppl_topk} use Video-MME, Qwen3-VL-8B, $k=32$ and BLIP-2-ITM scores on the 1\,fps grid.

\subsection{Prompts}
\label{sec:suppl_prompts}

All MLLM readouts use the chat template of the model with the system message ``You are a helpful assistant.''
The user turn contains one frame followed by the text below; nothing is generated beyond the first token, whose logits are read.
The boxes show the prompts as they are filled in for a Video-MME question.

\begin{promptbox}{Usefulness $u(t)$ (analysis only, \cref{eq:usefulness})}
\small\ttfamily
Question: How many kinds of fruit are shown in the video?\\
Options:\\
{['A. Apples.', 'B. Candles.', 'C. Berries.', 'D. The three kinds are of the same number.']}\\
Answer with the option letter only.
\end{promptbox}
\noindent The softmax is taken over the logits of the letters A--D, and $u(t)$ is the entry of the correct letter.
In this prompt the options are inserted as one list, as above.

\begin{promptbox}{Probe $y(t)$ (\cref{eq:probe}, used by ProbFaS)}
\small\ttfamily
Question: How many kinds of fruit are shown in the video?\\
Options:\\
A. Apples.\\
B. Candles.\\
C. Berries.\\
D. The three kinds are of the same number.\\
Does this video frame show information that helps answer the question? Answer Yes or No.
\end{promptbox}
\noindent $y(t)$ is the logit of the token ``Yes'' minus that of ``No''.
For benchmarks without options the question is followed directly by the last sentence.

\begin{promptbox}{Text scored by BLIP-2 and SigLIP}
\small\ttfamily
How many kinds of fruit are shown in the video?\\
A. Apples.\\
B. Candles.\\
C. Berries.\\
D. The three kinds are of the same number.
\end{promptbox}
\noindent CLIP-B/32 in \cref{tab:score_usefulness} receives the same text, truncated to its 77-token limit.

\subsection{Usefulness by score bin and on MLVU and LongVideoBench}
\label{sec:suppl_bins}

\begin{figure*}[h]
  \centering
  \includegraphics[width=0.84\textwidth]{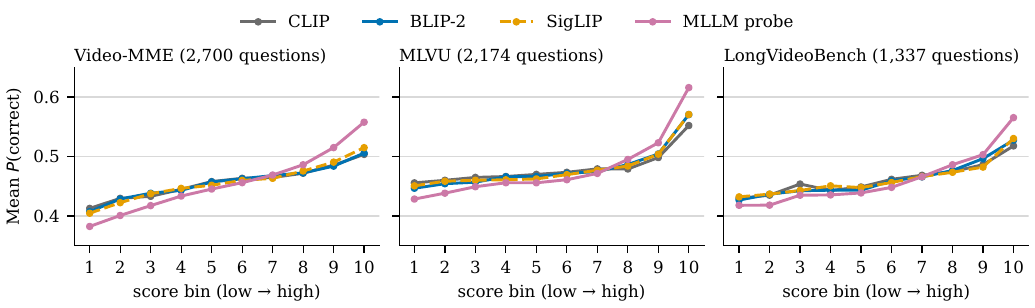}
  \caption{Mean usefulness $u(t)$, the probability of the correct option given one frame, by within-question score bin on Video-MME, MLVU and LongVideoBench (Qwen3-VL-8B).
  Bin 1 holds the lowest-scored tenth of each question's probed frames and bin 10 the highest.
  The vertical axis is restricted to $[0.35, 0.65]$.}
  \label{fig:score_bins}
\end{figure*}

\Cref{fig:score_bins} extends the right panel of \cref{fig:score_usefulness} from the top tenth of each score to all ten, and from Video-MME to MLVU and LongVideoBench.
For each question and score, the probed frames are ranked by the score and split into ten bins of equal size, from the lowest-scored tenth (bin 1) to the highest (bin 10).
The mean usefulness $u(t)$ (\cref{eq:usefulness}) of each bin is then averaged over questions.
A score that ranked frames as the MLLM uses them would rise steeply from bin 1 to bin 10.

On MLVU and LongVideoBench we repeat the analysis of \cref{sec:motivation} with the same answering model, scorers and prompts, on the frames that ProbFaS probes (62 and 51 per question on average, against 58 on Video-MME).
CLIP-B/32 and SigLIP-SO400M score these frames with the question and options, and BLIP-2 is read from its 1\,fps grid, as in ProbFaS.

\begin{table}[t]
  \centering
  \caption{Agreement of each score with single-frame usefulness on MLVU (2,174 questions) and LongVideoBench (1,337 questions), Qwen3-VL-8B.
  Columns as in \cref{tab:score_usefulness}; Negative and Top-frame in \%.}
  \label{tab:suppl_usefulness}
  \small
  \setlength{\tabcolsep}{3pt}
  \resizebox{\linewidth}{!}{%
  \begin{tabular}{lcccccc}
    \toprule
    & \multicolumn{3}{c}{MLVU} & \multicolumn{3}{c}{LongVideoBench} \\
    \cmidrule(lr){2-4}\cmidrule(lr){5-7}
    Score & Spearman & Negative & Top-frame & Spearman & Negative & Top-frame \\
    \midrule
    No image      & --   & --   & 42.9 & --   & --   & 41.6 \\
    CLIP-B/32     & 0.08 & 39.3 & 59.7 & 0.06 & 42.9 & 53.4 \\
    BLIP-2-ITM    & 0.13 & 35.3 & 61.0 & 0.07 & 41.2 & 57.6 \\
    SigLIP-SO400M & 0.11 & 37.0 & 61.9 & 0.06 & 42.1 & 56.6 \\
    Yes/no probe  & 0.22 & 31.6 & 67.2 & 0.14 & 38.6 & 63.3 \\
    \bottomrule
  \end{tabular}}
\end{table}

On all three benchmarks the three alignment curves nearly coincide and rise by only 0.09 to 0.12 from bin 1 to bin 10, while the probe rises by 0.15 to 0.19.
The probe's curve crosses them around bin 7: its upper bins hold more useful frames than those of any alignment score, and its lower bins less useful ones.
\Cref{tab:suppl_usefulness} gives the measures of \cref{tab:score_usefulness} on the two benchmarks.
The alignment scores again barely track usefulness.
Their median correlation is 0.08--0.13 on MLVU and 0.06--0.07 on LongVideoBench, and it is negative on 35--39\% and 41--43\% of questions.
Their top frame answers 59.7--61.9\% and 53.4--57.6\% of questions, against 42.9\% and 41.6\% with no image.
The probe is again clearly closer, with correlations of 0.22 and 0.14 and top-frame accuracies of 67.2\% and 63.3\%.
Unlike on Video-MME, BLIP-2 correlates more with usefulness than SigLIP-SO400M on both benchmarks, so a stronger alignment model does not help consistently.
Correlations fall with video length for every score, but the probe stays ahead in every duration group of Video-MME and LongVideoBench: from 0.34 against 0.10--0.18 on the shortest videos to 0.07 and 0.06 against at most 0.04 on the longest.

\subsection{Finding more peaks of the alignment score}
\label{sec:suppl_peak}
With a budget of a fraction $B$ of the grid, a first pass scores every $(2/B)$-th grid frame.
A second pass spends the other half of the budget, in eight batches, on the midpoints of the gaps between scored frames.
A gap is ranked by the expected excess of its maximum score over the 32nd-highest score so far, modeling the scores inside it as a Brownian bridge between its two scored ends, with the variance of that video's first-pass increments; nothing is fitted.
The control scores every $(1/B)$-th grid frame.
Both then select with the temporal DPP of \cref{sec:dpp} on the raw scores.
\Cref{tab:suppl_peak} shows that the peak search finds 1.7 to 2.2 times as many of the full grid's top-scoring frames at every budget, yet never raises accuracy significantly ($|z|\le0.9$).
Scoring fewer frames does cost accuracy (25\%: $z=-2.1$ against the full grid), but finding the score peaks does not recover it.

\begin{table}[h]
  \centering
  \caption{Scoring part of the 1\,fps grid (Video-MME, $k=32$). Recall: fraction of the full grid's 32 top-scoring frames that are scored. $z$: peak search against even scoring at the same budget, paired.}
  \label{tab:suppl_peak}
  \small
  \setlength{\tabcolsep}{3pt}
  \begin{tabular}{llcccccc}
    \toprule
    Scored & Rule & Recall & Acc. & Short & Med. & Long & $z$ \\
    \midrule
    100\% & every frame & 1.00 & 67.0 & 78.6 & 65.8 & 56.6 & -- \\
    \midrule
    50\% & even & 0.51 & 66.8 & 76.3 & 66.4 & 57.6 & \\
         & peak search & 0.86 & 67.1 & 77.8 & 67.7 & 55.9 & $0.6$ \\
    25\% & even & 0.30 & 65.5 & 75.0 & 65.2 & 56.3 & \\
         & peak search & 0.66 & 65.5 & 75.3 & 65.6 & 55.7 & $0.0$ \\
    12.5\% & even & 0.22 & 65.1 & 74.9 & 64.3 & 56.1 & \\
         & peak search & 0.49 & 64.4 & 74.6 & 63.3 & 55.4 & $-0.9$ \\
    \bottomrule
  \end{tabular}
\end{table}

\subsection{Steering by per-option probabilities}
\label{sec:suppl_steer}
On the frames probed in \cref{sec:motivation}, we select 32 with the same temporal DPP and three qualities: the BLIP-2 score, the usefulness $u(t)$, which uses the label, and the single-frame probability $P(w\mid f_t,q)$ of one wrong option $w$ drawn at random per question (\cref{tab:suppl_steer}).
The set built for $w$ makes the model answer $w$ on 24.2\% of questions, 13.6 points more than the BLIP-2 set, while the set built from $u(t)$ raises the correct answers by 8.7 points.
A set built from the probability of an option thus pushes the answer toward that option, right or wrong, and its gain with the label is leakage rather than evidence.
The yes/no probe names no option.

\begin{table}[h]
  \centering
  \caption{Answers after selecting frames by a per-option probability (Video-MME, $k=32$, \% of questions). $w$: a wrong option drawn at random per question.}
  \label{tab:suppl_steer}
  \small
  \setlength{\tabcolsep}{4pt}
  \begin{tabular}{lccc}
    \toprule
    DPP quality & Correct & Answers $w$ & Other wrong \\
    \midrule
    BLIP-2 score & 67.0 & 10.6 & 22.4 \\
    $u(t)=P(a^\star\mid f_t,q)$ & 75.7 & 8.4 & 15.9 \\
    $P(w\mid f_t,q)$ & 50.9 & 24.2 & 24.9 \\
    \bottomrule
  \end{tabular}
\end{table}

\subsection{Top-$k$ against the DPP}
\label{sec:suppl_topk}
On the full grid, the 32 top-scoring frames give 65.2\% (78.0, 62.4 and 55.2\% on short, medium and long videos), against 67.0\% (78.6, 65.8 and 56.6\%) for the temporal DPP on the same scores (paired $z=-2.5$).
Top-$k$ spends its budget on adjacent frames around the same peaks; the loss is largest on medium videos.

\section{Method details}
\label{sec:suppl_method}

\subsection{Implementation details}
\label{sec:suppl_impl}

\paragraph{Grid and alignment scores.}
The grid keeps decoder frames $0,\eta,2\eta,\ldots$ with $\eta=\max(\lfloor\mathrm{fps}\rfloor,1)$.
BLIP-2-ITM is \texttt{Salesforce/blip2-itm-vit-g}; $s(t)$ is the softmax probability of the ``match'' class of its image--text matching head, computed separately for every frame--question pair.
Before the calibration, $s(t)$ is clipped to $[10^{-4},1-10^{-4}]$ so that its logit is finite.

\paragraph{Probe.}
Probe frames are resized to at most 401,408 pixels, the default single-image setting of our probe implementation and about half the 786,432 pixels of the answer pass, which halves the cost of the probe; we did not tune this value.
A single image fills one two-frame temporal patch of Qwen3-VL, so a probe costs about as many visual tokens as one patch of the answer pass at that resolution.
Probes are batched 16 at a time.
The probe set is the union of $m/2$ DPP picks on $s(t)$ and $m/2$ frames at evenly spaced grid positions, so it has slightly fewer than $m$ distinct frames when the two halves share frames (median 62 of 64 on Video-MME).

\paragraph{Co-kriging.}
Margins are clipped to $[-15,15]$ before fitting.
The GP kernel is $c\cdot\mathrm{Mat\acute{e}rn}_{5/2}(\lambda)+\sigma_\varepsilon^2\,\delta_{tt'}$, with amplitude $c\in[10^{-3},10^{3}]$, length scale $\lambda\in[1,n]$ seconds initialized at $\max(n/64,1)$, and noise level $\sigma_\varepsilon^2\in[10^{-3},10^{3}]$, all on residuals standardized to unit variance and fitted by maximizing the marginal likelihood with two random restarts (fixed seed).
On Video-MME the fitted $\rho$ has a median of 1.12 and is non-positive for 9\% of questions; the median fitted length scale is 6.4\,s.

\paragraph{DPP and gap fill.}
The greedy MAP inverts $L_I$ with a jitter of $10^{-8}$ for numerical stability.
Gap fill keeps the chosen frames and distributes the missing ones over the empty stretches of the native timeline, including those before the first and after the last chosen frame, in proportion to their lengths.

\subsection{Co-kriging: estimator and validation}
\label{sec:suppl_cokrig}

\paragraph{Estimator.}
Under \cref{eq:cokrig}, with $\rho$ and $b$ fixed, the residuals at the probes are $\tilde{\boldsymbol\delta}_\mathcal{S}=\boldsymbol{\delta}_\mathcal{S}+\boldsymbol{\varepsilon}$ with $\boldsymbol{\varepsilon}\sim\mathcal{N}(0,\sigma_\varepsilon^2\mathbf{I})$.
The posterior of $\delta(t)$ at any grid time is Gaussian with mean $\hat\delta(t)$ given in \cref{sec:cokrig} and variance
\begin{equation}
v(t)=\kappa_\theta(t,t)-\kappa_\theta(t,\mathcal{S})\big(\kappa_\theta(\mathcal{S},\mathcal{S})+\sigma_\varepsilon^2\mathbf{I}\big)^{-1}\kappa_\theta(\mathcal{S},t).
\end{equation}
$\hat\delta(t)$ is the best linear unbiased predictor of $\delta(t)$ from $\tilde{\boldsymbol\delta}_\mathcal{S}$: among all predictors linear in the observed residuals, it has the smallest mean squared error, which equals $v(t)$~\cite{rasmussen2006gp}.
Consequently $g(t)$ is the best linear unbiased predictor of the noise-free probe response $\rho\,\mathrm{logit}\,s(t)+b+\delta(t)$ given the alignment score and the probes, with error variance $v(t)$.
$v(t)$ vanishes up to the noise level at the probes and grows with the distance to the nearest probe, which is why half of the probes are spread evenly over the timeline.
Kennedy and O'Hagan~\cite{kennedy2000cokriging} estimate $\rho$, $b$ and $\theta$ jointly by marginal likelihood; we estimate $\rho$ and $b$ first by least squares and then $\theta$ on the residuals, a plug-in approximation that is stable with only $m$ observations.
These guarantees hold under the model; whether the model fits is an empirical question, which we test next.

\paragraph{Validation against probing every frame.}
We probed every frame of the one-frame-per-second grid for all 2,700 Video-MME questions with the same model and prompt (2.8 million frame--question pairs, about 16 times the cost of $m=64$) and compare the calibrated curve $g(t)$, fitted from the 64 probes only, with these dense judgments (\cref{tab:suppl_dense}).
On frames that were not probed, $g(t)$ reaches a median within-question Spearman correlation of 0.56 with the dense judgment, against 0.40 for the raw alignment score, so the correction carries the answering model's judgment to frames it never saw.
Selecting from the dense judgments instead of from $g(t)$, with the same DPP, changes accuracy by only $+0.3$ points at $k=32$ ($z=0.4$); the frame sets share 22\% of frames exactly, and 88\% of the frames selected from $g(t)$ lie within one DPP length of a frame selected from the dense judgments.

\begin{table}[h]
  \centering
  \caption{Agreement with the dense judgment $y(t)$ on every grid frame of Video-MME (median over 2,700 questions of the within-question Spearman correlation).}
  \label{tab:suppl_dense}
  \small
  \setlength{\tabcolsep}{4pt}
  \begin{tabular}{lcc}
    \toprule
    Curve & All frames & Frames not probed \\
    \midrule
    Alignment score $\mathrm{logit}\,s(t)$ & 0.42 & 0.40 \\
    Calibrated $g(t)$ (64 probes) & 0.63 & 0.56 \\
    \bottomrule
  \end{tabular}
\end{table}

\subsection{Kernel of the discrepancy}
\label{sec:suppl_kernel}

The discrepancy $\delta(t)$ models how the bias of the alignment score drifts along the timeline.
We use a Mat\'ern-$\tfrac52$ kernel~\cite{rasmussen2006gp}, whose sample paths are twice differentiable: smooth enough to carry a bias across the gaps between probes, which are about $n/m$ seconds apart, yet able to change within a scene.
A Mat\'ern-$\tfrac12$ (exponential) kernel gives rough paths whose posterior mean reverts quickly to zero between probes, and a squared-exponential kernel gives very smooth paths that can oscillate between widely spaced probes.
In all cases the length scale is fitted per question, so the kernel family mainly sets how quickly the correction decays away from a probe.

\Cref{tab:suppl_kernel} changes only the kernel, at $k=32$ with everything else as in ProbFaS.
The two smooth kernels tie: the squared exponential is within 0.2 points of Mat\'ern-$\tfrac52$ on every benchmark ($|z|\le0.4$).
The rougher kernels lose about 0.5 points on Video-MME and 1 point on LongVideoBench, where the loss is significant for both ($z=-2.0$ and $-2.3$); MLVU is insensitive to the kernel.
This is consistent with the role of $\delta(t)$: a correction that fades within a few seconds of each probe leaves the frames between probes close to the raw alignment score, which is where the bias has to be carried.

The bottom of \cref{tab:suppl_kernel} replaces the GP, with the same probes, $\rho$ and $b$.
Calibration alone ($\hat\delta\equiv0$) loses 1.7--2.8 points, significantly on all three benchmarks ($|z|\ge2.6$).
Taking the residual of the nearest probe, or interpolating the residuals linearly, is behind the GP in five of six cases, by up to 1.5 points, and significantly in three; only the linear interpolation on MLVU ties.

\begin{table}[h]
  \centering
  \caption{Discrepancy $\hat\delta$ at $k=32$: kernel of the GP, or no GP (accuracy, \%; MLVU official score). In brackets: paired $z$ against Mat\'ern-$\tfrac52$, the default.}
  \label{tab:suppl_kernel}
  \small
  \setlength{\tabcolsep}{4pt}
  \begin{tabular}{lccc}
    \toprule
    $\hat\delta$ & Video-MME & MLVU & LVB \\
    \midrule
    \multicolumn{4}{l}{\emph{GP kernel}} \\
    Mat\'ern-$\tfrac12$ & 67.5 ($-1.0$) & 77.5 ($0.3$) & 68.1 ($-2.0$) \\
    Mat\'ern-$\tfrac32$ & 67.4 ($-1.7$) & 77.6 ($1.1$) & 68.2 ($-2.3$) \\
    Mat\'ern-$\tfrac52$ (default) & 67.9 & 77.2 & 69.2 \\
    Squared exponential & 67.9 ($-0.1$) & 77.1 ($-0.1$) & 69.3 ($0.4$) \\
    \midrule
    \multicolumn{4}{l}{\emph{No GP}} \\
    Calibration only & 65.9 ($-3.2$) & 75.5 ($-2.6$) & 66.4 ($-3.4$) \\
    Nearest probe & 67.1 ($-1.3$) & 76.2 ($-2.1$) & 67.7 ($-2.1$) \\
    Linear interpolation & 66.4 ($-2.5$) & 77.3 ($0.1$) & 68.1 ($-1.5$) \\
    \bottomrule
  \end{tabular}
\end{table}

\subsection{Temporal DPP: properties}
\label{sec:suppl_dpp}

\paragraph{Factorized gain.}
Let $D=\mathrm{diag}\big(\hat r(t_1),\ldots,\hat r(t_n)\big)$, so that $L=DCD$ with $C_{ij}=\exp\big(-(t_i-t_j)^2/(2\ell^2)\big)$.
For the set $I$ of frames selected so far, with $D_I$ the diagonal of $D$ restricted to $I$, and a candidate $i\notin I$, $L_{iI}=\hat r(t_i)\,C_{iI}D_I$ and $L_I^{-1}=D_I^{-1}C_I^{-1}D_I^{-1}$, hence
\begin{equation}
L_{iI}L_I^{-1}L_{Ii}=\hat r(t_i)^2\,C_{iI}C_I^{-1}C_{Ii},
\end{equation}
and with $C_{ii}=1$ the greedy gain is
\begin{equation}
\Delta(i\mid I)=\hat r(t_i)^2\,\nu(i\mid I),
\qquad
\nu(i\mid I)=1-C_{iI}C_I^{-1}C_{Ii}.
\end{equation}
$\nu(i\mid I)$ is the posterior variance at $t_i$ of a unit-variance process with covariance $C$ observed without noise at the selected times, so $0\le\nu\le1$; it is close to 0 within about $\ell$ of a selected frame and close to 1 when all selected frames are several $\ell$ away.
The gain is therefore relevance times temporal novelty.

\paragraph{Peaks and coverage.}
The first selected frame is the most relevant one, since $\nu(i\mid\emptyset)=1$.
A later frame is chosen only if its relevance outweighs its loss of novelty: a frame next to a chosen peak must be far more relevant than a frame in an uncovered part of the video.
Two limits show the trade-off.
When $\hat r$ is constant, the gain depends only on $\nu$ and the greedy spreads the frames over the timeline like a space-filling design; when $\hat r$ has a few isolated peaks far above the rest, the greedy takes the peaks first, at most one per neighborhood of width about $\ell$, as non-maximum suppression would.
The temporal DPP interpolates between these behaviors according to the calibrated relevance of each question, without a hand-set trade-off weight.
Since $\log\det L_I$ is submodular as a function of $I$, the greedy is the standard approximate MAP algorithm for DPPs~\cite{kulesza2012dpp,chen2018fastdpp}.

\paragraph{Empirically.}
On the raw alignment score, the temporal DPP is 1.8 points more accurate than the 32 highest-scoring frames (\cref{sec:suppl_topk}), and the selection-rule ablation in \cref{tab:ablation} compares the DPP with other rules on the calibrated relevance.
At $k=32$ on Video-MME (videos with more than 48 grid frames), the largest gap between consecutive ProbFaS frames is a median 17.5\% of the video, against 3.1\% for uniform sampling: the DPP concentrates frames on relevant moments rather than covering the video evenly.

\section{Experimental setup}
\label{sec:suppl_setup}

\subsection{Baseline implementations}
\label{sec:suppl_baselines}

Every baseline selects frames with its released code and default settings; where a release covers only some benchmarks or budgets, we ran its selector ourselves, as noted below.
The selected frame indices are decoded from the original video and passed to the answering model exactly as for ProbFaS.
A question whose list is empty falls back to $k$ uniformly sampled frames, except AKS at $k<32$ (below).

\paragraph{Uniform.}
The model's own video reader with $k$ evenly spaced frames.

\paragraph{BOLT~\cite{liu2025bolt}.}
CLIP ViT-L/14 similarities on the 1\,fps grid, from the released features, followed by inverse transform sampling with power 3 (2.5 on LongVideoBench), as in the released scripts.
For $k\le32$ we use the released frame lists, which keep repeated picks and are shown as they are: of the 32 frames at $k=32$, on average 30.3, 31.6 and 26.8 are distinct on Video-MME, MLVU and LongVideoBench.
For $k=64$ and 128, which are not released, we run its released selection script on the released features with the same settings.

\paragraph{AKS~\cite{tang2025aks}.}
BLIP-ITM-large scores on the 1\,fps grid, released for Video-MME and LongVideoBench and extracted with the same model and sampling for MLVU, followed by the released judge-and-split selector with $t_1=0.8$, $t_2=-100$, depth 5 and ratio 1.
After the split, the released script gives a segment at depth $d$ a quota of $\lfloor k/2^d\rfloor$ frames, which is zero for fully split segments when $k<32$; the lists are then empty for most questions (2,199 and 2,372 of 2,700 on Video-MME at $k=8$ and 16), so these cells are missing.
At $k=32$ our lists match the released Video-MME lists on all questions; they are capped at, not fixed to, $k$ frames (2,649 of 2,700 Video-MME lists have 32).

\paragraph{AdaQ~\cite{zhang2026adaq}.}
CLIP ViT-B/32 frame probabilities on the 1\,fps grid from its feature extractor, followed by its softmax and top-$p$ selector with $k_{\min}=k$, variance scale 0.5 and $p=0.95$.

\paragraph{QCA~\cite{peng2026qca}.}
BLIP-ITM-large scores and BLIP image features on the 1\,fps grid, or on 128 evenly spaced frames when the grid is shorter, followed by its selector with 16 segments, $\gamma=0.5$, $\beta=0.5$ and $\tau=1$.

\paragraph{MDP$^3$~\cite{sun2025mdp3}.}
The release wires its selector only into a Video-MME pipeline, so we call the selector directly on all three benchmarks.
It embeds 128 evenly spaced candidate frames, saved as JPEG as in the release, and the question with its options with SigLIP-SO400M, and runs its segment-wise conditional DPP with $\lambda=0.2$ and condition size 1.
The release hard-codes 8 selected frames; we set the number to $k$ and the segment size to 32 for $k\le32$ and 128 otherwise, so at $k\ge64$ a single segment remains and the selector reduces to a conditional DPP.
At $k=128$ the budget equals the candidate pool, and the lists are empty for 196, 26 and 123 questions of Video-MME, MLVU and LongVideoBench, which fall back to uniform frames; one LongVideoBench video (3 questions) also has empty lists at $k=16$--64.

\paragraph{GIFT~\cite{ma2026gift}.}
Its released SigLIP-SO400M scorer and irreplaceability selector (refine group 9), unchanged, on its default pool of 128 evenly spaced candidate frames, with the question and options as text; we export its frame indices.
At $k=128$ it keeps the whole pool, so its list is the 128 evenly spaced candidates, replayed like its other lists.
With LLaVA-Video-7B at 32 frames our replay gives 65.0 on Video-MME, as reported by GIFT, and 59.3 on LongVideoBench against its 60.2; its MLVU protocol differs from ours (its uniform baseline is 61.2, ours 65.0).

\paragraph{KFC~\cite{fang2026narkfc}.}
The frame-selection stage of Nar-KFC, without the text narratives that Nar-KFC adds for the frames it does not select, so that it competes on the frames alone like the other methods.
CLIP ViT-L/14 at 336\,px embeds the question and the frames of the 1\,fps grid; the edge score of two frames is their query similarity plus $\exp(-\text{frame similarity})$, with the query similarity on the diagonal, and its released greedy search with local search (radius 1) selects the $k$ frames, with the shrink rule of each benchmark's released call site.
The release solves the integer quadratic program with CPLEX on MLVU and uses the greedy search on Video-MME and LongVideoBench; we use the greedy search on all three.

\paragraph{FOCUS~\cite{zhu2026focus}.}
Its released keyframe selector with BLIP-ITM-large and its default coarse-to-fine bandit, at each $k$.

\paragraph{DIG~\cite{li2026dig}.}
The authors release, for each question, its query type, decided by Qwen3-Next-80B-A3B, and the frame rewards that Qwen3-VL-8B assigns to the shot frames of local questions.
We run the released refinement script on these at each $k$; global questions receive uniform frames.
Its frames are therefore chosen with the same answering model as ours.

\paragraph{A.I.R.~\cite{zou2026air}.}
Its released LMMs-Eval model wrapper with Qwen3-VL-8B in the loop and its default settings, at most $k$ frames and 786,432 pixels per frame: EVA02-CLIP-L similarities at 2\,fps, cached per video, a query-conditioned Gaussian mixture that splits the video into events, density-aware sampling within events, and up to six rounds of single-frame refinement by the MLLM.
Since the MLLM is inside the loop, its frames cannot be replayed with another answering model (\cref{tab:other_mllm}).

\paragraph{WFS-SB~\cite{chen2026wfssb}.}
For $k\le32$ we use the frame lists released by the authors, computed with BLIP-2-ITM on the 1\,fps grid, its default scorer.
For $k\ge64$, which are not released, we run its released selector on our BLIP-2-ITM grid with the default hyperparameters of its pipeline (wavelet denoising with db4, peak-based segment boundaries, adaptive budget allocation and diversity-aware selection within segments).
Our grid stores no image embeddings, so the diversity term uses temporal distance, which the released code does when no embeddings are given; questions whose grid has fewer than $k$ frames receive $k$ uniform frames, as in its pipeline.

\paragraph{FRAG~\cite{huang2025frag}.}
The answering model, Qwen3-VL-8B, scores 256 evenly spaced frames, one frame per forward pass, with FRAG's prompt, which asks whether the image gives the details needed to answer the question (the question alone, with the options ``A. yes'' and ``B. no''); the score is the probability of ``A'', and the $k$ highest-scoring frames are kept in temporal order.
FRAG used InternVL with one 448-pixel tile per frame; we pass each frame at the resolution of our probes, so one scoring pass costs as much as one probe of ProbFaS, and FRAG spends 256 of them per question against 64 for ProbFaS.
Frames tied at the $k$-th score, which the saturated bf16 probabilities produce, are spread evenly in time.

\paragraph{OTFS~\cite{zhao2026otfs}.}
Its released code with default settings, at all five budgets (the paper reports $k=32$ and 64).
Its evidence is the BLIP-ITM-large scores on the 1\,fps grid that AKS uses, the frame rewards that DIG releases for Qwen3-VL-8B (the paper uses those for Qwen2.5-VL-7B), linearly interpolated between the frames DIG scores, and DINOv2-base transitions between adjacent frames; questions that DIG's router marks as global get no reward or transition evidence, as in the paper.
A semi-unbalanced optimal transport fuses the three and a dynamic program extracts the frames, so OTFS also relies on DIG's 80B query router.
Its lists are shorter than $k$ only when the 1\,fps grid has fewer than $k$ frames.

\paragraph{Q-Frame~\cite{zhang2025qframe}.}
Its released preprocessor, ported to the Qwen3-VL wrapper of LMMs-Eval: 128 evenly spaced candidates are ranked by Long-CLIP-L at 336\,px with Gumbel-max sampling, and the top 4, next 8 and next 32 frames are passed at $\tfrac12$, $\tfrac14$ and $\tfrac18$ of the native resolution as a multi-image input.

\paragraph{FORTE~\cite{huang2026forte}, as reported.}
FORTE has no released code, so it is not in \cref{tab:main}.
It reports results with Qwen3-VL-8B at $k=64$ in its own setup, with BLIP scores of the question alone, and its uniform baselines differ from ours.
By gain over each setup's own uniform sampling, FORTE gains 2.2, 5.5 and 5.7 points on Video-MME, MLVU and LongVideoBench (69.9, 75.8 and 68.1; against 67.7, 70.3 and 62.4), and ProbFaS 2.1, 10.0 and 5.8 points (69.1, 78.1 and 69.0; against 67.0, 68.1 and 63.2; \cref{tab:main}).
FORTE reports budgets below 64 frames only as four-benchmark means, so we compare at $k=64$.

\subsection{LENS}
\label{sec:suppl_lens}
LENS first asks the answering MLLM, in a text-only call, how to split the $k$ images between a spatial and a temporal zoom.
Spatial-zoom frames are chosen by BLIP matching scores and keep their resolution, with query-irrelevant regions masked by CLIP attention.
Each temporal-zoom slot, chosen by a graph diffusion over the scores, is a hyperframe that tiles four consecutive candidate frames into a $2\times2$ image downscaled to the original frame size.
These images do not fit the frame-index protocol of the other methods, so we run the released pipeline end to end, with its default settings, BLIP as its scoring model and Qwen3-VL-8B as the answering model, which receives the $k$ images as a video, as in the released code.
We score its answers by the predicted option letter, like every other method.
Its own evaluation counts an answer as correct when the answer letter occurs anywhere in the prediction, which accepts, for example, ``(B) Driver's view'' for the answer D, and raises its question accuracy on MLVU and LongVideoBench by 0.8 to 2.4 points.
The pipeline fails on 3 MLVU and 5 LongVideoBench questions whose text exceeds the 512 tokens of its scoring model; we count them as wrong, which lowers its scores by at most 0.4 points.

\subsection{Other answering models}
\label{sec:suppl_other_mllm}

LLaVA-Video-7B-Qwen2~\cite{zhang2024llavavideo} runs through the \texttt{llava\_vid} wrapper of LMMs-Eval with the \texttt{qwen\_1\_5} conversation template and exactly 32 frames.
InternVL3.5-8B~\cite{wang2025internvl35} runs through the \texttt{internvl3\_5} wrapper in video mode, with two tiles and a thumbnail per frame and the prompt ``Frame $i$: \texttt{<image>}''.
Neither model receives frame timestamps, so uniform sampling and replayed lists reach the model in the same form.
With uniform sampling, LLaVA-Video-7B scores 62.5 on Video-MME and 58.5 on LongVideoBench at 32 frames, in line with the published 63.3 and 58.2 at 64 frames.
For the own-probe runs, each model judges the same probe set as Qwen3-VL-8B with the same yes/no prompt, each frame preprocessed as in its 32-frame answer pass; calibration, DPP and gap fill are unchanged.
For LLaVA-Video we also mapped its margins, per benchmark and without labels, to the mean and spread of the Qwen3-VL-8B margins; this gives 64.1, 70.1 and 60.5 on Video-MME, MLVU and LongVideoBench, against 64.7, 69.4 and 61.0 without the mapping.
MDP$^3$'s lists are empty for the three questions of one LongVideoBench video; for those questions every wrapper falls back to uniform frames, as in \cref{tab:main}.

\paragraph{Own probe.}
InternVL3.5-8B with its own probe is within about one point of the Qwen-probe lists on MLVU and LongVideoBench ($z=-1.1$ on both).
LLaVA-Video-7B with its own probe loses 0.9 to 2.9 points against the Qwen-probe lists ($z=-1.4$, $-3.6$ and $-2.4$ on Video-MME, MLVU and LongVideoBench), though it stays above uniform sampling.
Its ranking of the probe frames agrees less with that of Qwen3-VL-8B (median Spearman 0.53--0.61, against 0.61--0.70 for InternVL), and its yes/no margins span 8 to 12 times less (4 to 5 times less for InternVL).
Unlike for the 2B probe (\cref{sec:suppl_sensitivity}), the mapping of its margins above does not help consistently, so the limit is its ranking, not its scale.

\paragraph{Comparison with DAFS.}
DAFS~\cite{wang2026dafs}, a training-free selector that reads relevance from the attention of a 2B MLLM in one forward pass, reports results for the same two answering models in its own setup (\cref{tab:other_mllm}, bottom; resolution, prompts and candidate pool not stated).
Its uniform baselines differ from ours (62.9 against 60.2 for InternVL on LongVideoBench), so we compare gains over each setup's own uniform sampling, on Video-MME, MLVU and LongVideoBench: with InternVL, DAFS gains 3.0, 2.7 and 2.9 points and ProbFaS 2.2, 5.9 and 4.0 (own probe 2.5, 5.3 and 3.0); with LLaVA-Video, DAFS gains 3.1, 2.7 and 2.7 and ProbFaS 3.1, 7.3 and 4.6.
ProbFaS thus gains more on MLVU and LongVideoBench and as much or less on Video-MME; in absolute accuracy DAFS is ahead with InternVL on Video-MME and LongVideoBench.

\section{Additional results}
\label{sec:suppl_results}

\subsection{Selection rules on the calibrated relevance}
\label{sec:suppl_plugin}

For the selection-rule ablation of \cref{tab:ablation}, each question is calibrated exactly as in ProbFaS, with the same probes and the same fixed seed, and $\hat r=\sigma(g)$ on the 1\,fps grid replaces the rule's own score at $k=32$.
Each rule keeps the defaults of our baseline runs:
\begin{itemize}
  \item Top-$k$: the $k$ frames with the largest $\hat r$.
  \item BOLT: inverse transform sampling with power 3 (2.5 on LongVideoBench). The released BOLT lists keep repeated picks; on the sharper $\hat r$ this would leave about 22 distinct frames of 32 on Video-MME, so a repeated pick moves to the nearest unused grid frame.
  \item AKS: judge-and-split with its \texttt{meanstd} rule, threshold $t_1=0.8$ and depth 5.
  \item AdaQ: softmax and top-$p$ sampling, the core of its selector, with $k_{\min}=k$, variance scale 0.5 and $p=0.95$.
  \item WFS-SB: wavelet denoising, segment boundaries and per-segment MMR with its default settings.
  \item FORTE~\cite{huang2026forte}: the exact maximizer of $\sum_{i\in I}r_i+\beta\sum_{j}\log(1+d_j/\tau)$ over $|I|=k$, where $r$ is $\hat r$ min-max normalized per question, $d_j$ are the gaps between consecutive selected times including both ends of the video, $\tau=T/(k+1)$ for a video of $T$ seconds and $\beta=1$, its defaults, computed by its dynamic program over the whole 1\,fps grid; our implementation matches exhaustive search on 300 random small instances. Its adaptive scoring stage is not used, since $\hat r$ is known on every frame.
\end{itemize}
Questions whose grid has at most $k$ frames take the whole grid, and lists shorter than $k$ are completed by the gap fill of \cref{sec:dpp}.
As a check, the temporal DPP on the same $g$ reproduces the ProbFaS lists of \cref{tab:main} on every question.
QCA, MDP$^3$, GIFT and KFC select from image features, FOCUS queries its scorer inside its bandit, DIG and A.I.R.\ query the MLLM during selection, OTFS fuses DIG's MLLM rewards, and LENS runs its own pipeline, so their rules cannot take $\hat r$ as input; FRAG's rule is top-$k$.
\Cref{tab:suppl_plugin} gives the paired comparison with ProbFaS and each method with its own scorer.
Top-$k$ ties the temporal DPP on LongVideoBench and trails it by 1.1 and 0.8 points on Video-MME and MLVU; judge-and-split mostly ends with one frame in each of 32 segments, close to uniform sampling, and the exact relevance-plus-coverage optimum of FORTE ties the DPP on Video-MME but loses 3.6 and 2.0 points on MLVU and LongVideoBench.
The calibration transfers: $\hat r$ raises BOLT and AKS over their own scores on all three benchmarks, by 1.4 to 5.7 points, and AdaQ and WFS-SB on two of the three; their own runs also differ in grid and sampling, so we read these gains as indicative.

\begin{table}[h]
  \centering
  \caption{Selection rules on the calibrated relevance $\hat r$ at $k=32$ (accuracy, \%). Own: the method with its own scorer (\cref{tab:main}). $z$: paired against ProbFaS, negative when the rule is worse. Best per $\hat r$ column in \textbf{bold}, second \underline{underlined}.}
  \label{tab:suppl_plugin}
  \small
  \setlength{\tabcolsep}{3pt}
  \resizebox{\linewidth}{!}{%
  \begin{tabular}{l ccc ccc ccc}
    \toprule
    & \multicolumn{3}{c}{Video-MME} & \multicolumn{3}{c}{MLVU} & \multicolumn{3}{c}{LongVideoBench} \\
    \cmidrule(lr){2-4}\cmidrule(lr){5-7}\cmidrule(lr){8-10}
    Rule & Own & $\hat r$ & $z$ & Own & $\hat r$ & $z$ & Own & $\hat r$ & $z$ \\
    \midrule
    Temporal DPP (ProbFaS) & -- & \underline{67.9} & -- & -- & \textbf{77.2} & -- & -- & \textbf{69.2} & -- \\
    Top-$k$ & -- & 66.8 & $-1.9$ & -- & \underline{76.4} & $-1.2$ & -- & \textbf{69.2} & $0.0$ \\
    BOLT & 64.7 & 66.1 & $-2.8$ & 70.5 & 72.3 & $-5.8$ & 62.8 & \underline{68.5} & $-0.8$ \\
    AKS & 65.6 & 67.0 & $-1.3$ & 68.1 & 70.2 & $-8.1$ & 61.5 & 65.3 & $-4.1$ \\
    AdaQ & 64.1 & 67.0 & $-1.3$ & 72.5 & 71.9 & $-6.7$ & 63.4 & 67.1 & $-2.4$ \\
    WFS-SB & 66.2 & 66.9 & $-1.7$ & 75.0 & 73.5 & $-5.3$ & 66.9 & 67.8 & $-1.7$ \\
    FORTE & -- & \textbf{68.0} & $0.1$ & -- & 73.6 & $-4.6$ & -- & 67.2 & $-2.6$ \\
    \bottomrule
  \end{tabular}
  }
\end{table}

\subsection{Probe budget, probe model and sensitivity}
\label{sec:suppl_sensitivity}

\paragraph{Dense probing.}
Judging every frame of the grid (\cref{tab:probe_budget}) adds 0.2--0.4 points at $k\le32$ on Video-MME and 0.7 and 0.8 points at $k=64$ and 128, mostly on long videos.
It is not an upper bound: at $k=32$ it adds 0.8 points on MLVU and loses 1.0 on LongVideoBench ($z=0.3$ and $-1.2$; \cref{tab:ablation}).
Its gain at large $k$ needs near-dense probing and is not reached by doubling $m$.

\paragraph{Probe model.}
Probing with Qwen3-VL-2B while answering with 8B falls to the BLIP-2-only level on all three benchmarks (\cref{tab:ablation}).
Its judgments rank frames similarly to the 8B model (median Spearman about 0.7 on shared frames), but its yes/no margins span about three times less, so the corrected quality $\sigma(g)$ becomes nearly flat and the DPP degrades toward uniform spacing.
Multiplying the 2B margins by one temperature fitted on Video-MME ($\times2.4$) recovers half of the loss on LongVideoBench (67.2 against 65.0, and 69.2 for the 8B probe).
Probing with the answering model itself avoids this calibration.

\paragraph{Sensitivity.}
We swept the exponent $\gamma$ of the DPP quality $\hat r^\gamma=\sigma(g)^\gamma$ over $\{0.5,1,2,4\}$ and the DPP length over $\{0.5,1,2\}$ times its default on a 270-question slice of Video-MME.
No setting differs significantly: the best one on the slice ($\gamma=2$, half length) scored 68.2, 76.6 and 68.1 on the full Video-MME, MLVU and LongVideoBench, against 67.9, 77.2 and 69.2 for the default.
Only removing the quality ($\gamma=0$, a purely temporal DPP) clearly hurt, by 4.5 points on the slice.

\subsection{Variants that did not help}
\label{sec:suppl_negative}

\Cref{tab:suppl_negative} lists three variants we tried to spend the probes or the frames differently; none improves on ProbFaS.

\begin{table}[h]
  \centering
  \caption{Variants of ProbFaS (accuracy, \%). ``Default'' is ProbFaS with $m=64$ single-frame probes.}
  \label{tab:suppl_negative}
  \small
  \setlength{\tabcolsep}{3pt}
  \begin{tabular}{lccccc}
    \toprule
    & \multicolumn{3}{c}{Video-MME} & MLVU & LVB \\
    \cmidrule(lr){2-4}\cmidrule(lr){5-5}\cmidrule(lr){6-6}
    Variant & $k{=}8$ & 32 & 128 & 32 & 32 \\
    \midrule
    Default & 63.9 & 67.9 & 70.4 & 77.2 & 69.2 \\
    GP-UCB second probe round & 63.0 & 67.4 & 70.2 & -- & -- \\
    Two-frame clip probe & -- & 67.9 & 70.1 & -- & 68.7 \\
    Frames paired with neighbors & -- & 68.0 & 70.4 & 76.2 & 69.5 \\
    \midrule
    Probe every frame (reference) & 64.3 & 68.2 & 71.2 & 78.0 & 68.2 \\
    \bottomrule
  \end{tabular}
\end{table}

\paragraph{GP-UCB second probe round.}
Instead of placing all probes at once, 32 probes are placed by the score-peak and uniform rule, the calibration is fitted, and the remaining 32 probes go where $g(t)+2\sqrt{v(t)}$ is largest, updating $v(t)$ after each pick as if the probe had been observed~\cite{srinivas2010gpucb}.
The acquisition does find more relevant frames (46\% of second-round probes are judged helpful, against 33\% in the first round), but accuracy does not improve; it is the extrapolation of the probes, not their placement, that matters.

\paragraph{Two-frame clip probe.}
Each probe shows the MLLM the clip $(t,t+1\,\mathrm{s})$ instead of a single frame, at the same token cost, so that it can see motion.
Its margins are about 1.3 times narrower than single-frame margins; the LongVideoBench entry rescales them by 1.3 (67.8 without rescaling).
It helps motion and counting questions and costs detail questions about equally.

\paragraph{Frames paired with neighbors.}
Qwen3-VL merges every two consecutive input frames into one token group.
Selecting $k/2$ frames and showing each with its neighbor one second later makes every group a 1\,s clip, but shows half as many distinct moments; it ties on Video-MME and LongVideoBench and loses on MLVU, where the order and plot tasks need many distinct events.

\subsection{Results by video duration and question type}
\label{sec:suppl_detailed}

\Cref{tab:suppl_dur_videomme,tab:suppl_dur_mlvu,tab:suppl_dur_lvb} break the results of \cref{tab:main} down by video duration, \cref{tab:suppl_mlvu_tasks} by MLVU task, and \cref{tab:suppl_vmme_cats,tab:suppl_lvb_cats} by Video-MME and LongVideoBench question category, each for every frame budget.
All cells except MLVU M-avg, the mean over tasks, are question accuracy recomputed from the per-question outputs of the runs in \cref{tab:main}.
Averaged over the five budgets, ProbFaS gains most over uniform sampling on questions about one or a few specific moments: MLVU counting, order, needle and plot QA (11.5 to 25.7 points), Video-MME OCR and temporal reasoning (11.3 and 7.1 points), and the LongVideoBench perception questions, which refer to a single moment (level 1, 10.1 points against 5.3 on relation questions).
It gains least on questions that need many moments or the whole video, such as Video-MME counting (1.9 points), MLVU topic reasoning (0.0 points) and Video-MME information synopsis (1.3 points).
The gain is larger on longer videos of MLVU (7.1 points on 3--7\,min videos, 11.0 to 13.1 on longer ones) and LongVideoBench (0.0 points on 8--15\,s videos, 10.6 on 15--60\,min ones), but on Video-MME it is largest on medium videos (5.9 points, against 3.4 on short and 2.6 on long ones).

\subsection{Cost measurement}
\label{sec:suppl_cost}

\Cref{tab:cost} in the main paper gives the selection times by video duration, with videos on local disk and decoding included.
ProbFaS takes 11.5\,s per question, about a quarter of LENS, QCA and AKS, a sixth of A.I.R., a seventh of DIG and a twelfth of OTFS; DIG and OTFS in addition call an 80B LLM to route each query.
It is close to AdaQ, FRAG and FOCUS, and about three times BOLT and twice MDP$^3$.
The 64 probes cost 3.5\,s and grow only slightly with the length of the video (2.7 to 4.0\,s), since $m$ is fixed.
The rest is the BLIP-2 grid, which grows linearly with the length and dominates on long videos (19.5 of 23.7\,s); it is the grid WFS-SB also scores, so ProbFaS costs WFS-SB plus 3.5\,s.

\paragraph{Protocol.}
\Cref{tab:cost} uses 30 Video-MME videos, 10 per duration group drawn with a fixed seed, with all their 90 questions (8.3 hours of video).
Each method runs as its own job on one GH200 GPU with 16 CPU cores and the videos copied to node-local disk, so that file-system load does not enter.
Model loading is excluded, and the first video is processed once more as a warm-up and not recorded.
Times include video decoding where the method decodes; work done once per video, such as image features, is divided among the questions on that video.
Baselines run with their released code as is, except BOLT, WFS-SB and the reward step of DIG. BOLT is timed with our implementation of its pipeline, CLIP-L/14 on the 1\,fps grid once per video followed by inverse transform sampling; it uses the 336\,px model rather than the 224\,px model of the release, so its time is an upper estimate. WFS-SB runs its released selector (wavelet denoising, adaptive budget allocation and diversity-aware intra-segment selection) on the ProbFaS BLIP-2 grid, which adds 7\,ms per question.
The released code measures diversity in the intra-segment step with BLIP-2 image embeddings when they are given and by temporal distance otherwise; as in our WFS-SB runs of \cref{tab:main}, the grid stores no embeddings, so the timing uses temporal distance.
FOCUS and A.I.R.\ do not expose per-question times, so we divide the wall-clock time of the whole run by 90; for A.I.R.\ this includes the answer pass and model loading (77.5\,s), hence the approximate selection time of 75\,s.
DIG first picks shot frames from DINOv2 features at 2\,fps and then asks the answering MLLM for a reward on each shot frame with its prompt; one reward takes 0.81\,s at batch 16, with 21, 93 and 448 shot frames per local question on short, medium and long videos.
We time up to 32 rewards per question and extrapolate to all shot frames.
Its 80B query router is not timed, and the vLLM deployment of its release would generate faster than our implementation, so the reward part is an upper estimate.

For ProbFaS, BLIP-2's ViT-g image encoder runs once per frame and is reused by the questions of the same video, which gives identical scores (largest difference 0) to running the whole model on every frame--question pair.
Without this reuse, as in our original extraction, the grid takes 13.3\,s and ProbFaS 16.9\,s per question.
The 64 probes include choosing and decoding the probe frames and run in batches of 16.
The answer pass of Qwen3-VL-8B takes about 2.0\,s at $k=32$ (11.0k input tokens on average) and 6.5\,s at $k=128$ (43.6k tokens).
Its input is the same for every selector at equal $k$, and its time on identical inputs varied by up to 1.5$\times$ with the order of runs, so we give one number per $k$.

LENS runs selection and answering in one process; a timer in its loop separates the two.
Under the protocol above it takes 20.3, 53.4 and 196.9\,s of selection per question on short, medium and long videos (90.2\,s overall, the mean of two runs on different nodes, 93.7 and 86.6\,s), most of it decoding every frame at full resolution once per question.
Its full Video-MME run, on the same questions with the videos on the shared file system, took about half as long, 16.0, 29.7 and 86.5\,s per question without the answer pass, for reasons we could not isolate; \cref{tab:cost} reports these lower numbers.

DAFS~\cite{wang2026dafs} has no public code, so it is not in \cref{tab:cost}.
It scores frames with one forward pass of a 2B MLLM over at most 32,000 visual tokens and reports 0.47\,s of selection and 4.90\,s of candidate decoding per question on short Video-MME videos, on an A100 and with a Qwen2.5-VL-7B answering model.
Its scoring pass is likely cheaper than our BLIP-2 grid on long videos, but we cannot time it under our protocol.

\subsection{Qualitative examples}
\label{sec:suppl_qualitative}

\begin{figure*}[p]
    \centering
    \includegraphics[width=\textwidth,height=0.4\textheight,keepaspectratio]{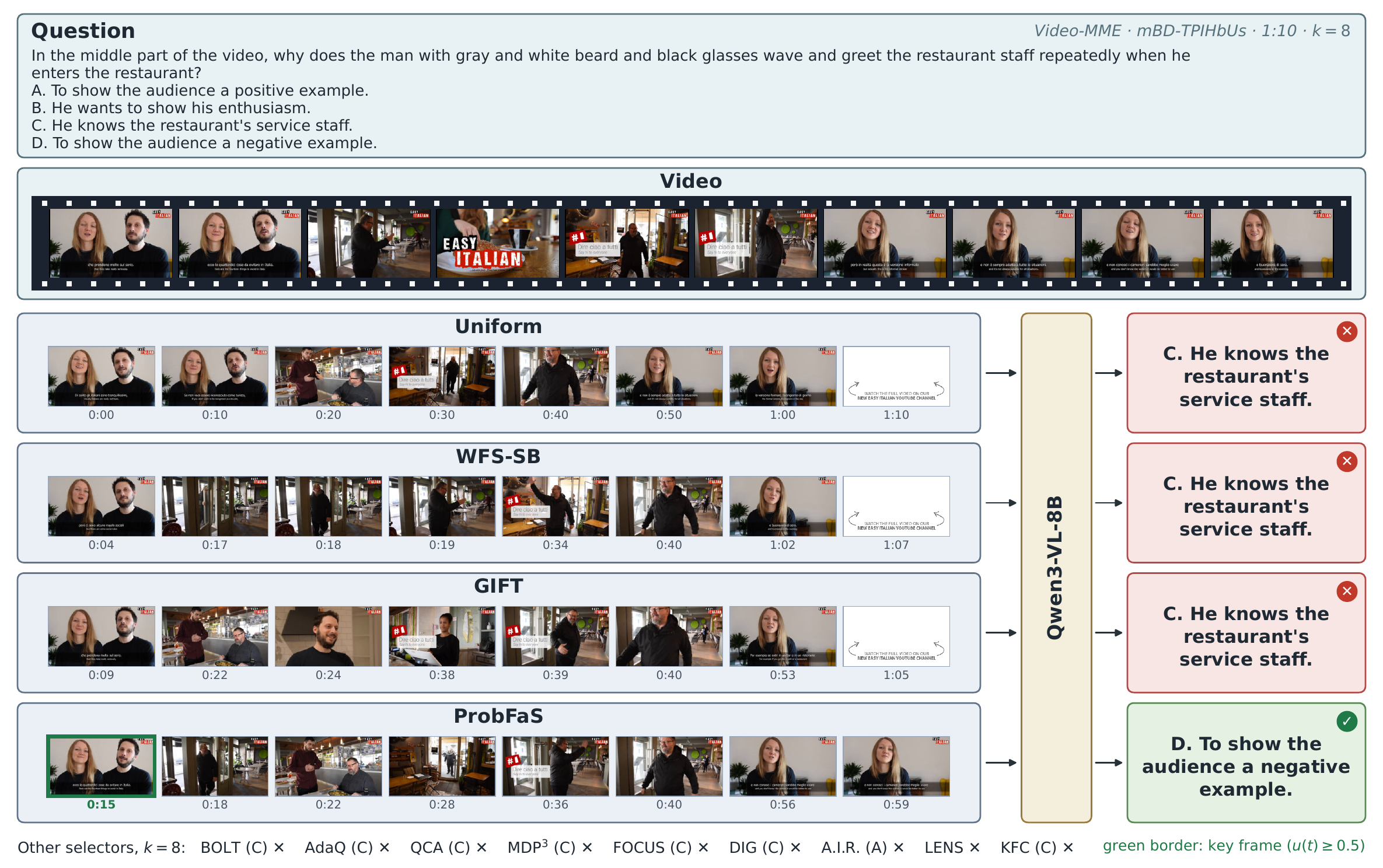}
    \caption{Video-MME, action reasoning, 70-s video from a language-learning channel, $k=8$.
    ProbFaS selects the introduction at 0:15, whose subtitle announces ``fourteen things to avoid in Italy'', so the greeting shown later is a negative example.
    Uniform sampling, WFS-SB and GIFT show the same hosts at other moments, and all fourteen baselines with a $k=8$ result answer wrongly.}
    \label{fig:qual_vmme_action}
    \vspace{2.5em}
    \includegraphics[width=\textwidth,height=0.4\textheight,keepaspectratio]{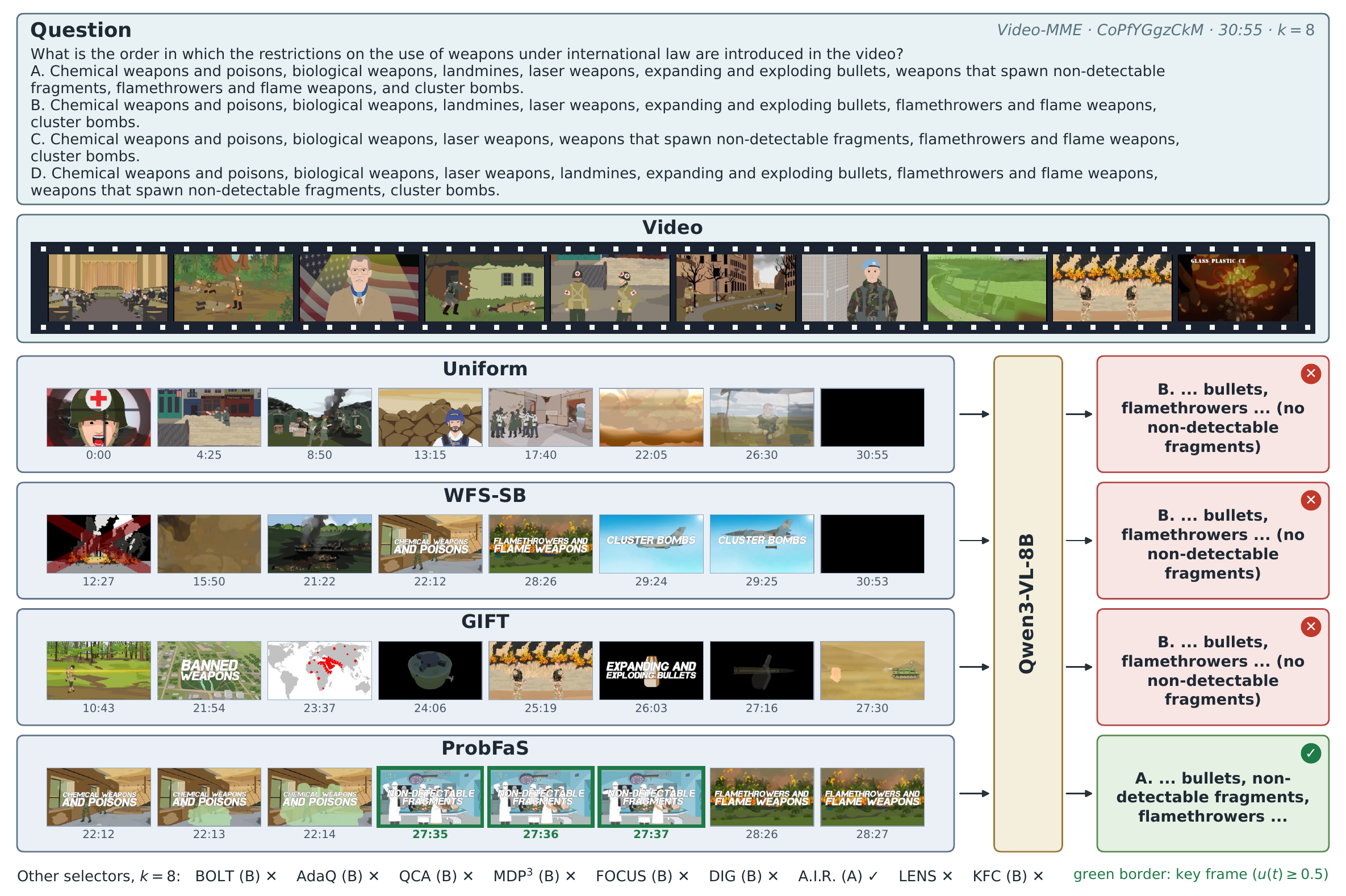}
    \caption{Video-MME, temporal reasoning, 31-min animated explainer, $k=8$.
    Options A and B list the same weapons except the ``weapons that spawn non-detectable fragments''.
    ProbFaS selects the title card of that section (27:35--27:37).
    Uniform sampling, WFS-SB and GIFT miss it and choose B, as does every other selector except A.I.R.\ and OTFS.}
    \label{fig:qual_vmme_temporal}
\end{figure*}

\begin{figure*}[p]
    \centering
    \includegraphics[width=\textwidth,height=0.4\textheight,keepaspectratio]{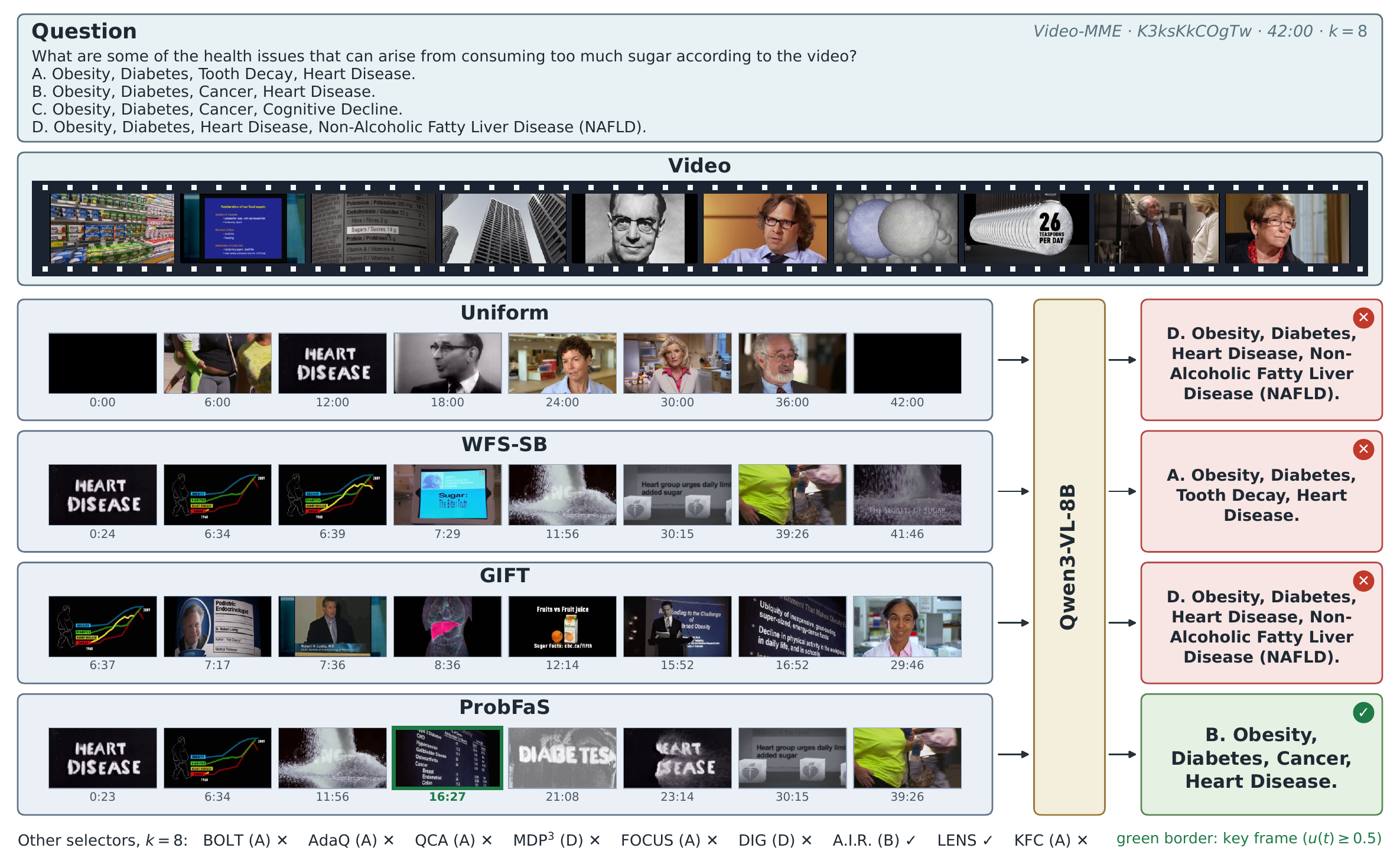}
    \caption{Video-MME, object recognition, 42-min documentary about sugar, $k=8$.
    ProbFaS selects a slide at 16:27 that lists coronary heart disease and cancer among the diseases attributable to obesity, which singles out option B.
    Uniform sampling, WFS-SB and GIFT miss it and answer D or A; of the other selectors, only A.I.R., LENS, FRAG and OTFS answer correctly.}
    \label{fig:qual_vmme_object}
    \vspace{2.5em}
    \includegraphics[width=\textwidth,height=0.4\textheight,keepaspectratio]{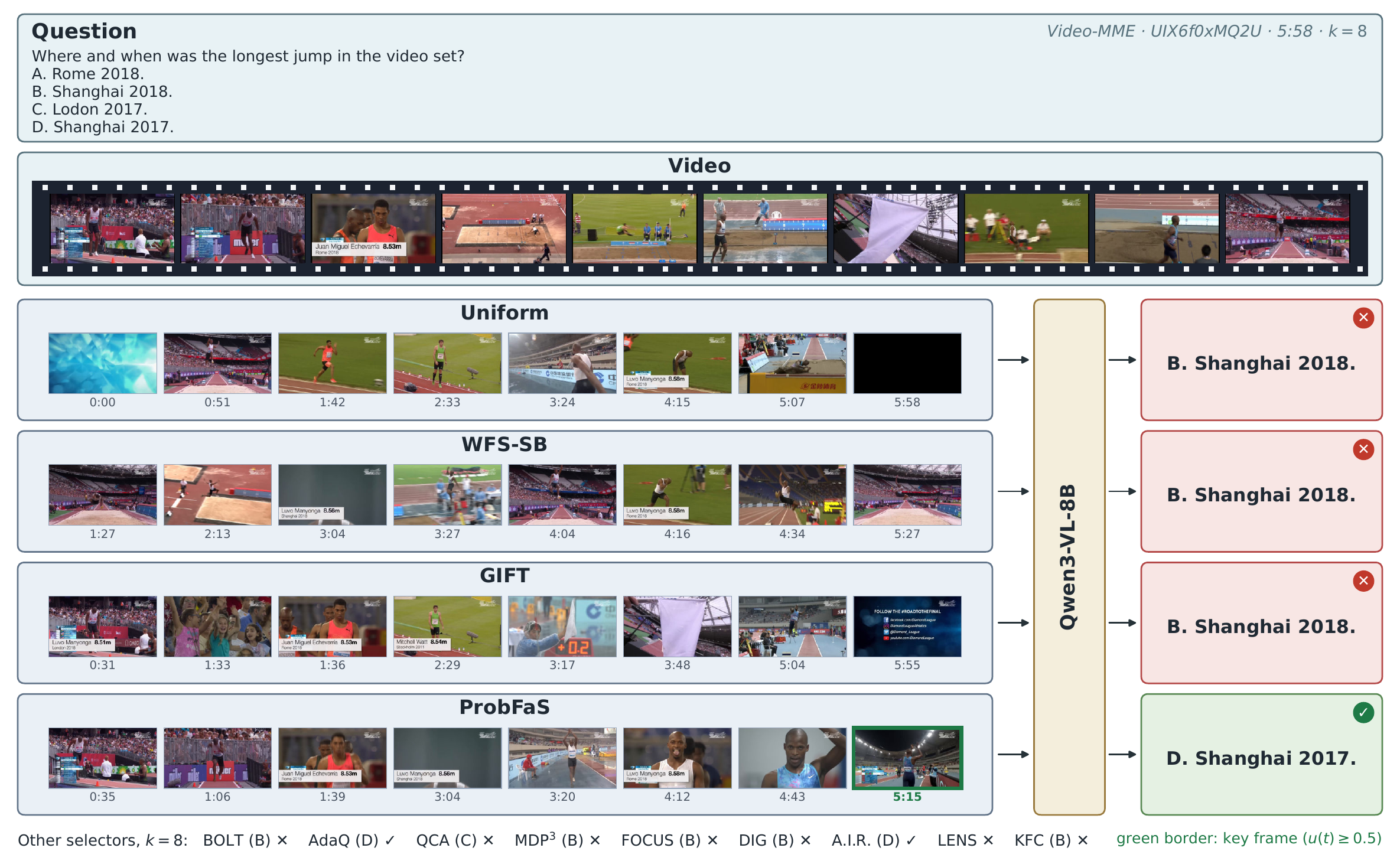}
    \caption{Video-MME, OCR, 6-min compilation of Diamond League long jumps, $k=8$.
    The answer is read from overlays: a name card at 4:41 (``Luvo Manyonga 8.61m, Shanghai 2017'') and the scoreboard at 5:15, which ProbFaS selects.
    Uniform sampling, WFS-SB and GIFT show other moments and answer B, the answer without any frame.
    AdaQ and KFC select the name card, but only AdaQ answers correctly; A.I.R., FRAG and OTFS are also correct.}
    \label{fig:qual_vmme_ocr}
\end{figure*}

\begin{figure*}[p]
    \centering
    \includegraphics[width=\textwidth,height=0.4\textheight,keepaspectratio]{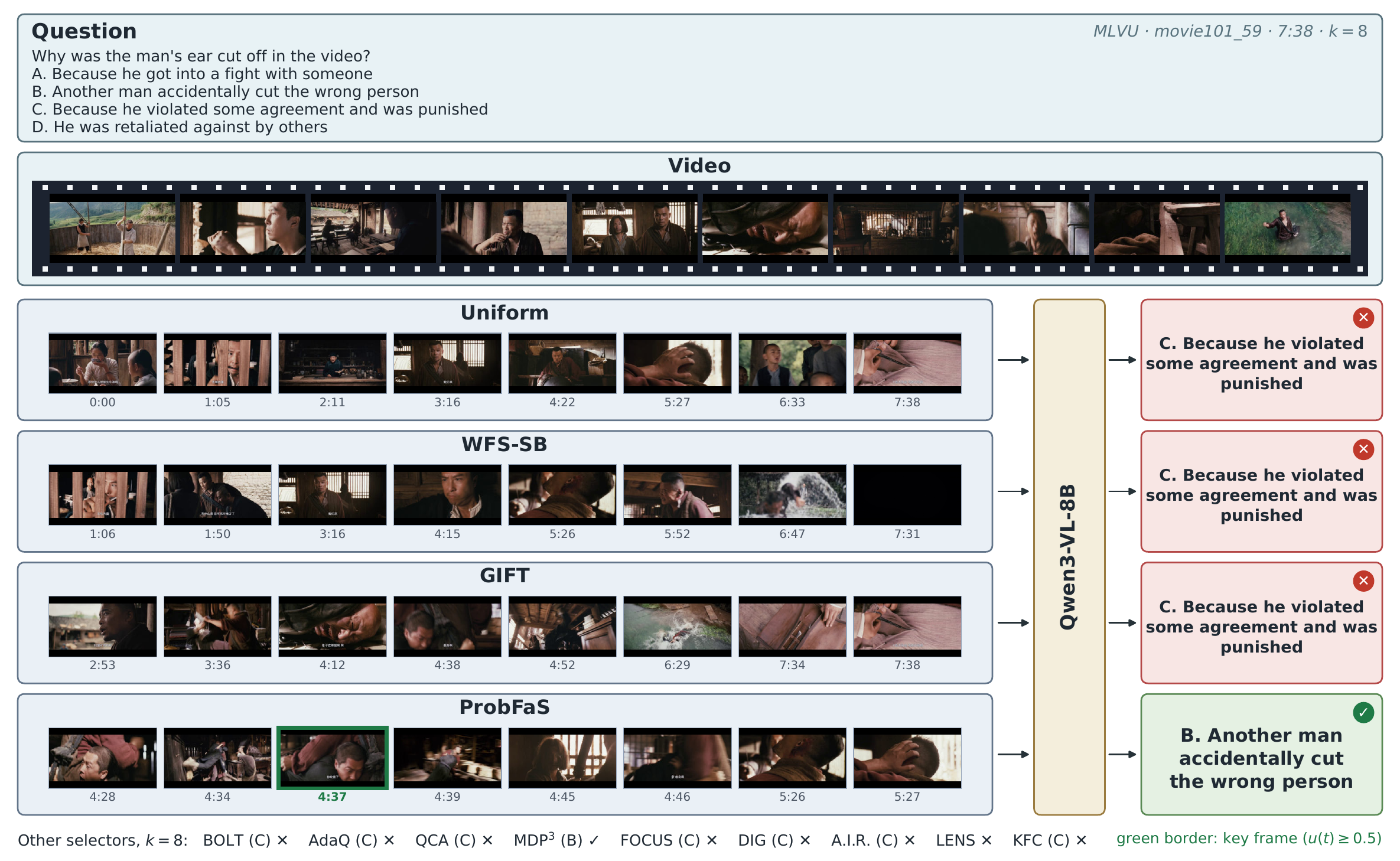}
    \caption{MLVU, plot QA, 8-min film excerpt, $k=8$.
    The evidence is a Chinese subtitle shown for about one second at 4:37 (``You cut the wrong one'').
    ProbFaS and MDP$^3$ select it and answer correctly.
    GIFT's frame 0.7\,s later shows only the next line (``Help!''), and every other selector answers wrongly.}
    \label{fig:qual_mlvu_plot}
    \vspace{2.5em}
    \includegraphics[width=\textwidth,height=0.4\textheight,keepaspectratio]{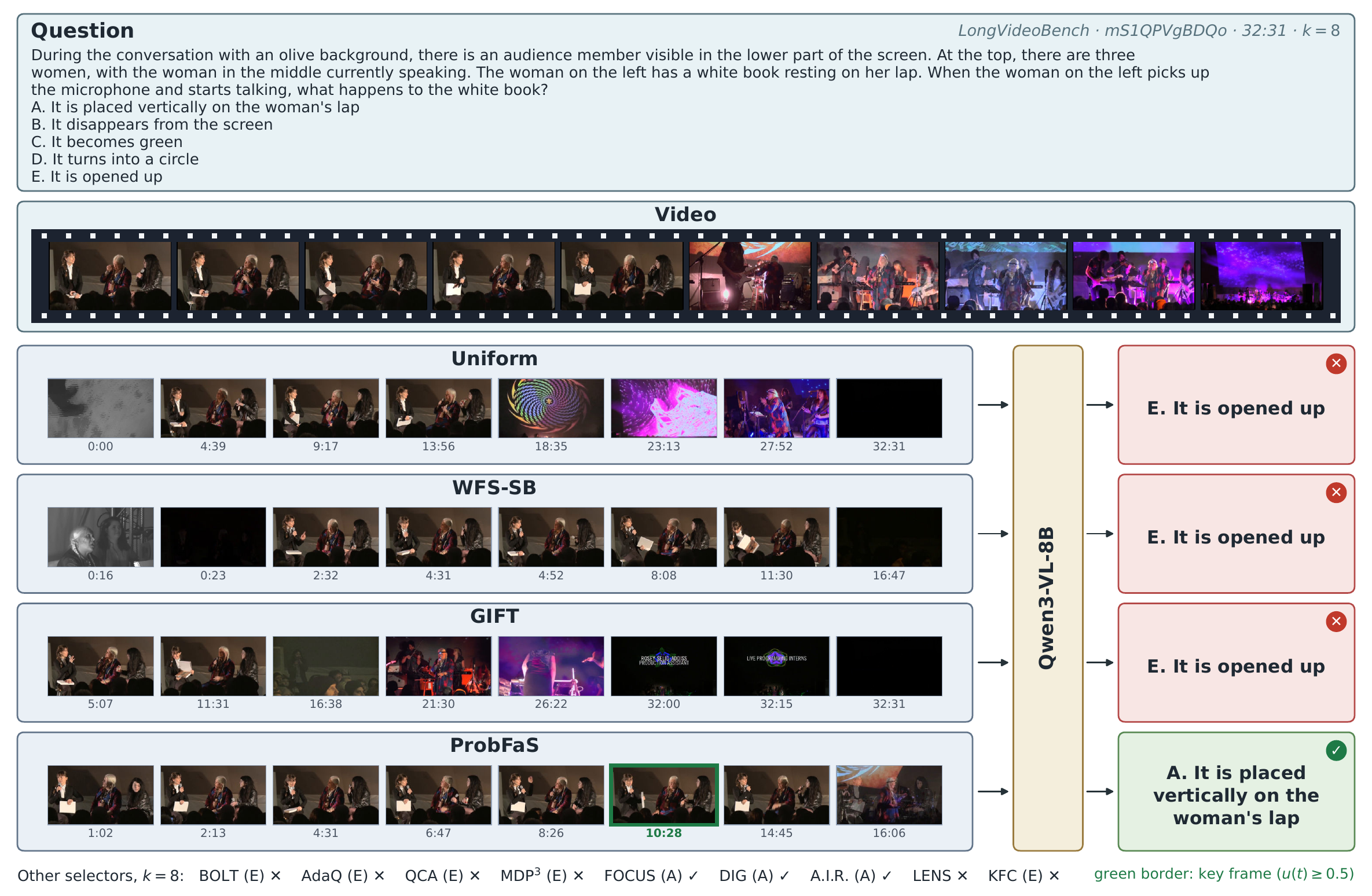}
    \caption{LongVideoBench, attribute change (SAA), 33-min panel recording, $k=8$.
    ProbFaS selects the moment the question refers to (10:28): the woman on the left has taken the microphone and holds the white book upright on her lap.
    Uniform sampling, WFS-SB and GIFT show the panel before or after this moment.
    Ten of the fourteen baselines with a $k=8$ result answer wrongly; FOCUS, DIG, A.I.R. and FRAG answer correctly.}
    \label{fig:qual_lvb_attr}
\end{figure*}

\Cref{fig:teaser_qual} in the main paper shows a Video-MME OCR question on which ProbFaS selects the key frame and the eight baselines scored below miss it.
\Cref{fig:qual_vmme_action,fig:qual_vmme_temporal,fig:qual_vmme_object,fig:qual_vmme_ocr,fig:qual_mlvu_plot,fig:qual_lvb_attr}, at the end of the supplementary material, add six more at $k=8$, each of a different question type: action reasoning, temporal reasoning, object recognition and OCR on Video-MME, plot QA on MLVU and attribute change on LongVideoBench, in videos of 70\,s to 42\,min.
We chose them among the questions that ProbFaS answers correctly while uniform sampling and most strong selectors answer wrongly.
Each figure shows the frames of uniform sampling, WFS-SB, GIFT and ProbFaS in temporal order and lists the answers of the other selectors below; LENS shows hyperframes, so it has no row.
All frames are embedded at their full resolution and can be enlarged.
To check that the selected frames explain the answers, we scored every frame selected by uniform sampling, BOLT, AdaQ, QCA, MDP$^3$, GIFT, KFC, WFS-SB and ProbFaS with the single-frame usefulness $u(t)$ of \cref{eq:usefulness}.
A green border marks a key frame, one from which Qwen3-VL-8B alone gives the correct answer with $u(t)\geq0.5$.
In all seven examples ProbFaS selects at least one key frame.
Of the eight baselines checked, only MDP$^3$ on the MLVU example and AdaQ and KFC on the long-jump example select one; MDP$^3$ and AdaQ answer correctly, KFC does not.
Most wrong answers are the answer the model gives without any frame: without the evidence, the selectors fall back on the language prior.

\begin{table*}[p] 
\centering
\caption{Accuracy (\%) by video duration on Video-MME, Qwen3-VL-8B, for every frame budget $k$. Video-MME durations: short $<$2\,min, medium 4--15\,min, long 30--60\,min. LENS runs its own pipeline (\cref{sec:setup}). Best in \textbf{bold}, second \underline{underlined}.}
\label{tab:suppl_dur_videomme}
\resizebox{0.75\textwidth}{!}{%
\begin{tabular}{lccccc ccccc ccccc}
\toprule
& \multicolumn{5}{c}{Short ($n{=}900$)} & \multicolumn{5}{c}{Medium ($n{=}900$)} & \multicolumn{5}{c}{Long ($n{=}900$)} \\
\cmidrule(lr){2-6}\cmidrule(lr){7-11}\cmidrule(lr){12-16}
Method $\backslash$ $k$ & 8 & 16 & 32 & 64 & 128 & 8 & 16 & 32 & 64 & 128 & 8 & 16 & 32 & 64 & 128 \\
\midrule
Uniform & 65.3 & 69.9 & 76.3 & 78.9 & \textbf{80.7} & 54.0 & 57.6 & 59.8 & 66.2 & 68.0 & 49.7 & 53.2 & 57.1 & 55.8 & 58.8 \\
BOLT & 69.4 & 73.4 & 76.0 & 78.8 & 79.0 & 54.4 & 59.6 & 63.1 & 65.9 & 69.2 & 51.0 & 51.9 & 55.0 & 56.8 & 59.1 \\
AKS & -- & -- & 77.7 & 78.7 & 79.1 & -- & -- & 63.0 & 66.4 & 69.8 & -- & -- & 56.1 & 58.0 & 59.8 \\
AdaQ & 67.2 & 71.0 & 76.8 & 79.4 & 79.6 & 53.4 & 60.3 & 60.8 & 65.2 & 67.8 & 50.1 & 52.8 & 54.8 & 57.2 & 58.7 \\
QCA & 71.0 & 72.3 & 76.9 & 78.9 & 79.7 & 57.1 & 58.3 & 63.4 & 66.6 & 69.2 & 52.0 & 54.0 & 56.3 & 58.1 & 60.0 \\
MDP$^3$ & 71.6 & 75.0 & \textbf{79.0} & \underline{80.1} & \underline{80.6} & 57.8 & 60.6 & 64.8 & 67.2 & 68.2 & 50.0 & 52.8 & 54.7 & 58.0 & 59.3 \\
FOCUS & 69.0 & 69.9 & 72.1 & 75.0 & 74.3 & 56.8 & 58.1 & 60.9 & 63.4 & 63.6 & 51.8 & 53.9 & 55.8 & 56.2 & 56.8 \\
DIG & 69.0 & 72.1 & 75.8 & 77.8 & 78.2 & 58.2 & 60.8 & 63.4 & 67.4 & 68.9 & 50.8 & 53.7 & 56.1 & 58.1 & 59.4 \\
A.I.R. & 71.6 & 72.6 & 74.9 & 74.8 & 75.1 & 59.0 & 62.8 & 61.8 & 63.1 & 63.6 & 51.3 & 52.9 & 52.2 & 53.6 & 53.6 \\
LENS & 72.2 & 75.3 & 75.6 & 78.2 & 80.0 & \underline{60.9} & \underline{63.8} & \underline{66.6} & \textbf{70.3} & \textbf{70.6} & \underline{54.1} & \textbf{57.2} & \textbf{58.8} & \underline{58.6} & \underline{61.9} \\
GIFT & \underline{72.7} & \textbf{77.1} & \textbf{79.0} & \textbf{80.9} & 80.3 & 58.1 & 60.4 & 64.3 & 67.2 & 68.2 & 51.9 & 54.0 & 55.9 & 57.4 & 58.3 \\
KFC & 70.3 & 73.2 & 75.3 & \underline{80.1} & 80.4 & 56.2 & 59.2 & 62.3 & 65.1 & 69.7 & 50.6 & 52.9 & 55.4 & 57.0 & 57.9 \\
WFS-SB & 71.8 & 74.9 & 78.4 & 78.9 & 79.7 & 60.0 & 62.1 & 64.6 & 68.9 & 69.6 & 52.3 & 54.4 & 55.7 & 56.4 & 59.8 \\
FRAG & 70.1 & 72.4 & 76.6 & 77.1 & 80.1 & 58.6 & 60.1 & 62.0 & 65.8 & 68.4 & 52.1 & 53.8 & 55.9 & 57.3 & 59.2 \\
OTFS & 71.3 & 72.7 & 77.7 & 78.3 & 79.1 & 59.2 & 61.6 & 65.8 & \underline{69.4} & 69.6 & 52.8 & \underline{56.4} & 56.7 & 58.2 & \textbf{62.7} \\
\midrule
ProbFaS & \textbf{74.0} & \underline{76.1} & \underline{78.8} & 79.2 & 79.9 & \textbf{62.8} & \textbf{65.4} & \textbf{67.8} & 69.2 & \underline{70.1} & \textbf{54.9} & 55.4 & \underline{57.2} & \textbf{59.0} & 61.2 \\
\bottomrule
\end{tabular}}
\end{table*}

\begin{table*}[p]
\centering
\caption{Accuracy (\%) by video duration on MLVU dev, Qwen3-VL-8B, for every frame budget $k$. MLVU cells are question accuracy within each duration bin (the official score is the mean over tasks, so it is not the mean of these cells). LENS runs its own pipeline (\cref{sec:setup}). Best in \textbf{bold}, second \underline{underlined}.}
\label{tab:suppl_dur_mlvu}
\resizebox{0.9\textwidth}{!}{%
\begin{tabular}{lccccc ccccc ccccc ccccc}
\toprule
& \multicolumn{5}{c}{3--7 min ($n{=}528$)} & \multicolumn{5}{c}{7--10 min ($n{=}1154$)} & \multicolumn{5}{c}{10--20 min ($n{=}354$)} & \multicolumn{5}{c}{$>$20 min ($n{=}138$)} \\
\cmidrule(lr){2-6}\cmidrule(lr){7-11}\cmidrule(lr){12-16}\cmidrule(lr){17-21}
Method $\backslash$ $k$ & 8 & 16 & 32 & 64 & 128 & 8 & 16 & 32 & 64 & 128 & 8 & 16 & 32 & 64 & 128 & 8 & 16 & 32 & 64 & 128 \\
\midrule
Uniform & 63.1 & 68.2 & 74.4 & 77.5 & 80.1 & 52.6 & 57.5 & 62.3 & 67.9 & 73.7 & 59.3 & 63.8 & 65.5 & 69.8 & 74.6 & 47.1 & 52.2 & 55.8 & 52.2 & 60.9 \\
BOLT & 68.8 & 72.5 & 72.2 & 78.6 & 79.9 & 62.7 & 67.9 & 72.1 & 74.9 & 77.1 & 67.8 & 70.6 & 74.6 & 77.4 & \underline{79.1} & 58.0 & 62.3 & 60.9 & 66.7 & 68.8 \\
AKS & -- & -- & 75.8 & 77.5 & 79.2 & -- & -- & 67.5 & 73.1 & 73.7 & -- & -- & 69.5 & 72.0 & 74.3 & -- & -- & 62.3 & 63.8 & 67.4 \\
AdaQ & 72.0 & 75.0 & 78.6 & 80.3 & \underline{82.0} & 65.3 & 67.9 & 72.4 & \underline{76.6} & \underline{78.5} & 70.6 & 72.9 & 75.7 & \underline{79.1} & 77.1 & 55.8 & 63.0 & 63.8 & 63.0 & 67.4 \\
QCA & \underline{75.6} & 76.1 & 79.2 & 79.4 & 80.9 & 63.9 & 66.4 & 72.9 & 76.5 & 76.8 & 68.9 & 68.9 & 72.3 & 75.7 & 74.9 & 57.2 & 58.0 & 63.0 & \textbf{71.0} & \underline{69.6} \\
MDP$^3$ & \textbf{76.3} & 75.4 & 78.0 & 80.7 & 79.7 & 66.1 & 68.2 & 72.6 & 72.7 & 74.0 & 66.1 & 71.2 & 70.9 & 70.6 & 74.0 & 55.8 & 56.5 & 55.8 & 58.7 & 58.0 \\
FOCUS & 72.9 & 75.6 & 78.0 & 76.9 & 77.3 & 66.1 & 70.0 & 72.7 & 72.4 & 74.8 & 70.3 & 73.7 & \underline{77.1} & 77.1 & 75.7 & 60.1 & 60.9 & 65.2 & 66.7 & \textbf{70.3} \\
DIG & 68.8 & 73.7 & 76.5 & 79.5 & 79.2 & 62.8 & 66.6 & 69.4 & 71.8 & 73.3 & 68.1 & 68.6 & 71.8 & 74.0 & 75.4 & 55.1 & 58.7 & 62.3 & 62.3 & \textbf{70.3} \\
A.I.R. & 71.2 & 77.3 & 75.8 & 77.1 & 76.9 & 68.9 & 70.9 & 72.7 & 72.8 & 72.1 & \textbf{72.6} & 72.9 & 73.7 & 74.0 & 72.6 & \underline{63.0} & \underline{65.2} & 65.2 & 65.2 & 68.8 \\
LENS & 73.5 & 73.7 & 76.5 & 78.8 & 81.1 & 65.9 & 65.9 & 69.5 & 69.8 & 73.9 & 70.1 & 71.5 & 73.2 & 74.3 & 75.4 & 59.4 & 60.9 & 58.0 & 65.2 & 62.3 \\
GIFT & 74.1 & \underline{77.7} & 78.6 & 78.8 & 80.5 & 66.2 & 71.6 & 73.0 & 73.2 & 73.7 & 67.8 & 70.3 & 72.6 & 74.9 & 74.6 & 55.8 & 60.1 & 58.0 & 61.6 & 56.5 \\
KFC & 75.0 & 76.7 & \underline{80.1} & 79.4 & 81.4 & 64.0 & 70.6 & 72.9 & 75.0 & 77.2 & 68.1 & \underline{74.0} & 76.0 & 76.8 & 77.4 & 57.2 & 62.3 & 59.4 & 60.1 & 61.6 \\
WFS-SB & 72.9 & 75.2 & 79.0 & \underline{81.4} & 81.6 & 67.6 & 72.3 & 76.0 & 76.0 & 77.4 & 71.2 & 72.0 & 75.4 & 77.4 & 78.0 & \underline{63.0} & 63.8 & 68.8 & 68.8 & \textbf{70.3} \\
FRAG & 71.8 & 75.8 & 79.2 & 79.7 & 80.9 & 68.3 & 70.8 & 72.0 & 72.7 & 75.4 & 69.8 & 70.9 & 74.6 & 75.4 & 76.8 & 58.7 & 57.2 & 62.3 & 60.1 & 61.6 \\
OTFS & 73.3 & 77.1 & 77.3 & \textbf{81.6} & 80.3 & \textbf{71.0} & \textbf{74.2} & \underline{76.2} & 76.2 & 78.0 & 72.0 & 72.0 & 75.1 & 77.4 & 78.0 & \textbf{65.2} & \textbf{70.3} & \underline{69.6} & \underline{70.3} & 68.8 \\
\midrule
ProbFaS & 75.4 & \textbf{79.2} & \textbf{80.3} & 81.2 & \textbf{82.6} & \underline{70.4} & \underline{72.9} & \textbf{78.3} & \textbf{79.1} & \textbf{78.6} & \underline{72.3} & \textbf{74.9} & \textbf{80.2} & \textbf{79.4} & \textbf{81.1} & 59.4 & 64.5 & \textbf{70.3} & 68.8 & \textbf{70.3} \\
\bottomrule
\end{tabular}}
\end{table*}

\begin{table*}[p]
\centering
\caption{Accuracy (\%) by video duration on LongVideoBench val, Qwen3-VL-8B, for every frame budget $k$. LongVideoBench duration groups as defined by the benchmark. LENS runs its own pipeline (\cref{sec:setup}). Best in \textbf{bold}, second \underline{underlined}.}
\label{tab:suppl_dur_lvb}
\resizebox{0.9\textwidth}{!}{%
\begin{tabular}{lccccc ccccc ccccc ccccc}
\toprule
& \multicolumn{5}{c}{8--15\,s ($n{=}189$)} & \multicolumn{5}{c}{15--60\,s ($n{=}172$)} & \multicolumn{5}{c}{3--10\,min ($n{=}412$)} & \multicolumn{5}{c}{15--60\,min ($n{=}564$)} \\
\cmidrule(lr){2-6}\cmidrule(lr){7-11}\cmidrule(lr){12-16}\cmidrule(lr){17-21}
Method $\backslash$ $k$ & 8 & 16 & 32 & 64 & 128 & 8 & 16 & 32 & 64 & 128 & 8 & 16 & 32 & 64 & 128 & 8 & 16 & 32 & 64 & 128 \\
\midrule
Uniform & 68.3 & 74.1 & \underline{77.2} & 79.4 & \textbf{79.9} & 66.9 & 70.9 & 75.6 & 76.7 & 76.2 & 57.0 & 57.5 & 58.7 & 63.8 & 64.3 & 45.7 & 48.8 & 51.2 & 53.2 & 56.9 \\
BOLT & \underline{70.4} & 72.5 & 72.0 & 73.0 & 73.0 & 70.9 & \underline{76.2} & 74.4 & 76.2 & 75.0 & 58.5 & 61.4 & 63.8 & 67.2 & 68.0 & 50.0 & 50.7 & 55.3 & 56.9 & 58.7 \\
AKS & -- & -- & 70.4 & 70.4 & 70.4 & -- & -- & 72.1 & 75.0 & 75.0 & -- & -- & 60.9 & 66.3 & 67.2 & -- & -- & 55.7 & 58.7 & 58.3 \\
AdaQ & 69.3 & 70.9 & 70.9 & 70.9 & 70.9 & 66.9 & 71.5 & 72.1 & 72.7 & 72.7 & 60.0 & 60.4 & 65.5 & \underline{67.5} & 67.0 & 52.3 & 54.6 & 56.6 & 58.7 & 60.3 \\
QCA & \textbf{71.4} & \textbf{75.7} & \textbf{78.8} & \underline{79.9} & \underline{79.4} & 69.8 & 68.0 & 76.7 & 77.9 & 76.2 & 59.2 & 60.0 & 65.0 & 65.3 & 68.4 & 52.0 & 55.1 & 57.8 & 58.9 & 58.3 \\
MDP$^3$ & \textbf{71.4} & 73.0 & 76.2 & 78.8 & \textbf{79.9} & 71.5 & 73.3 & \textbf{79.1} & \textbf{79.1} & 76.2 & 60.0 & 62.4 & 62.1 & 64.1 & 63.6 & 50.9 & 52.3 & 55.0 & 53.5 & 55.0 \\
FOCUS & 67.7 & 70.9 & 70.9 & 73.0 & 75.7 & 66.9 & 69.8 & 72.7 & 77.3 & 76.7 & 61.7 & 61.9 & 66.3 & 65.0 & 65.5 & 53.9 & 56.6 & 58.7 & 58.3 & 60.1 \\
DIG & 69.3 & 72.5 & 72.0 & 77.2 & 77.8 & 70.9 & 68.6 & 70.9 & 70.9 & 71.5 & 59.7 & 64.6 & 65.8 & \underline{67.5} & \underline{68.9} & 53.2 & 54.8 & 58.2 & 61.3 & 61.2 \\
A.I.R. & \underline{70.4} & 70.9 & 71.4 & 72.0 & 72.0 & 73.3 & \underline{76.2} & 76.2 & 77.9 & \underline{77.3} & 63.3 & 64.1 & 62.4 & 63.8 & 63.6 & 55.1 & 56.9 & 56.2 & 55.7 & 56.7 \\
LENS & 67.7 & \underline{74.6} & 76.2 & 75.1 & 76.7 & 72.1 & \underline{76.2} & 72.1 & 75.0 & 75.0 & 62.4 & \underline{65.8} & 61.7 & 65.5 & 65.5 & 52.0 & 56.0 & 57.1 & 59.0 & 58.5 \\
GIFT & \underline{70.4} & \underline{74.6} & 76.7 & \textbf{80.4} & \textbf{79.9} & 68.6 & 70.9 & \underline{77.9} & \textbf{79.1} & \textbf{77.9} & 60.4 & \underline{65.8} & \underline{68.4} & \underline{67.5} & 64.3 & 51.2 & 51.8 & 52.1 & 56.2 & 57.3 \\
KFC & \underline{70.4} & 73.5 & 76.7 & 79.4 & \underline{79.4} & 68.6 & 75.6 & 76.2 & \textbf{79.1} & 76.2 & 57.8 & 62.4 & 65.0 & 66.7 & 65.0 & 47.7 & 50.9 & 55.5 & 58.3 & 59.4 \\
WFS-SB & \textbf{71.4} & \textbf{75.7} & 76.2 & 78.8 & \underline{79.4} & \textbf{75.0} & 73.3 & 77.3 & 77.9 & 76.2 & 63.1 & 64.3 & 68.2 & 66.7 & 66.7 & \underline{56.2} & 57.8 & \underline{59.8} & \underline{62.1} & \textbf{64.4} \\
FRAG & \underline{70.4} & 69.8 & 74.1 & 72.5 & 77.8 & 66.9 & 68.6 & 70.9 & 72.7 & 73.8 & 63.6 & 63.1 & 62.1 & 66.7 & 67.2 & 55.9 & 58.2 & 58.3 & 59.2 & 58.7 \\
OTFS & 69.3 & 70.4 & 70.4 & 70.4 & 70.4 & \underline{74.4} & \textbf{76.7} & 75.6 & 75.0 & 75.0 & \textbf{68.0} & 65.5 & 67.7 & 66.0 & \textbf{69.2} & 55.3 & \underline{58.3} & 59.0 & 59.8 & \underline{63.1} \\
\midrule
ProbFaS & 69.8 & 74.1 & 76.2 & 79.4 & \underline{79.4} & 72.7 & 75.6 & \textbf{79.1} & \underline{78.5} & \textbf{77.9} & \underline{66.0} & \textbf{69.4} & \textbf{71.1} & \textbf{69.2} & 68.2 & \textbf{59.9} & \textbf{61.5} & \textbf{62.4} & \textbf{62.6} & 62.5 \\
\bottomrule
\end{tabular}}
\end{table*}

\begin{table*}[t]
\centering
\caption{Accuracy (\%) per task on MLVU dev, Qwen3-VL-8B. M-avg is the official score, the mean of the seven task accuracies. Best per column and budget in \textbf{bold}, second \underline{underlined}.}
\label{tab:suppl_mlvu_tasks}
\small
\setlength{\tabcolsep}{4pt}
\resizebox{!}{0.47\textheight}{%
\begin{tabular}{llcccccccc}
\toprule
$k$ & Method & Anomaly & Count & Ego & Needle & Order & PlotQA & Topic & M-avg \\
\midrule
\multirow{15}{*}{8} & Uniform & 68.0 & 19.9 & 51.4 & 58.3 & 46.7 & 56.6 & 85.2 & 55.2 \\
 & BOLT & 66.0 & 38.3 & 54.8 & 79.2 & 49.0 & 66.8 & \textbf{89.4} & 63.4 \\
 & AdaQ & 72.0 & 38.8 & 56.0 & 81.7 & 54.4 & 71.1 & 85.9 & 65.7 \\
 & QCA & 70.0 & 32.0 & 63.6 & 74.1 & 59.8 & 72.2 & 84.4 & 65.2 \\
 & MDP$^3$ & 70.0 & 32.5 & 59.7 & 76.1 & 62.2 & 74.0 & 87.5 & 66.0 \\
 & FOCUS & 67.5 & 37.9 & 64.5 & 81.1 & 59.5 & 71.1 & 81.7 & 66.2 \\
 & DIG & 66.5 & 26.7 & 57.1 & 75.8 & 53.7 & 69.8 & \underline{88.2} & 62.5 \\
 & A.I.R. & 64.0 & \textbf{50.0} & 65.6 & 81.1 & 54.4 & 74.4 & 84.8 & 67.8 \\
 & LENS & 67.5 & 45.6 & 58.0 & 76.9 & 59.1 & 73.8 & 84.4 & 66.5 \\
 & GIFT & 70.5 & 37.9 & 61.6 & 72.7 & \textbf{66.8} & 69.9 & 86.7 & 66.6 \\
 & KFC & \underline{73.5} & 33.5 & 58.0 & 77.7 & \underline{65.6} & 67.9 & 84.4 & 65.8 \\
 & WFS-SB & 69.0 & 36.4 & 61.9 & \textbf{83.4} & 59.1 & \underline{75.3} & 82.9 & 66.9 \\
 & FRAG & 70.5 & 42.7 & \textbf{67.9} & 81.4 & 45.6 & 75.1 & 81.7 & 66.4 \\
 & OTFS & 72.0 & 40.3 & \underline{67.3} & 81.7 & 63.3 & \textbf{76.3} & 84.4 & \underline{69.3} \\
 & ProbFaS & \textbf{75.0} & \underline{49.5} & 62.8 & \underline{82.8} & 56.0 & 75.1 & 87.8 & \textbf{69.9} \\
\midrule
\multirow{15}{*}{16} & Uniform & 70.5 & 20.4 & 57.7 & 64.5 & 50.2 & 64.0 & 87.8 & 59.3 \\
 & BOLT & 68.5 & 44.7 & 61.1 & \underline{82.8} & 56.0 & 72.0 & 88.2 & 67.6 \\
 & AdaQ & 71.0 & 43.7 & 58.8 & 82.3 & 63.7 & 74.2 & 87.1 & 68.7 \\
 & QCA & 74.5 & 35.0 & 65.1 & 72.1 & 62.2 & 73.7 & 86.7 & 67.0 \\
 & MDP$^3$ & 72.5 & 36.9 & 63.9 & 76.1 & 65.6 & 74.0 & 87.5 & 68.1 \\
 & FOCUS & 73.0 & 41.7 & 64.2 & 79.7 & 64.1 & 77.6 & 86.3 & 69.5 \\
 & DIG & 71.0 & 28.6 & 63.4 & 78.3 & 55.2 & 74.6 & \textbf{89.0} & 65.7 \\
 & A.I.R. & 72.0 & 47.6 & 66.8 & 81.4 & 62.9 & 78.3 & 84.8 & 70.5 \\
 & LENS & 69.0 & 42.7 & 59.1 & 75.5 & 60.2 & 73.7 & 88.2 & 66.9 \\
 & GIFT & 72.5 & 48.1 & 64.2 & 77.2 & \underline{70.7} & 76.1 & 87.8 & 70.9 \\
 & KFC & \textbf{77.5} & 43.7 & 63.1 & 78.3 & \textbf{81.9} & 71.4 & 85.9 & 71.7 \\
 & WFS-SB & 72.0 & 46.6 & 67.0 & 82.5 & 66.4 & 77.0 & 82.9 & 70.6 \\
 & FRAG & 74.0 & \textbf{53.4} & 68.5 & 78.6 & 45.9 & \underline{79.0} & 85.2 & 69.2 \\
 & OTFS & 70.5 & 46.6 & \textbf{69.3} & \underline{82.8} & 70.3 & \textbf{80.9} & 84.4 & \underline{72.1} \\
 & ProbFaS & \underline{75.5} & \underline{51.0} & \underline{68.8} & \textbf{83.4} & 62.5 & 78.7 & \underline{88.6} & \textbf{72.6} \\
\midrule
\multirow{16}{*}{32} & Uniform & 73.5 & 30.6 & 61.4 & 71.8 & 50.6 & 69.8 & 88.6 & 63.7 \\
 & BOLT & 72.0 & 52.4 & 63.6 & 82.3 & 60.2 & 75.3 & 87.8 & 70.5 \\
 & AKS & 73.0 & 45.6 & 63.4 & 75.8 & 54.4 & 74.2 & \textbf{90.5} & 68.1 \\
 & AdaQ & 74.5 & 48.5 & 63.9 & 82.5 & 70.3 & 78.8 & 88.6 & 72.5 \\
 & QCA & 77.5 & 48.5 & 67.9 & 78.9 & 69.9 & 77.4 & 87.5 & 72.5 \\
 & MDP$^3$ & \underline{78.5} & 45.6 & 65.3 & 76.9 & 67.2 & 76.6 & \underline{90.1} & 71.5 \\
 & FOCUS & 76.0 & 47.1 & 70.2 & 80.3 & 71.0 & 78.3 & 86.3 & 72.7 \\
 & DIG & 74.0 & 35.4 & 65.3 & 79.4 & 60.2 & 78.1 & 89.4 & 68.8 \\
 & A.I.R. & 70.0 & 49.5 & 67.3 & 81.4 & 68.3 & 77.7 & 85.9 & 71.5 \\
 & LENS & 71.5 & 45.1 & 63.1 & 74.9 & 61.4 & 79.4 & 89.0 & 69.2 \\
 & GIFT & 77.0 & 50.5 & 64.5 & 79.2 & 71.8 & 76.4 & 87.5 & 72.4 \\
 & KFC & 76.5 & 48.1 & 63.4 & 80.0 & \textbf{79.2} & 77.6 & 88.6 & 73.3 \\
 & WFS-SB & 76.0 & \textbf{59.7} & 69.3 & \textbf{83.9} & 69.9 & 80.0 & 86.3 & \underline{75.0} \\
 & FRAG & \textbf{79.5} & 55.3 & \underline{70.5} & 77.7 & 55.6 & 79.2 & 87.8 & 72.2 \\
 & OTFS & 69.5 & 56.8 & 69.3 & 82.5 & 71.4 & \underline{81.1} & 89.0 & 74.2 \\
 & ProbFaS & 76.0 & \underline{59.2} & \textbf{72.2} & \underline{83.4} & \underline{78.4} & \textbf{83.7} & 87.8 & \textbf{77.2} \\
\midrule
\multirow{16}{*}{64} & Uniform & 77.0 & 37.9 & 61.9 & 73.8 & 59.8 & 75.5 & 90.5 & 68.1 \\
 & BOLT & 74.0 & 57.3 & 67.0 & \textbf{83.1} & 65.6 & 82.2 & 89.4 & 74.1 \\
 & AKS & 75.5 & 57.8 & \underline{71.0} & 76.3 & 56.4 & 78.5 & 89.7 & 72.2 \\
 & AdaQ & 76.5 & 59.2 & 65.1 & 82.3 & 75.7 & \textbf{83.3} & 89.0 & 75.9 \\
 & QCA & 77.5 & \underline{60.7} & 69.9 & 80.8 & 71.8 & 80.0 & 90.5 & 75.9 \\
 & MDP$^3$ & \underline{79.5} & 49.0 & 64.8 & 77.7 & 62.9 & 80.0 & 90.5 & 72.1 \\
 & FOCUS & 74.5 & 53.9 & 67.0 & 80.3 & 69.9 & 77.0 & 87.5 & 72.9 \\
 & DIG & 76.5 & 35.9 & 69.0 & 81.1 & 62.5 & 81.6 & 89.7 & 70.9 \\
 & A.I.R. & 71.0 & 49.5 & 66.2 & 82.3 & 69.1 & 78.7 & 86.3 & 71.9 \\
 & LENS & 78.0 & 43.7 & 66.5 & 76.6 & 64.5 & 78.3 & 89.0 & 70.9 \\
 & GIFT & 78.5 & 51.5 & 65.1 & 77.7 & 69.9 & 78.3 & \textbf{91.3} & 73.2 \\
 & KFC & \textbf{80.5} & 51.0 & 67.6 & 78.6 & \underline{76.4} & 78.7 & 89.4 & 74.6 \\
 & WFS-SB & 77.5 & 59.7 & 69.9 & \textbf{83.1} & 68.7 & \underline{82.9} & 88.2 & 75.7 \\
 & FRAG & 77.5 & 54.4 & 67.9 & 78.0 & 57.1 & 81.6 & \underline{90.9} & 72.5 \\
 & OTFS & \underline{79.5} & 55.8 & 70.2 & \underline{82.8} & 74.5 & 80.9 & 90.1 & \underline{76.3} \\
 & ProbFaS & 76.5 & \textbf{64.6} & \textbf{71.9} & \underline{82.8} & \textbf{79.5} & 82.4 & 89.4 & \textbf{78.1} \\
\midrule
\multirow{16}{*}{128} & Uniform & 78.5 & 48.5 & 66.2 & 79.2 & 66.0 & 80.9 & \textbf{92.4} & 73.1 \\
 & BOLT & 74.5 & 59.2 & 70.5 & \textbf{84.8} & 71.0 & 83.5 & 88.6 & 76.0 \\
 & AKS & 79.5 & \underline{60.2} & 67.6 & 79.4 & 62.2 & 77.9 & 91.3 & 74.0 \\
 & AdaQ & 80.0 & 57.8 & \underline{71.3} & 82.0 & 74.5 & 83.7 & 91.3 & \underline{77.2} \\
 & QCA & 79.0 & 57.8 & 71.0 & 82.3 & 69.9 & 81.4 & 89.4 & 75.8 \\
 & MDP$^3$ & 80.0 & 49.0 & 66.8 & 77.7 & 70.3 & 78.7 & 90.9 & 73.3 \\
 & FOCUS & 72.0 & 55.3 & 68.8 & 82.5 & 71.0 & 80.1 & 86.3 & 73.7 \\
 & DIG & 77.0 & 38.3 & 69.9 & \underline{83.9} & 64.1 & 82.9 & 90.5 & 72.4 \\
 & A.I.R. & 72.0 & 48.1 & 68.2 & 80.8 & 67.6 & 77.7 & 85.9 & 71.5 \\
 & LENS & 75.5 & 46.1 & 66.8 & 79.2 & 70.7 & \underline{83.9} & 90.1 & 73.2 \\
 & GIFT & 78.5 & 47.6 & 65.1 & 79.4 & 67.6 & 80.7 & 91.6 & 72.9 \\
 & KFC & \textbf{82.0} & 54.4 & 69.0 & 79.4 & \underline{76.4} & 82.0 & 90.9 & 76.3 \\
 & WFS-SB & 80.0 & \underline{60.2} & \textbf{71.9} & 81.4 & 71.4 & 83.7 & 89.4 & 76.8 \\
 & FRAG & \underline{80.5} & 55.3 & 71.0 & 78.6 & 66.4 & 80.9 & \underline{92.0} & 75.0 \\
 & OTFS & 77.5 & 57.8 & \underline{71.3} & 80.8 & 74.9 & 83.5 & 90.9 & 76.7 \\
 & ProbFaS & \underline{80.5} & \textbf{61.7} & 70.5 & 82.8 & \textbf{78.4} & \textbf{84.4} & 90.9 & \textbf{78.4} \\
\bottomrule
\end{tabular}}
\end{table*}

\begin{table*}[t]
\centering
\caption{Accuracy (\%) per Video-MME task category, Qwen3-VL-8B, for every frame budget $k$ (number of questions in brackets). Best per column and budget in \textbf{bold}, second \underline{underlined}.}
\label{tab:suppl_vmme_cats}
\small
\setlength{\tabcolsep}{4pt}
\resizebox{!}{0.4\textheight}{%
\begin{tabular}{llcccccccccccc}
\toprule
$k$ & Method & Act.\ reas. (285) & Act.\ rec. (313) & Attr. (222) & Count (268) & Synopsis (323) & OCR (139) & Obj.\ reas. (454) & Obj.\ rec. (354) & Spat.\ perc. (54) & Spat.\ reas. (56) & Temp.\ perc. (55) & Temp.\ reas. (177) \\
\midrule
\multirow{15}{*}{8} & Uniform & 51.6 & 53.4 & 66.7 & 35.4 & 71.2 & 53.2 & 55.3 & 61.0 & 53.7 & 73.2 & 63.6 & \underline{49.7} \\
 & BOLT & 54.0 & 53.4 & 75.2 & 34.7 & 71.5 & 66.2 & 55.9 & 62.4 & 63.0 & 73.2 & 70.9 & 45.8 \\
 & AdaQ & 50.9 & 52.4 & 75.2 & 32.1 & 73.1 & 66.9 & 55.1 & 59.6 & 63.0 & 75.0 & 58.2 & 43.5 \\
 & QCA & 55.1 & 54.3 & 77.9 & \underline{40.3} & 74.3 & 62.6 & 58.1 & 63.0 & 66.7 & 76.8 & 70.9 & 45.8 \\
 & MDP$^3$ & 56.1 & 57.8 & 75.2 & 35.1 & 74.9 & 66.9 & 56.2 & 61.9 & \underline{72.2} & 76.8 & 63.6 & 48.6 \\
 & FOCUS & \textbf{57.5} & 55.9 & 77.0 & 36.6 & 74.0 & 64.7 & 55.1 & 64.7 & 57.4 & \underline{78.6} & 58.2 & 42.4 \\
 & DIG & 53.3 & 56.5 & 74.3 & 36.2 & 71.5 & 67.6 & 58.4 & 65.0 & 59.3 & 75.0 & \underline{72.7} & 43.5 \\
 & A.I.R. & 54.0 & \underline{60.4} & 74.3 & 36.9 & 71.5 & 70.5 & 58.1 & 67.2 & \textbf{74.1} & \textbf{80.4} & 70.9 & 42.4 \\
 & LENS & 56.5 & \textbf{60.7} & 75.7 & 38.8 & \textbf{77.1} & 68.3 & 61.2 & 67.5 & 68.5 & 75.0 & 65.5 & 48.6 \\
 & GIFT & \underline{57.2} & 56.5 & 76.1 & 35.4 & \underline{76.5} & 71.2 & 57.9 & 65.3 & 70.4 & \underline{78.6} & \underline{72.7} & 44.1 \\
 & KFC & 55.4 & 58.8 & 73.9 & 34.0 & 73.7 & 65.5 & 58.1 & 62.7 & 70.4 & 75.0 & 61.8 & 38.4 \\
 & WFS-SB & 55.1 & 60.1 & 75.2 & 37.7 & 75.2 & 66.9 & \underline{61.7} & 65.5 & 63.0 & 76.8 & 67.3 & 46.3 \\
 & FRAG & 53.0 & 57.2 & \textbf{80.6} & 35.8 & 71.2 & \textbf{75.5} & 57.3 & \underline{67.8} & 70.4 & 69.6 & 65.5 & 41.8 \\
 & OTFS & 56.1 & 56.5 & 75.7 & 38.8 & 71.2 & \underline{72.7} & 61.2 & 66.7 & 66.7 & 73.2 & 63.6 & 47.5 \\
 & ProbFaS & 55.8 & 59.7 & \underline{78.8} & \textbf{43.7} & 74.9 & 71.9 & \textbf{62.1} & \textbf{68.9} & \underline{72.2} & \textbf{80.4} & \textbf{74.5} & \textbf{53.1} \\
\midrule
\multirow{15}{*}{16} & Uniform & 58.2 & 54.0 & 74.8 & 41.0 & 76.2 & 58.3 & 59.5 & 61.0 & 59.3 & 78.6 & 67.3 & 50.3 \\
 & BOLT & 55.8 & 58.8 & 76.1 & 39.6 & 74.3 & 62.6 & 62.6 & 67.2 & 66.7 & 75.0 & 72.7 & 44.6 \\
 & AdaQ & 53.0 & 56.5 & \textbf{82.4} & 40.3 & 75.9 & 64.7 & 58.1 & 65.3 & \textbf{74.1} & 73.2 & 70.9 & 49.7 \\
 & QCA & 56.1 & 53.7 & 79.7 & \underline{42.2} & 76.8 & 65.5 & 60.1 & 64.1 & 63.0 & 73.2 & 72.7 & 50.8 \\
 & MDP$^3$ & 57.2 & 62.0 & 79.3 & 41.0 & 76.2 & 68.3 & 61.2 & 63.3 & \textbf{74.1} & 80.4 & 67.3 & 49.2 \\
 & FOCUS & 54.7 & 54.0 & 77.5 & 38.4 & 76.8 & 69.1 & 59.7 & 65.5 & 59.3 & 76.8 & 65.5 & 44.6 \\
 & DIG & 58.6 & 58.8 & 77.9 & 40.3 & 74.9 & 69.1 & 59.9 & 67.2 & 63.0 & 78.6 & 69.1 & 46.9 \\
 & A.I.R. & 56.8 & \underline{63.6} & 79.3 & 36.9 & 73.7 & 71.9 & \underline{63.2} & 67.2 & \underline{72.2} & \underline{82.1} & 63.6 & 42.4 \\
 & LENS & \textbf{59.6} & 62.0 & \underline{80.6} & \textbf{44.4} & \textbf{79.3} & 72.7 & \underline{63.2} & \underline{71.2} & 68.5 & 78.6 & \textbf{78.2} & 48.0 \\
 & GIFT & 54.7 & 59.7 & 77.5 & 40.7 & \underline{78.9} & 70.5 & 61.7 & 68.9 & \underline{72.2} & \textbf{83.9} & \underline{76.4} & \textbf{53.7} \\
 & KFC & 57.5 & 60.7 & 75.7 & 39.6 & 74.3 & 68.3 & 62.3 & 66.1 & 64.8 & 76.8 & 63.6 & 42.4 \\
 & WFS-SB & 57.5 & 59.7 & 77.0 & 39.6 & 76.5 & 71.2 & \textbf{63.7} & \textbf{72.6} & 61.1 & 78.6 & 65.5 & 50.8 \\
 & FRAG & 54.0 & 56.9 & \underline{80.6} & 41.0 & 74.3 & \underline{74.8} & 59.3 & 69.8 & 68.5 & 73.2 & 67.3 & 45.8 \\
 & OTFS & \underline{58.9} & 60.1 & \underline{80.6} & 39.6 & 76.2 & 70.5 & 63.0 & 67.8 & 68.5 & 73.2 & 65.5 & 51.4 \\
 & ProbFaS & 57.5 & \textbf{64.9} & \underline{80.6} & \underline{42.2} & 78.3 & \textbf{77.7} & 63.0 & 70.3 & 68.5 & 80.4 & \underline{76.4} & \underline{53.1} \\
\midrule
\multirow{16}{*}{32} & Uniform & 59.6 & 62.3 & 78.4 & 43.3 & 78.9 & 73.4 & 62.3 & 68.1 & 66.7 & \underline{80.4} & 74.5 & 45.8 \\
 & BOLT & 57.5 & 58.8 & 79.3 & 46.3 & 79.3 & 69.8 & 63.0 & 71.5 & 64.8 & 73.2 & \underline{76.4} & 50.3 \\
 & AKS & 57.2 & 62.9 & 79.7 & 44.0 & \textbf{80.2} & 75.5 & 63.2 & 69.8 & 66.7 & \textbf{82.1} & 74.5 & 53.7 \\
 & AdaQ & 55.8 & 61.0 & \underline{82.0} & 42.9 & 77.7 & 74.1 & 61.7 & 70.9 & \underline{72.2} & 73.2 & 65.5 & 46.9 \\
 & QCA & 60.7 & 62.3 & 78.4 & \underline{47.4} & 78.0 & 72.7 & 63.0 & 67.5 & 70.4 & 76.8 & \textbf{78.2} & 55.9 \\
 & MDP$^3$ & \underline{61.4} & 64.2 & 78.4 & 45.5 & 79.3 & 73.4 & 64.3 & 69.2 & 70.4 & 78.6 & \textbf{78.2} & 53.1 \\
 & FOCUS & 60.4 & 61.3 & 80.6 & 41.0 & 75.2 & 69.8 & 62.1 & 65.8 & 63.0 & 69.6 & 63.6 & 46.9 \\
 & DIG & 57.9 & 62.6 & 80.6 & 42.5 & 77.4 & 76.3 & 64.8 & 70.1 & 70.4 & \textbf{82.1} & 70.9 & 46.9 \\
 & A.I.R. & 58.6 & 62.6 & 77.5 & 37.7 & 73.1 & 72.7 & 61.9 & 71.2 & 66.7 & \textbf{82.1} & 61.8 & 44.1 \\
 & LENS & \textbf{62.1} & 64.2 & 78.4 & 45.9 & \underline{79.6} & \underline{77.0} & 64.8 & 70.6 & 68.5 & 76.8 & \underline{76.4} & \underline{58.2} \\
 & GIFT & 58.9 & 64.2 & 78.8 & \underline{47.4} & 78.3 & 74.1 & 63.4 & 71.5 & \textbf{74.1} & 76.8 & 74.5 & 57.1 \\
 & KFC & 57.2 & 63.3 & 76.6 & 44.0 & 76.8 & 71.9 & 62.3 & 70.6 & 70.4 & 76.8 & \textbf{78.2} & 47.5 \\
 & WFS-SB & 58.9 & \textbf{66.1} & 78.8 & 41.4 & 78.0 & 74.8 & 64.8 & \underline{72.9} & 66.7 & \underline{80.4} & 74.5 & 54.8 \\
 & FRAG & 56.8 & 62.6 & 81.1 & 45.9 & 77.1 & \underline{77.0} & 62.1 & \underline{72.9} & 66.7 & 69.6 & 63.6 & 46.9 \\
 & OTFS & 56.8 & 65.5 & 79.7 & \textbf{48.1} & 78.6 & 74.1 & \underline{65.2} & \underline{72.9} & 70.4 & 76.8 & \underline{76.4} & 53.1 \\
 & ProbFaS & 58.6 & \underline{65.8} & \textbf{82.9} & 44.8 & 78.0 & \textbf{82.0} & \textbf{65.4} & \textbf{75.7} & 68.5 & \textbf{82.1} & 70.9 & \textbf{58.8} \\
\midrule
\multirow{16}{*}{64} & Uniform & 60.7 & 65.2 & 79.7 & 46.3 & 79.3 & 72.7 & 63.2 & 72.3 & \underline{75.9} & \underline{82.1} & 80.0 & 55.9 \\
 & BOLT & 58.9 & 64.5 & 81.1 & 45.9 & 77.7 & 75.5 & 64.5 & 75.1 & 74.1 & 73.2 & 76.4 & 57.6 \\
 & AKS & \underline{61.1} & 64.5 & \underline{82.4} & \underline{48.5} & 79.3 & 77.0 & 63.7 & 72.6 & 70.4 & 80.4 & 78.2 & 58.8 \\
 & AdaQ & \underline{61.1} & 66.5 & \underline{82.4} & 46.6 & 78.6 & 74.1 & 62.6 & 72.6 & 66.7 & 80.4 & \underline{81.8} & 58.2 \\
 & QCA & \underline{61.1} & 65.5 & 80.6 & 47.4 & \textbf{81.7} & 74.1 & 62.3 & 71.8 & 70.4 & 80.4 & \textbf{83.6} & \textbf{64.4} \\
 & MDP$^3$ & 60.0 & 65.2 & 82.0 & 47.0 & 80.8 & 73.4 & 65.4 & 74.0 & \underline{75.9} & \underline{82.1} & \underline{81.8} & 62.7 \\
 & FOCUS & 59.3 & 61.7 & 79.7 & 44.4 & 78.9 & 69.8 & 62.8 & 72.0 & 63.0 & 73.2 & 61.8 & 52.5 \\
 & DIG & 58.2 & 65.2 & \textbf{84.7} & 47.8 & \underline{81.4} & 76.3 & \underline{66.7} & 72.9 & 72.2 & \underline{82.1} & 70.9 & 50.8 \\
 & A.I.R. & 57.5 & 63.9 & 79.3 & 40.3 & 74.9 & 71.9 & 63.7 & 70.3 & 66.7 & \underline{82.1} & 63.6 & 44.1 \\
 & LENS & 60.7 & 66.5 & 79.3 & 46.6 & 81.1 & 78.4 & 66.1 & 74.9 & \textbf{81.5} & 78.6 & \textbf{83.6} & \underline{63.3} \\
 & GIFT & \underline{61.1} & \textbf{68.4} & 80.2 & \textbf{48.9} & 77.7 & 76.3 & 65.9 & \underline{75.4} & 72.2 & \underline{82.1} & 76.4 & 58.2 \\
 & KFC & 58.6 & 65.2 & 77.5 & 47.8 & 79.3 & 74.8 & 65.2 & 74.0 & 74.1 & \textbf{83.9} & 76.4 & 57.6 \\
 & WFS-SB & 58.6 & 65.5 & 81.5 & 44.8 & 78.6 & \underline{80.6} & \textbf{67.0} & 72.3 & \underline{75.9} & 80.4 & \underline{81.8} & 61.0 \\
 & FRAG & 57.9 & 63.6 & 80.2 & 47.0 & 79.3 & \textbf{83.5} & 63.7 & 73.2 & 68.5 & 75.0 & 70.9 & 54.2 \\
 & OTFS & 60.0 & \underline{67.1} & 80.6 & 47.0 & 81.1 & 77.7 & 66.5 & 74.6 & 72.2 & 78.6 & 80.0 & 59.3 \\
 & ProbFaS & \textbf{63.5} & 66.5 & \underline{82.4} & 47.0 & 78.9 & 78.4 & 66.5 & \textbf{75.7} & 74.1 & 80.4 & 78.2 & 60.5 \\
\midrule
\multirow{16}{*}{128} & Uniform & 62.1 & 68.4 & \textbf{85.1} & 49.6 & 78.3 & 76.3 & 66.5 & 75.1 & 68.5 & 80.4 & \textbf{83.6} & 55.9 \\
 & BOLT & 60.7 & 66.8 & 81.1 & 47.4 & 78.9 & 78.4 & 65.9 & 76.6 & \textbf{75.9} & 78.6 & 76.4 & 65.5 \\
 & AKS & \underline{63.2} & \textbf{70.0} & 82.4 & 47.4 & 79.9 & \underline{80.6} & 64.8 & 75.7 & 70.4 & 80.4 & \textbf{83.6} & 61.0 \\
 & AdaQ & 60.0 & 67.1 & 82.4 & 48.9 & 79.9 & 75.5 & 66.1 & 73.7 & 72.2 & \underline{82.1} & 70.9 & 62.7 \\
 & QCA & 61.8 & 67.7 & 82.4 & 50.0 & \underline{81.1} & 77.0 & 66.1 & 76.3 & 68.5 & 80.4 & 78.2 & 62.7 \\
 & MDP$^3$ & 60.4 & 67.4 & 83.3 & 49.3 & 80.5 & 76.3 & 67.8 & 74.3 & 70.4 & \underline{82.1} & 80.0 & 61.0 \\
 & FOCUS & 54.0 & 61.7 & 82.9 & 43.7 & 78.0 & 73.4 & 63.7 & 70.3 & 70.4 & 75.0 & 70.9 & 52.5 \\
 & DIG & 60.0 & 67.7 & \underline{84.2} & 47.0 & 79.9 & 79.1 & 68.1 & 75.7 & 70.4 & \underline{82.1} & 74.5 & 52.5 \\
 & A.I.R. & 58.9 & 62.9 & 80.6 & 38.8 & 75.5 & 70.5 & 63.2 & 71.8 & 66.7 & 80.4 & 65.5 & 46.3 \\
 & LENS & 61.1 & 68.7 & 82.4 & 48.5 & \textbf{82.0} & 79.1 & \textbf{69.6} & \underline{77.4} & \underline{74.1} & 78.6 & 76.4 & \underline{67.2} \\
 & GIFT & \textbf{63.5} & \underline{69.0} & \underline{84.2} & 46.3 & 79.9 & 76.3 & 65.9 & 74.9 & 70.4 & \textbf{83.9} & \underline{81.8} & 54.2 \\
 & KFC & 60.7 & 68.1 & 82.4 & \textbf{51.5} & 79.9 & 79.1 & 67.2 & 73.4 & 70.4 & \underline{82.1} & \underline{81.8} & 58.2 \\
 & WFS-SB & 62.1 & 67.4 & 81.1 & \underline{50.4} & 78.9 & 79.9 & \underline{68.3} & 74.6 & 70.4 & \underline{82.1} & 76.4 & 63.3 \\
 & FRAG & 60.0 & 67.1 & 82.4 & 48.1 & 80.5 & \textbf{81.3} & 67.4 & 74.9 & 68.5 & 78.6 & 76.4 & 62.1 \\
 & OTFS & 62.5 & 67.4 & 83.8 & 48.5 & \underline{81.1} & \underline{80.6} & \underline{68.3} & 77.1 & 70.4 & 80.4 & 78.2 & 64.4 \\
 & ProbFaS & 62.5 & 67.7 & 82.4 & 47.4 & 80.2 & \underline{80.6} & 67.8 & \textbf{78.0} & 70.4 & \underline{82.1} & 76.4 & \textbf{67.8} \\
\bottomrule
\end{tabular}}
\end{table*}

\begin{table*}[t]
\centering
\caption{Accuracy (\%) per LongVideoBench question category, Qwen3-VL-8B, for every frame budget $k$ (number of questions in brackets). Level-1 (perception) categories: S2E, S2O, S2A, E2O, O2E, T2E, T2O, T2A; the others are level-2 (relation). Best per column and budget in \textbf{bold}, second \underline{underlined}.}
\label{tab:suppl_lvb_cats}
\small
\setlength{\tabcolsep}{4pt}
\resizebox{!}{0.4\textheight}{%
\begin{tabular}{llccccccccccccccccc}
\toprule
$k$ & Method & E2O (65) & E3E (94) & O2E (87) & O3O (66) & S2A (88) & S2E (93) & S2O (72) & SAA (72) & SOS (81) & SSS (97) & T2A (79) & T2E (65) & T2O (76) & T3E (73) & T3O (74) & TAA (82) & TOS (73) \\
\midrule
\multirow{15}{*}{8} & Uniform & 58.5 & 66.0 & 62.1 & 56.1 & 69.3 & 61.3 & 51.4 & 48.6 & 64.2 & 40.2 & 53.2 & 53.8 & 63.2 & 46.6 & 52.7 & 52.4 & 32.9 \\
 & BOLT & 66.2 & 64.9 & 66.7 & 54.5 & 73.9 & 67.7 & 59.7 & 55.6 & 65.4 & 34.0 & 64.6 & 66.2 & 60.5 & 52.1 & 48.6 & \textbf{56.1} & 31.5 \\
 & AdaQ & 63.1 & 63.8 & \textbf{71.3} & 51.5 & 79.5 & 71.0 & 62.5 & 51.4 & 56.8 & 37.1 & 70.9 & 60.0 & 68.4 & 45.2 & 48.6 & 53.7 & \textbf{42.5} \\
 & QCA & 66.2 & 64.9 & 67.8 & 42.4 & 80.7 & 74.2 & 66.7 & 52.8 & 59.3 & 35.1 & 73.4 & 66.2 & \textbf{75.0} & 42.5 & 40.5 & \underline{54.9} & 39.7 \\
 & MDP$^3$ & 63.1 & 66.0 & 62.1 & 54.5 & 79.5 & 68.8 & 62.5 & 52.8 & 66.7 & \underline{46.4} & 70.9 & 61.5 & 60.5 & 45.2 & \underline{59.5} & 46.3 & 35.6 \\
 & FOCUS & 70.8 & 64.9 & \underline{70.1} & 50.0 & 79.5 & 72.0 & 68.1 & 52.8 & 63.0 & 29.9 & 70.9 & 64.6 & 71.1 & 47.9 & 50.0 & \textbf{56.1} & 35.6 \\
 & DIG & 61.5 & 62.8 & 65.5 & 53.0 & 75.0 & 67.7 & 65.3 & 56.9 & 60.5 & 43.3 & 68.4 & 61.5 & 69.7 & \textbf{54.8} & 55.4 & 53.7 & 38.4 \\
 & A.I.R. & \underline{73.8} & \underline{67.0} & \textbf{71.3} & \textbf{62.1} & 79.5 & 75.3 & 62.5 & \textbf{61.1} & \underline{67.9} & 35.1 & \textbf{81.0} & \underline{67.7} & 68.4 & 38.4 & 51.4 & \underline{54.9} & 38.4 \\
 & LENS & \underline{73.8} & \textbf{68.1} & 65.5 & 47.0 & 79.5 & 69.9 & 68.1 & 55.6 & 59.3 & 36.1 & 70.9 & 66.2 & 71.1 & 41.1 & 54.1 & \underline{54.9} & 37.0 \\
 & GIFT & 61.5 & \textbf{68.1} & 63.2 & 48.5 & 72.7 & 71.0 & 59.7 & \underline{58.3} & 65.4 & 41.2 & 70.9 & 66.2 & 61.8 & 46.6 & 55.4 & \textbf{56.1} & 31.5 \\
 & KFC & 56.9 & 61.7 & 67.8 & 47.0 & 73.9 & 64.5 & 61.1 & 52.8 & 56.8 & 40.2 & 70.9 & 63.1 & 55.3 & 43.8 & 50.0 & 52.4 & \underline{41.1} \\
 & WFS-SB & 72.3 & 64.9 & 66.7 & \underline{57.6} & \underline{83.0} & 68.8 & 69.4 & 54.2 & 61.7 & 42.3 & 75.9 & \textbf{70.8} & \textbf{75.0} & \underline{53.4} & \textbf{64.9} & \underline{54.9} & 34.2 \\
 & FRAG & \textbf{80.0} & 63.8 & 67.8 & 53.0 & 79.5 & \textbf{77.4} & \underline{75.0} & 52.8 & 65.4 & 34.0 & 74.7 & 64.6 & \underline{73.7} & 47.9 & 48.6 & 52.4 & 38.4 \\
 & OTFS & 69.2 & 66.0 & 69.0 & 56.1 & 80.7 & \underline{76.3} & \textbf{76.4} & 54.2 & \textbf{69.1} & \textbf{47.4} & \underline{77.2} & \textbf{70.8} & \underline{73.7} & 47.9 & 51.4 & 52.4 & \underline{41.1} \\
 & ProbFaS & \textbf{80.0} & \textbf{68.1} & \textbf{71.3} & 56.1 & \textbf{85.2} & 72.0 & 73.6 & \textbf{61.1} & \textbf{69.1} & \textbf{47.4} & 75.9 & \underline{67.7} & 71.1 & 52.1 & 56.8 & \underline{54.9} & 38.4 \\
\midrule
\multirow{15}{*}{16} & Uniform & 61.5 & 68.1 & 63.2 & 56.1 & 73.9 & 67.7 & 54.2 & 55.6 & 64.2 & 44.3 & 57.0 & 61.5 & 57.9 & 49.3 & 54.1 & 54.9 & 35.6 \\
 & BOLT & 72.3 & 62.8 & 66.7 & 56.1 & 76.1 & 66.7 & 63.9 & 56.9 & 65.4 & 40.2 & 68.4 & 66.2 & 65.8 & 50.7 & 54.1 & \underline{58.5} & 35.6 \\
 & AdaQ & 73.8 & 69.1 & \underline{73.6} & \textbf{63.6} & 78.4 & 61.3 & 66.7 & 62.5 & 59.3 & 35.1 & 68.4 & 61.5 & 68.4 & 52.1 & 44.6 & \underline{58.5} & 39.7 \\
 & QCA & 64.6 & 66.0 & 70.1 & 56.1 & 78.4 & 68.8 & 69.4 & 59.7 & 64.2 & 47.4 & 72.2 & 67.7 & \underline{73.7} & 42.5 & 47.3 & 52.4 & 35.6 \\
 & MDP$^3$ & 67.7 & 67.0 & 70.1 & 51.5 & 76.1 & 65.6 & 58.3 & 61.1 & \underline{72.8} & 48.5 & 67.1 & 67.7 & 60.5 & 50.7 & \textbf{63.5} & 54.9 & 30.1 \\
 & FOCUS & 69.2 & \underline{70.2} & \underline{73.6} & 51.5 & 81.8 & 71.0 & \underline{72.2} & 55.6 & 65.4 & 37.1 & 72.2 & 66.2 & 71.1 & 45.2 & 54.1 & 56.1 & 37.0 \\
 & DIG & 69.2 & 62.8 & 67.8 & 60.6 & 79.5 & \underline{73.1} & 59.7 & \underline{63.9} & 64.2 & 42.3 & 75.9 & 67.7 & 68.4 & 50.7 & 54.1 & 56.1 & 38.4 \\
 & A.I.R. & 72.3 & 68.1 & 72.4 & 60.6 & 80.7 & \textbf{74.2} & 62.5 & \textbf{65.3} & 66.7 & 46.4 & \underline{79.7} & 67.7 & 69.7 & 42.5 & 54.1 & \textbf{63.4} & 30.1 \\
 & LENS & 75.4 & \underline{70.2} & 63.2 & 59.1 & \textbf{86.4} & 72.0 & 66.7 & 62.5 & 66.7 & 48.5 & 74.7 & 61.5 & \underline{73.7} & \underline{53.4} & 54.1 & \underline{58.5} & \textbf{42.5} \\
 & GIFT & 67.7 & 61.7 & 70.1 & \underline{62.1} & 78.4 & \underline{73.1} & 62.5 & \textbf{65.3} & 66.7 & 47.4 & 72.2 & 60.0 & 67.1 & \textbf{54.8} & 55.4 & 54.9 & 27.4 \\
 & KFC & 69.2 & 63.8 & 67.8 & 51.5 & 75.0 & 66.7 & 63.9 & 56.9 & 65.4 & 39.2 & 75.9 & \textbf{72.3} & \underline{73.7} & 46.6 & 47.3 & 57.3 & \underline{41.1} \\
 & WFS-SB & 73.8 & 68.1 & 72.4 & 59.1 & 81.8 & \textbf{74.2} & 70.8 & 58.3 & 61.7 & 48.5 & 75.9 & 66.2 & 71.1 & 50.7 & \underline{62.2} & \underline{58.5} & 37.0 \\
 & FRAG & \underline{78.5} & 67.0 & 71.3 & 57.6 & 79.5 & \underline{73.1} & 70.8 & 56.9 & 65.4 & 37.1 & 77.2 & 67.7 & 71.1 & 52.1 & 48.6 & 52.4 & 39.7 \\
 & OTFS & 75.4 & 69.1 & \textbf{75.9} & 56.1 & \underline{83.0} & 68.8 & 66.7 & 56.9 & 71.6 & \underline{50.5} & \textbf{82.3} & 64.6 & \textbf{75.0} & 46.6 & 55.4 & 56.1 & 39.7 \\
 & ProbFaS & \textbf{81.5} & \textbf{72.3} & 71.3 & \underline{62.1} & \underline{83.0} & \textbf{74.2} & \textbf{75.0} & 62.5 & \textbf{74.1} & \textbf{55.7} & 78.5 & \underline{70.8} & 71.1 & \textbf{54.8} & 59.5 & 57.3 & \textbf{42.5} \\
\midrule
\multirow{16}{*}{32} & Uniform & 64.6 & 70.2 & 65.5 & 60.6 & 75.0 & 66.7 & 61.1 & 52.8 & 72.8 & 48.5 & 62.0 & 61.5 & 67.1 & 47.9 & 54.1 & 57.3 & 32.9 \\
 & BOLT & 70.8 & \textbf{75.5} & 67.8 & 57.6 & 78.4 & 68.8 & 66.7 & 61.1 & 69.1 & 41.2 & 67.1 & 67.7 & 67.1 & 54.8 & 54.1 & \underline{59.8} & 37.0 \\
 & AKS & 69.2 & 66.0 & 70.1 & 57.6 & 79.5 & 67.7 & 65.3 & 58.3 & 70.4 & 47.4 & 70.9 & 61.5 & 73.7 & 47.9 & 48.6 & 51.2 & 35.6 \\
 & AdaQ & 70.8 & 68.1 & 69.0 & \textbf{65.2} & 72.7 & 68.8 & \underline{70.8} & 65.3 & 70.4 & 45.4 & 74.7 & 63.1 & 69.7 & 46.6 & 54.1 & 58.5 & \textbf{43.8} \\
 & QCA & 72.3 & 64.9 & 72.4 & \underline{63.6} & 81.8 & 75.3 & 68.1 & 63.9 & 70.4 & 49.5 & 74.7 & 67.7 & 77.6 & \textbf{58.9} & 52.7 & 57.3 & 39.7 \\
 & MDP$^3$ & 69.2 & 67.0 & 67.8 & 60.6 & 79.5 & 69.9 & 65.3 & 58.3 & 71.6 & 52.6 & 69.6 & \underline{69.2} & 64.5 & 50.7 & 56.8 & \textbf{63.4} & 35.6 \\
 & FOCUS & 75.4 & 70.2 & \underline{74.7} & 57.6 & \underline{83.0} & \textbf{78.5} & 68.1 & 55.6 & 70.4 & 32.0 & 77.2 & 61.5 & \textbf{81.6} & 50.7 & 52.7 & \textbf{63.4} & \underline{42.5} \\
 & DIG & \underline{78.5} & 64.9 & 67.8 & \underline{63.6} & 76.1 & 73.1 & 65.3 & 61.1 & \underline{75.3} & 45.4 & 70.9 & 67.7 & 71.1 & 53.4 & 59.5 & 56.1 & 41.1 \\
 & A.I.R. & 70.8 & 73.4 & 72.4 & \underline{63.6} & 78.4 & 74.2 & 65.3 & 61.1 & 64.2 & 42.3 & 77.2 & 64.6 & 68.4 & 41.1 & 51.4 & 57.3 & 38.4 \\
 & LENS & 69.2 & 68.1 & 66.7 & 57.6 & 80.7 & 75.3 & 65.3 & \textbf{68.1} & 65.4 & 43.3 & 77.2 & 66.2 & 71.1 & 52.1 & 55.4 & 53.7 & 35.6 \\
 & GIFT & 75.4 & 67.0 & 71.3 & 59.1 & 78.4 & 71.0 & 63.9 & 63.9 & 69.1 & \underline{55.7} & 68.4 & 66.2 & 67.1 & 54.8 & 58.1 & 57.3 & 37.0 \\
 & KFC & 69.2 & 73.4 & 69.0 & 54.5 & 81.8 & 69.9 & 65.3 & 61.1 & 70.4 & 51.5 & \textbf{79.7} & 66.2 & 72.4 & 47.9 & 50.0 & 58.5 & \underline{42.5} \\
 & WFS-SB & \underline{78.5} & 73.4 & \textbf{75.9} & 60.6 & 80.7 & 73.1 & \underline{70.8} & 63.9 & 69.1 & 53.6 & 77.2 & 66.2 & \underline{78.9} & 50.7 & \textbf{67.6} & 57.3 & 37.0 \\
 & FRAG & 73.8 & 64.9 & 71.3 & 57.6 & 77.3 & 76.3 & 69.4 & 65.3 & 67.9 & 39.2 & 75.9 & 64.6 & 72.4 & 53.4 & 55.4 & 52.4 & 39.7 \\
 & OTFS & 76.9 & 68.1 & \underline{74.7} & 62.1 & \underline{83.0} & 69.9 & 66.7 & 59.7 & 69.1 & 48.5 & \underline{78.5} & \textbf{70.8} & 72.4 & 54.8 & 59.5 & 56.1 & 41.1 \\
 & ProbFaS & \textbf{80.0} & \underline{74.5} & 67.8 & \underline{63.6} & \textbf{86.4} & \underline{77.4} & \textbf{75.0} & \underline{66.7} & \textbf{77.8} & \textbf{61.9} & \textbf{79.7} & 67.7 & 76.3 & \underline{56.2} & \underline{63.5} & 56.1 & 41.1 \\
\midrule
\multirow{16}{*}{64} & Uniform & 70.8 & 68.1 & 64.4 & 63.6 & 77.3 & 67.7 & 62.5 & \underline{66.7} & 77.8 & 49.5 & 69.6 & 66.2 & 72.4 & 46.6 & 59.5 & 53.7 & 37.0 \\
 & BOLT & 67.7 & 69.1 & 70.1 & 60.6 & 83.0 & 72.0 & 69.4 & 63.9 & 76.5 & 44.3 & 78.5 & 64.6 & 69.7 & 52.1 & 56.8 & 61.0 & 39.7 \\
 & AKS & 69.2 & 67.0 & 70.1 & 62.1 & \underline{85.2} & 72.0 & \underline{70.8} & 61.1 & \textbf{80.2} & 52.6 & 72.2 & 66.2 & 72.4 & 43.8 & 54.1 & 54.9 & \underline{42.5} \\
 & AdaQ & 69.2 & 73.4 & \textbf{74.7} & \underline{68.2} & 79.5 & 71.0 & 68.1 & 63.9 & 71.6 & 48.5 & 73.4 & 64.6 & 72.4 & 52.1 & 55.4 & 56.1 & 38.4 \\
 & QCA & 70.8 & 71.3 & \textbf{74.7} & 62.1 & 79.5 & 73.1 & 69.4 & 63.9 & 77.8 & 47.4 & 77.2 & 67.7 & 75.0 & \textbf{58.9} & 56.8 & 58.5 & 39.7 \\
 & MDP$^3$ & 73.8 & 70.2 & 71.3 & 62.1 & 71.6 & 72.0 & 58.3 & 61.1 & 69.1 & 48.5 & 70.9 & 67.7 & 72.4 & 52.1 & 58.1 & 58.5 & \underline{42.5} \\
 & FOCUS & \underline{78.5} & 71.3 & 71.3 & 56.1 & 83.0 & 74.2 & 66.7 & 58.3 & 64.2 & 38.1 & \textbf{81.0} & 63.1 & \textbf{84.2} & 43.8 & 58.1 & \underline{62.2} & \textbf{47.9} \\
 & DIG & \textbf{80.0} & 69.1 & 70.1 & \textbf{71.2} & 84.1 & 72.0 & 66.7 & 59.7 & 74.1 & 47.4 & 78.5 & \underline{70.8} & 75.0 & \underline{57.5} & 62.2 & 57.3 & 39.7 \\
 & A.I.R. & 75.4 & 74.5 & 72.4 & 66.7 & 79.5 & 72.0 & 66.7 & 63.9 & 66.7 & 40.2 & \underline{79.7} & 64.6 & 68.4 & 39.7 & 51.4 & 52.4 & 41.1 \\
 & LENS & 69.2 & \underline{76.6} & 70.1 & 66.7 & 83.0 & 74.2 & 62.5 & 65.3 & 71.6 & 52.6 & 77.2 & 67.7 & 68.4 & 54.8 & 58.1 & 54.9 & 32.9 \\
 & GIFT & 73.8 & 70.2 & 69.0 & 63.6 & 81.8 & 71.0 & 68.1 & 62.5 & 75.3 & \underline{56.7} & 69.6 & \textbf{72.3} & 73.7 & 53.4 & 56.8 & 61.0 & 41.1 \\
 & KFC & 73.8 & 74.5 & \textbf{74.7} & 62.1 & 81.8 & 72.0 & 62.5 & 62.5 & 76.5 & 55.7 & 77.2 & 66.2 & 72.4 & 52.1 & 62.2 & \textbf{63.4} & 35.6 \\
 & WFS-SB & 75.4 & 72.3 & \underline{73.6} & 59.1 & \underline{85.2} & \underline{77.4} & 66.7 & 65.3 & 72.8 & \underline{56.7} & \underline{79.7} & 69.2 & \underline{77.6} & \textbf{58.9} & \textbf{66.2} & 58.5 & 34.2 \\
 & FRAG & 75.4 & 68.1 & \textbf{74.7} & 60.6 & 80.7 & \textbf{78.5} & \textbf{72.2} & 59.7 & 69.1 & 37.1 & 77.2 & 63.1 & 75.0 & 53.4 & 60.8 & 61.0 & 39.7 \\
 & OTFS & 75.4 & 67.0 & 71.3 & 59.1 & 83.0 & 65.6 & 68.1 & 63.9 & 71.6 & 49.5 & 73.4 & 67.7 & 76.3 & 53.4 & \underline{63.5} & 58.5 & 39.7 \\
 & ProbFaS & \textbf{80.0} & \textbf{77.7} & \underline{73.6} & 63.6 & \textbf{87.5} & 75.3 & \underline{70.8} & \textbf{68.1} & \underline{79.0} & \textbf{59.8} & \textbf{81.0} & 69.2 & 75.0 & 47.9 & 58.1 & 58.5 & \underline{42.5} \\
\midrule
\multirow{16}{*}{128} & Uniform & 66.2 & 74.5 & 71.3 & 65.2 & 80.7 & 69.9 & 63.9 & 62.5 & 70.4 & 55.7 & 73.4 & 66.2 & 75.0 & 50.7 & 55.4 & 58.5 & 38.4 \\
 & BOLT & 73.8 & 72.3 & 69.0 & \textbf{71.2} & 84.1 & 69.9 & \textbf{70.8} & 62.5 & \underline{75.3} & 47.4 & \underline{78.5} & 63.1 & 71.1 & 50.7 & 55.4 & 59.8 & 39.7 \\
 & AKS & 73.8 & 72.3 & 71.3 & 62.1 & 84.1 & 71.0 & \textbf{70.8} & \underline{66.7} & 74.1 & 47.4 & 75.9 & 63.1 & 73.7 & 45.2 & 55.4 & 53.7 & 39.7 \\
 & AdaQ & 72.3 & 72.3 & \textbf{73.6} & 68.2 & 81.8 & 71.0 & \underline{69.4} & 63.9 & 70.4 & 54.6 & 75.9 & 58.5 & 68.4 & 49.3 & 54.1 & 59.8 & \textbf{43.8} \\
 & QCA & 73.8 & 71.3 & \underline{72.4} & \underline{69.7} & 84.1 & \underline{76.3} & \underline{69.4} & 61.1 & 74.1 & 49.5 & \underline{78.5} & \underline{67.7} & \underline{76.3} & 50.7 & 56.8 & 59.8 & 39.7 \\
 & MDP$^3$ & 69.2 & 73.4 & 71.3 & 63.6 & 77.3 & 65.6 & 61.1 & 62.5 & 72.8 & 53.6 & 68.4 & \underline{67.7} & 68.4 & 52.1 & 55.4 & 59.8 & 39.7 \\
 & FOCUS & \textbf{80.0} & 68.1 & \underline{72.4} & 65.2 & 79.5 & \textbf{80.6} & 68.1 & 59.7 & 69.1 & 34.0 & \textbf{82.3} & 66.2 & \textbf{77.6} & \underline{54.8} & 60.8 & \textbf{65.9} & \underline{41.1} \\
 & DIG & 75.4 & 69.1 & 70.1 & 68.2 & 84.1 & 73.1 & 66.7 & 65.3 & \textbf{80.2} & 51.5 & 77.2 & \textbf{72.3} & \underline{76.3} & 52.1 & 59.5 & 57.3 & \textbf{43.8} \\
 & A.I.R. & 73.8 & 74.5 & 69.0 & 65.2 & 83.0 & 71.0 & \underline{69.4} & 62.5 & 69.1 & 42.3 & \underline{78.5} & 64.6 & 69.7 & 42.5 & 52.7 & 53.7 & 38.4 \\
 & LENS & \underline{76.9} & 74.5 & \underline{72.4} & 66.7 & 83.0 & 74.2 & 65.3 & 65.3 & 63.0 & 55.7 & 75.9 & \underline{67.7} & 67.1 & 49.3 & 60.8 & 52.4 & 37.0 \\
 & GIFT & 69.2 & 74.5 & 71.3 & \underline{69.7} & 80.7 & 68.8 & 65.3 & 63.9 & 70.4 & 53.6 & 73.4 & 64.6 & 72.4 & 50.7 & 59.5 & 59.8 & 38.4 \\
 & KFC & 73.8 & 74.5 & \underline{72.4} & 65.2 & 80.7 & 72.0 & 63.9 & \textbf{69.4} & 71.6 & 55.7 & 73.4 & 63.1 & 72.4 & 53.4 & 58.1 & 58.5 & \underline{41.1} \\
 & WFS-SB & \underline{76.9} & \textbf{80.9} & 71.3 & 63.6 & \textbf{86.4} & \underline{76.3} & \textbf{70.8} & 65.3 & 71.6 & \textbf{61.9} & \underline{78.5} & \underline{67.7} & \underline{76.3} & \textbf{57.5} & \textbf{66.2} & 54.9 & 35.6 \\
 & FRAG & 75.4 & 72.3 & 71.3 & 68.2 & 81.8 & 74.2 & \underline{69.4} & 63.9 & 70.4 & 47.4 & 75.9 & 63.1 & 71.1 & 52.1 & 56.8 & \underline{62.2} & \textbf{43.8} \\
 & OTFS & \underline{76.9} & 73.4 & 70.1 & \underline{69.7} & 81.8 & 74.2 & 66.7 & 63.9 & \underline{75.3} & 54.6 & 75.9 & 63.1 & \underline{76.3} & 53.4 & \underline{64.9} & 61.0 & \textbf{43.8} \\
 & ProbFaS & 75.4 & \underline{75.5} & \underline{72.4} & \textbf{71.2} & \underline{85.2} & \underline{76.3} & \underline{69.4} & \textbf{69.4} & 74.1 & \underline{56.7} & \underline{78.5} & 64.6 & 73.7 & 53.4 & 60.8 & 61.0 & \textbf{43.8} \\
\bottomrule
\end{tabular}}
\end{table*}

\end{document}